\documentclass[onecolumn,journal,compsoc]{IEEEtran}
\usepackage[T1]{fontenc}
\usepackage[latin9]{inputenc}
\usepackage{geometry}
\usepackage{color}
\usepackage{array}
\usepackage{float}
\usepackage{multirow}
\usepackage{amsmath}
\usepackage{amssymb}
\usepackage{graphicx}
\usepackage[unicode=true,
 bookmarks=true,bookmarksnumbered=true,bookmarksopen=true,bookmarksopenlevel=1,
 breaklinks=false,pdfborder={0 0 0},pdfborderstyle={},backref=false,colorlinks=false]
 {hyperref}
\hypersetup{pdftitle={Your Title},
 pdfauthor={Your Name},
 pdfpagelayout=OneColumn, pdfnewwindow=true, pdfstartview=XYZ, plainpages=false}

\makeatletter

\providecommand{\tabularnewline}{\\}
\floatstyle{ruled}
\newfloat{algorithm}{tbp}{loa}
\providecommand{\algorithmname}{Algorithm}
\floatname{algorithm}{\protect\algorithmname}

\let\oldforeign@language\foreign@language
\DeclareRobustCommand{\foreign@language}[1]{%
  \lowercase{\oldforeign@language{#1}}}

\usepackage[caption=false,font=footnotesize]{subfig}
\usepackage{cite}

\makeatother

\begin{document}
\title{An Adaptive EM Accelerator for Unsupervised Learning of Gaussian Mixture
Models}
\author{Truong Nguyen, Guangye Chen,~and~Luis Chacón\IEEEcompsocitemizethanks{\IEEEcompsocthanksitem{Truong Nguyen is with the Applied Mathematics and Plasma Physics Group,
Theoretical Division, Los Alamos National Laboratory, NM 87545, USA,
e-mail: \protect\href{http://tbnguyen@lanl.gov}{tbnguyen@lanl.gov}.}\IEEEcompsocthanksitem{Guangye Chen is with the Applied Mathematics and Plasma Physics Group,
Theoretical Division, Los Alamos National Laboratory, NM 87545, USA,
e-mail: \protect\href{http://gchen@lanl.gov}{gchen@lanl.gov}.}\IEEEcompsocthanksitem{Luis Chacón is with the Applied Mathematics and Plasma Physics Group,
Theoretical Division, Los Alamos National Laboratory, NM 87545, USA,
e-mail: \protect\href{http://chacon@lanl.gov}{chacon@lanl.gov}.}}
\thanks{Manuscript submitted September 25, 2020.}
}
\markboth{IEEE Transactions on Pattern Analysis and Machine Intelligence}{Nguyen, Chen and Chacón : An Adaptive EM Accelerator for Unsupervised
Learning of Gaussian Mixture Models.}

\IEEEtitleabstractindextext{
\begin{abstract}
We propose an Anderson Acceleration (AA) scheme for the \emph{adaptive}
Expectation-Maximization (EM) algorithm for unsupervised learning
a finite mixture model from multivariate data (Figueiredo and Jain
2002). The proposed algorithm is able to determine the optimal number
of mixture components autonomously, and converges to the optimal solution
much faster than its non-accelerated version. The success of the AA-based
algorithm stems from several developments rather than a single breakthrough
(and without these, our tests demonstrate that AA fails catastrophically).
To begin, we ensure the monotonicity of the likelihood function (a
the key feature of the standard EM algorithm) with a recently proposed
monotonicity-control algorithm (Henderson and Varahdan 2019), enhanced
by a novel monotonicity test with little overhead. We propose nimble
strategies for AA to preserve the positive definiteness of the Gaussian
weights and covariance matrices strictly, and to conserve up to the
second moments of the observed data set exactly. Finally, we employ
a K-means clustering algorithm using the gap statistic to avoid excessively
overestimating the initial number of components, thereby maximizing
performance. We demonstrate the accuracy and efficiency of the algorithm
with several synthetic data sets that are mixtures of Gaussians distributions
of known number of components, as well as data sets generated from
particle-in-cell simulations. Our numerical results demonstrate speed-ups
with respect to non-accelerated EM of up to $60\times$ when the exact
number of mixture components is known, and between a few and more
than an order of magnitude with component adaptivity.
\end{abstract}

}

\thispagestyle{empty}
\addtocounter{page}{-1}
\noindent Preprint Notice:\\[0.25em]
\noindent © 2020 IEEE. Personal use of this material is permitted. Permission from IEEE must be obtained for all other uses, in any current or future media, including reprinting/republishing this material for advertising or promotional purposes, creating new collective works, for resale or redistribution to servers or lists, or reuse of any copyrighted component of this work in other works.

\pagebreak

\maketitle

\IEEEpeerreviewmaketitle{}

\IEEEpubid{}


\section{Introduction}

\IEEEPARstart{T}{he} Gaussian mixture model (GMM) is a probabilistic
model that assumes all the observed data points are generated from
a mixture of a finite number of Gaussian (normal) distributions \cite{titterington1985statistical,mclachlan2004finite,mclachlan2019finite,fruhwirth2019handbook}.
It has wide applications in pattern recognition and unsupervised machine
learning \cite{bishop2006pattern,murphy2012probabilistic}, big data
analytics \cite{liu2016bigdata,ullah2019BayesianMM,BouchonMeunier2006informationprocessing,yu2014},
and image segmentation and denoising \cite{Bi2018GMM-IS,Deledalle2018GGMM-ID,Kalti2014-IS,TMNguyen2011DGMM-IS,Farnoosh2008GMM-IS},
as well as recent applications in applied and computational physics,
e.g., gas kinetic \cite{alekseenko2018bci} and plasma kinetic algorithms
\cite{chen2020CR-EM-GM,dupuis2020characterizing}. Some other applications
of GMM can be found in \cite{Barb2011GMM-Radiology,Reynolds2000GMM-SpeakerVerification,Plataniotis200GMM-SignalProcess,yu2015gaussian}.
Of interest here is a parametric probability density function family
of the form $f(\boldsymbol{x};\boldsymbol{\theta})=\,\sum_{k=1}^{K}\omega_{k}G_{k}(\boldsymbol{x};\boldsymbol{\theta}_{k})$,
where $G_{k}$ is a Gaussian distribution parameterized by $\boldsymbol{\theta}_{k}$,
$\omega_{k}$ is the a positive weight under the constraint of $\sum_{k=1}^{K}\omega_{k}=1$,
and $K$ is the total number of Gaussian components. A common iterative
approach to estimate the parameters of the Gaussian distributions
in GMM is the Expectation-Maximization (EM) algorithm, which is based
on the maximum likelihood principle \cite{dempster1977maximum,redner1984mixture,mclachlan2004finite,mclachlan2007algorithm}.
EM is well known for its robustness, as it is guaranteed to converge
monotonically to a local maximum. The standard EM algorithm for GMM
(EM-GMM) is conceptually simple and easy to implement. However, the
performance is highly dependent on the initial guess of the Gaussian
parameters and the separation of the Gaussian components (convergence
can be very slow when Gaussian components are not well separated).
A further difficulty is that the maximum likelihood principle alone
cannot determine the number of components \cite{titterington1985statistical,mclachlan2014number}.
The number of Gaussian components is usually unknown in practice,
which makes the proper choice of the number of Gaussian components
crucial for the optimal performance of EM-GMM. Choosing too many components
would result in overfitting the data set, and potentially worsening
the convergence rate of EM-GMM. Alternatively, choosing too few components
could result in under-fitting, leading to model predictions that may
miss important structures of the data.

To resolve the number of components issue in GMM, a recent and well
adopted \emph{adaptive} EM algorithm was proposed \cite{figueiredo2000unsupervised,figueiredo2002unsupervised}
that can automatically converge on the optimal number of Gaussian
components. By employing a ``minimum-message-length (MML)'' Bayesian
information criterion \cite{wallace2005statistical,hansen2001modelselection},
the method allows users to start with a relatively large number of
Gaussian components and gradually converge to the optimal number of
groups during EM iterative procedure. Adaptive EM introduces a modified
M-step for the Gaussian weights capable of eliminating unimportant
components, which is only a simple extension of standard EM-GMM and
makes it attractive for practitioners when compared to other methods
(e.g., variational Bayesian model \cite{corduneanu2001variational},
and those reviewed in Ref. \cite{mclachlan2014number}). Several drawbacks
of standard EM, such as sensitivity to initialization, and possible
convergence to singular solutions, are also largely avoided \cite{figueiredo2000unsupervised,figueiredo2002unsupervised}.
However, the slow convergence problems of standard EM were left unaddressed.

Many methods have been proposed to date to accelerate the convergence
rate of standard\emph{ (non-adaptive)} EM \cite{mclachlan2007algorithm,lange2013optimization},
which may be roughly categorized into EM extension algorithms and
gradient-based algorithms. The algorithms in the first category are
mainly developed within the EM framework based on statistical considerations,
which include ECM \cite{meng1993maximum}, ECME \cite{liu1994ecme},
SAGE \cite{fessler1994space}, AECM \cite{meng1997algorithm}, PX-EM
\cite{liu1998parameter}, and CEMM \cite{celeux2001component}, etc.
We note that Figueiredo and Jain's original paper for adaptive EM-GMM
employed CEMM, but mainly for the purpose of avoiding elimination
of all Gaussian components at the beginning of the iteration in some
situations \cite{figueiredo2002unsupervised}. Algorithms in the second
category treat standard EM as a fixed-point iteration map, and seek
accelerations using various gradient-based methods, including Aitken-Steffensen-type
\cite{wolfe1970pattern,dempster1977maximum,louis1982finding,varadhan2008squarem,berlinet2007acceleration},
conjugate gradient (CG) \cite{jamshidian1993conjugate,salakhutdinov2003optimization,he2012dynamic},
quasi-Newton (QN) \cite{lange1995quasi,jamshidian1997acceleration,henderson2019daarem,zhou2011quasi}
and Newton-Raphson \cite{meilijson1989fast,lange1995gradient} methods,
etc. In the context of the GMM density estimation, the Newton-Raphson
method requires computation of the second derivative of the log-likelihood
function (i.e., the Hessian matrix \cite{mclachlan2004finite}), which
is considered too complicated to be practical, especially when compared
to standard EM. QN and CG methods avoid the difficulty by using approximations
that involve only the first derivatives of the log-likelihood function
(i.e., the score function \cite{mclachlan2004finite}), which are
much easier to obtain, and still often gain much acceleration over
the standard EM when it is very slow. However, one common feature
for many gradient-based EM-accelerators is that they require line-searching
and careful monitoring or safeguards, a consequence of the lack of
automatic monotone convergence of the likelihood function. The line-search
step (e.g., needed in Refs. \cite{jamshidian1993conjugate,lange1995quasi,jamshidian1997acceleration,salakhutdinov2003optimization,zhou2011quasi})
determines the step-size in some gradient direction in order to make
progress in increasing the likelihood function. This often requires
multiple likelihood function evaluations, which is one of the most
expensive operations in EM-GMM due to the need to evaluate multiple
exponential functions on the sample data. The situation is similar
for other strategies such as globalization \cite{varadhan2008squarem},
algorithm restart \cite{berlinet2007acceleration}, or monitoring
of the progress \cite{zhou2011quasi}. The need for many additional
likelihood function evaluations can offset much (or even all) of the
algorithmic acceleration afforded by these solver strategies, leading
to virtually no wall-clock-time advantage.

In this study, we explore acceleration of EM-GMM using Anderson Acceleration
(AA) \cite{anderson1965}, which can be viewed as a variant of QN
\cite{fang2009two}. We note that AA has been explored before for
EM-GMM \cite{walker2011anderson,plasse2013algorithm}. In Ref. \cite{walker2011anderson},
a reduced mixture problem (i.e., estimating only the means of a three-component
univariate Gaussian mixture) was successfully accelerated by AA. Later,
Ref. \cite{plasse2013algorithm} successfully extended the method
to two-component multivariate Gaussian mixtures, suggesting  potential
for AA as an EM-accelerator for GMM. However, both studies assumed
a known number of mixture components, and both employed a large number
of samples in their tests ($10^{5}$ in Ref. \cite{walker2011anderson}
and $10^{6}$ in Ref. \cite{plasse2013algorithm}), presumably robustifying
the AA iterations. For smaller and more realistic data sets and more
complicated applications, as in some of our tests, the AA implementation
in Refs. \cite{walker2011anderson,plasse2013algorithm} may break
positiveness of Gaussian parameters and may converge to sub-optimal
solutions \cite{henderson2019daarem} and even fail catastrophically
(as we will show). To remedy those drawbacks, we will employ a restarted/regularized
version of AA to accelerate \emph{adaptive} EM while monitoring the
monotonicity of the likelihood function (a critical step \cite{henderson2019daarem}),
which has not previously been tested with GMM. In fact, to the best
of our knowledge, no gradient-based EM accelerators have been applied
to the adaptive EM-GMM algorithm. 

The success of our proposed algorithm stems from various ingredients.
Firstly, we improve the monotonicity control step proposed in Ref.
\cite{henderson2019daarem} with a new, very low overhead monotonicity
test. Secondly, we ensure that the algorithm preserves positive-definiteness
of Gaussian weights and covariance matrices, and conserves up to second
moments of the observed data set exactly, just as in a standard EM.
Lastly, it is well known that choosing the initial number of Gaussians
well can significantly affect the performance of adaptive EM. To obtain
a good estimate for the initial number of components, we complement
our method with a reliable initialization routine using K-means clustering
with the gap statistic \cite{Tibshirani2001gapstatistics}. As a result,
our AA-based algorithm delivers significant efficiency gains versus
the non-accelerated EM while converging to the same (optimal) solution.

The rest of the paper is organized as follows. Section \ref{sec:EM-GMM-algorithms}
introduces the basic concepts of both standard and adaptive EM-GMM.
Section \ref{sec:Nonlinear-Anderson-acceleration-of-EM} briefly summarizes
a recent attempt at EM acceleration, the exact-line-search (ELS) EM
(which we will use as a benchmark for performance). We then present
an overview of AA, its challenges of aggressive application to EM-GMM,
and strategies to address these challenges proposed in the literature.
Section \ref{sec:AMEM} proposes our solution for accelerating adaptiv\textcolor{black}{e
EM with monotonicity control for the likelihood function. Key elements
include the design of an efficient approach for monotonicity control
in AA, the use of a regularization term \cite{henderson2019daarem}
to combine the robustness of EM and local convergence speed of AA,
and the careful selection of the initial number of Gaussian components
using K-means clustering based on the gap statistic approach \cite{Tibshirani2001gapstatistics}.
Also described are our solutions for strict moment conservation and
preservation of positive-definiteness of Gaussian weights and covariance
matrices. Section \ref{sec:Numerical-Results}} demonstrates the fidelity
and efficiency of the proposed acceleration scheme over its non-accelerated
counterpart for several synthetic data sets, as well as real data
sets generated from particle-in-cell simulations \cite{chen2020CR-EM-GM},
where we demonstrate significant algorithmic and wall-clock-time speed-ups.
Finally, we conclude in Section \ref{sec:Conclusion}.

\section{The Expectation Maximization (EM) algorithm for Gaussian mixtures
(EM-GMM)\label{sec:EM-GMM-algorithms}}

A Gaussian mixture (GM) of $K$ components is defined to be a convex
combination of $K$ Gaussian distributions $G_{k},\,k=1,\cdots,K$
of the following form

\begin{equation}
f(\boldsymbol{x})=\,\sum_{k=1}^{K}\omega_{k}G_{k}(\boldsymbol{x};\boldsymbol{\mu}_{k},\boldsymbol{\Sigma}_{k})\,,\label{eq:mix-dist}
\end{equation}
where $\omega_{k},\,\boldsymbol{\mu}_{k},\,\boldsymbol{\Sigma}_{k}$
are the weight, mean and covariance matrix, respectively, of the $k\mbox{th}$
Gaussian in the mixture. Note that $\omega_{k}\geq0$, and $\boldsymbol{\Sigma}_{k}$
are symmetric-positive-definite matrices. The Gaussian distribution
$G_{k}$ is defined as
\begin{equation}
G_{k}(\boldsymbol{x};\boldsymbol{\mu}_{k},\boldsymbol{\Sigma}_{k})=\frac{1}{\sqrt{(2\pi)^{D}\vert\boldsymbol{\Sigma}_{k}\vert}}e^{-\frac{1}{2}(\boldsymbol{x}-\boldsymbol{\mu}_{k})^{T}\boldsymbol{\Sigma}_{k}^{-1}(\boldsymbol{x}-\boldsymbol{\mu}_{k})}
\end{equation}
where both $\boldsymbol{x}$ and $\boldsymbol{\mu}$ are $D$-dimensional
column vectors, the superscript $T$ denotes transpose, and $\vert\boldsymbol{\Sigma}_{k}\vert$
is the determinant of the covariance matrix.

\subsection{The standard EM-GMM algorithm\label{subsec:The-standard-EM}}

The goal is to find the Gaussian parameters $\boldsymbol{\theta}=(\boldsymbol{\theta}_{1},...,\boldsymbol{\theta}_{K})$
where $\boldsymbol{\theta}_{k}=(\omega_{k},\boldsymbol{\mu}_{k},\boldsymbol{\Sigma}_{k})$
for $k=1,...,K$ that maximize the log-likelihood of the Gaussian
mixture model \cite{everitt2014finite}. The log-likelihood is written
as
\begin{equation}
\begin{aligned}\mathcal{L}(\boldsymbol{\theta}) & =\mbox{ln}\bigg(\prod_{j=1}^{N}\bigg[f(\boldsymbol{x}_{j})\bigg]^{\zeta_{j}}\bigg)+\eta\,\bigg(\sum_{k=1}^{K}\omega_{k}-1\bigg)\\
 & =\sum_{j=1}^{N}\zeta_{j}\;\mbox{ln}\bigg(\sum_{k=1}^{K}\omega_{k}G_{k}(\boldsymbol{x};\boldsymbol{\mu}_{k},\boldsymbol{\Sigma}_{k})\bigg)+\eta\,\bigg(\sum_{k=1}^{K}\omega_{k}-1\bigg)\,,
\end{aligned}
\label{eq:llh-gmm}
\end{equation}
for $N$ independent samples $\boldsymbol{X}=(\boldsymbol{x}_{1},...,\boldsymbol{x}_{N})$
(presumably) drawn from $f(\boldsymbol{x})$, with each sample $\boldsymbol{x}_{j}$
having a weight $\zeta_{j}\,$. Here, $\eta\,\bigg(\sum_{k=1}^{K}\omega_{k}-1\bigg)$
is the Lagrange-multiplier term that enforces the normalization constraint
$\sum_{k=1}^{K}\omega_{k}=1$, i.e., $f(\boldsymbol{x})$ is normalized
to unity. We assume that $\sum_{j=1}^{N}\zeta_{j}=N$, and the sample
weights $\zeta_{j},\;j=1,\ldots,N$ account for the cases with non-identical
samples \cite{hasselblad1966estimation}.

In order to maximize the log-likelihood function $\mathcal{L}(\boldsymbol{\theta})$,
we solve the following score equations:

\begin{equation}
\frac{\partial\mathcal{L}(\boldsymbol{\theta})}{\partial\boldsymbol{\theta}}=0\,,\mbox{ and}\;\frac{\partial\mathcal{L}(\boldsymbol{\theta})}{\partial\eta}=0.\label{eq:dL-dtheta=00003Dzero}
\end{equation}
We obtain (see Ref. \cite{bishop2006pattern} and Appendix \ref{sec:Derivative-LLH-appendix}):

\begin{equation}
\boldsymbol{\mu}_{k}=\frac{1}{N_{k}}\sum_{j=1}^{N}r_{jk}\boldsymbol{x}_{j}\;,\label{eq:muk}
\end{equation}
 
\begin{equation}
\boldsymbol{\Sigma}_{k}=\frac{1}{N_{k}}\sum_{j=1}^{N}r_{jk}(\boldsymbol{x}_{j}-\boldsymbol{\mu}_{k})(\boldsymbol{x}_{j}-\boldsymbol{\mu}_{k})^{T},\label{eq:sigmak}
\end{equation}
\begin{equation}
\omega_{k}=\frac{N_{k}}{N},\label{eq:omegak}
\end{equation}
where $r_{jk}$ is the responsibility value of point $\boldsymbol{x}_{j}$
within the $k\mbox{th}$ Gaussian, and is defined as:

\begin{equation}
r_{jk}=\frac{\zeta_{j}\,\omega_{k}G_{k}(\boldsymbol{x}_{j};\,\boldsymbol{\mu}_{k},\boldsymbol{\Sigma}_{k})}{\sum_{l=1}^{K}\omega_{l}G_{l}(\boldsymbol{x}_{j};\,\boldsymbol{\mu}_{l},\boldsymbol{\Sigma}_{l})},\label{eq:responsibilities}
\end{equation}
with $N_{k}=\sum_{j=1}^{N}r_{jk}$ and $\zeta_{j}=\sum_{k=1}^{K}r_{jk}$.

EM-GMM is an iterative procedure to find the solution to (\ref{eq:muk}),
(\ref{eq:sigmak}) and (\ref{eq:omegak}), and can be done in the
following distinct steps until convergence:
\begin{itemize}
\item Expectation-step: Evaluate the responsibilities $r_{jk}$ for $j=1,\ldots,N$
and $k=1,\ldots,K$ using (\ref{eq:responsibilities}).
\item Maximization-step: Update the Gaussian parameters for $k=1,\ldots,K$
using (\ref{eq:muk}), (\ref{eq:sigmak}) and (\ref{eq:omegak}).
\end{itemize}
Convergence of the algorithm is assessed by monitoring the log-likelihood
function (\ref{eq:llh-gmm}). We note that several quantities used
in the computation of responsibilities for $\theta^{(it)}$, $r_{jk}^{(it)}\big(\theta^{(it)}\big)$,
can be re-used in the evaluation of $\mathcal{L}(\theta^{(it)})$
for efficiency. We remark that each EM iteration ensures the monotonic
increase of the likelihood function during the iteration \cite{dempster1977maximum},
and conserves up to the second moments of the data set \cite{behboodian1970mixture,chen2020CR-EM-GM}.

\subsection{The adaptive EM-GMM algorithm\label{subsec:The-adaptive-EM}}

In practice, the number of Gaussian components in the Gaussian mixture
is often unknown. Therefore, a way to select the proper number of
mixture components is needed. Component adaptivity can be accomplished
by employing the minimum message length (MML) criterion \cite{figueiredo2000unsupervised,hansen2001modelselection,figueiredo2002unsupervised,mackay2003information,wallace2005statistical,chen2020CR-EM-GM}
to penalize the log-likelihood function. Instead of using (\ref{eq:llh-gmm})
for the log-likelihood, we use the following penalized log-likelihood
function:

\begin{equation}
\begin{aligned}\mathcal{PL}(\boldsymbol{\theta}) & =\sum_{j=1}^{N}\zeta_{j}\,\mbox{ln}\bigg(\sum_{k=1}^{K}\omega_{k}G_{k}(\boldsymbol{x};\boldsymbol{\mu}_{k},\boldsymbol{\Sigma}_{k})\bigg)+\eta\,\bigg(\sum_{k=1}^{K}\omega_{k}-1\bigg)\\
 & -\frac{d}{2}\mbox{ln}(N)-\frac{T}{2}\sum_{k=1}^{K}\mbox{ln}(\omega_{k})\,,
\end{aligned}
\label{eq:pllh-gmm}
\end{equation}
where $d$ is the total number of parameters in the Gaussian mixture
and $T=\frac{D(D+3)}{2}$. For a detailed derivation of (\ref{eq:pllh-gmm}),
we refer the reader to Ref. \cite{chen2020CR-EM-GM}. The last two
terms in (\ref{eq:pllh-gmm}) are the penalization terms, and they
play an important role in determining the optimal number of components
by removing unnecessary Gaussian components in order to avoid over-fitting
the data.

In order to maximize the penalized log-likelihood function (\ref{eq:pllh-gmm}),
the same approach as described in Section \ref{subsec:The-standard-EM}
results in the same formulas for updating the Gaussians' means $\boldsymbol{\mu}_{k}$
and covariance matrices $\boldsymbol{\Sigma}_{k}$ in the M-step,
i.e., (\ref{eq:muk}) and (\ref{eq:sigmak}), respectively. However,
due to the presence of the penalization terms, the equation for the
Gaussians' weights is modified as (see Appendix \ref{sec:Derivative-LLH-appendix}):
\begin{equation}
\omega_{k}=\frac{N_{k}-\frac{T}{2}}{N-\frac{TK}{2}}\;,\label{eq:adapt_omegak}
\end{equation}
provided that $N_{k}>T/2$. If $N_{k}<T/2$, $\omega_{k}<0$, indicating
that the $k\mbox{th}$ Gaussian should be killed. Also, (\ref{eq:adapt_omegak})
advises that one should start with more Gaussian components than the
exact number of Gaussian components in the GMM \cite{rousseau2011asymptotic}.

Putting things together, assuming $K$ components at each iteration
of adaptive EM-GMM, we do the following:
\begin{enumerate}
\item Evaluate the responsibilities $r_{jk}$ using (\ref{eq:responsibilities}).
\item Compute $\omega_{k}^{(*)}=\mbox{max}\bigg(\frac{N_{k}-T/2}{N-TK/2},0\bigg)$.
\item If $\omega_{k}^{(*)}=0$ then kill the $k\mbox{th}$ Gaussian: set
$\boldsymbol{\mu}_{k}=0$, $\boldsymbol{\Sigma}_{k}=0$ and $K=K-1.$
Otherwise, update the mean and covariance matrix of the $k\mbox{th}$
Gaussian using (\ref{eq:muk}) and (\ref{eq:sigmak}), respectively.
\item Re-normalize the Gaussian weights: $\omega_{k}=\omega_{k}^{(*)}\big/\big(\sum_{k=1}^{K}\omega_{k}^{(*)}\big)$.
\item Check for convergence by monitoring the penalized log-likelihood function
(\ref{eq:pllh-gmm}).
\end{enumerate}
Steps 1 to 5 are repeated until convergence. We refer to Step 1 as
the E-step and Steps 2-4 as the M-step. In practice, we can perform
the iterations in a simultaneous approach or a component-wise approach
\cite{celeux2001component}. In the simultaneous EM, we perform the
E-step for all available Gaussian components in the mixture, then
update their parameters in the M-step. In the component-wise EM, we
perform the E-M steps for one Gaussian component and then move on
the the next one until we reach the final component in the mixture.
The simultaneous approach is faster but the algorithm could possibly
kill all Gaussians at once in some situations (e.g., when the starting
number of components, $K_{init}$, is much larger than the model's
exact number of components, $K_{model}$) \cite{figueiredo2002unsupervised}.
The component-wise approach is slower (because it requires updating
the Gaussian mixture's probability density function once a component
is updated) but it can prevent such a problem \cite{figueiredo2002unsupervised}.
For efficiency, here we follow the simultaneous approach while relying
on an extended form of K-means clustering \cite{raykov2016adaptiveKmeans,Darken1990fastadaptkmeans,scikit-learn-selectNG,Rousseeuw1987silhouette,Nanjundan2019silhouette,Tibshirani2001gapstatistics}
to provide a good initial guess for the number of components. This
will be discussed later in this study.

We note that both the accelerated and non-accelerated versions of
the adaptive EM-GMM iteration do not conserve the moments of the observed
data set due to the presence of the penalization terms. To recover
conservation, we perform a final standard EM-GMM step after convergence
\cite{chen2020CR-EM-GM}.

\section{Nonlinear acceleration of the EM algorithm\label{sec:Nonlinear-Anderson-acceleration-of-EM}}

\subsection{State of the art in accelerated EM-GMM: Exact Line Search method
(ELS-EM)\label{subsec:ELS-EM}}

A recent attempt to accelerate EM-GMM is the so-called exact line
search EM (ELS-EM) introduced by Xiang et al. \cite{xiang2020exact}.
The strategy of the method is to search for an improved solution,
$\boldsymbol{\theta}^{(new)}=\big(\omega_{k}^{(new)},\boldsymbol{\mu}_{k}^{(new)},\boldsymbol{\Sigma}_{k}^{(new)}\big)$,
which is along the line joining the current and previous iterates,
before updating the Gaussian parameters in the M-step. In particular,
we want to find $\boldsymbol{\rho}^{(it)}=(\rho_{\omega_{k}},\rho)$,
where $\rho_{\omega_{k}}$ is the step size for each Gaussian weight
$\omega_{k}$ and $\rho$ is the common step size for all Gaussian
means and covariance matrices in GMM \cite{xiang2020exact}, such
that

\begin{equation}
\begin{aligned}\omega_{k}^{(new)} & =\omega_{k}^{(it-1)}+\rho_{\omega_{k}}(\omega_{k}^{(it)}-\omega_{k}^{(it-1)})\,,\\
\boldsymbol{\mu}_{k}^{(new)} & =\boldsymbol{\mu}_{k}^{(it-1)}+\rho\,(\boldsymbol{\mu}_{k}^{(it)}-\boldsymbol{\mu}_{k}^{(it-1)})\,,\\
\boldsymbol{\Sigma}_{k}^{(new)} & =\boldsymbol{\Sigma}_{k}^{(it-1)}+\rho\,(\boldsymbol{\Sigma}_{k}^{(it)}-\boldsymbol{\Sigma}_{k}^{(it-1)})\,,
\end{aligned}
\label{eq:els-solution}
\end{equation}
maximizes the log-likelihood. Here, the superscripts indicate solutions
at current $(it)\mbox{th}$ and previous $(it-1)\mbox{th}$ iteration,
respectively. Details for the computation of the step sizes $\boldsymbol{\rho}^{(it)}$
can be found in Ref. \cite{xiang2020exact}. If $\mathcal{L}(\boldsymbol{\theta}^{(new)})>\mathcal{L}(\boldsymbol{\theta}^{(it)}),$
we use $\boldsymbol{\theta}^{(new)}$ to update the Gaussian parameters
in M-step instead of using $\boldsymbol{\theta}^{(it)}.$ The sketch
for one iteration of ELS-EM \cite{xiang2020exact} is as follows:
\begin{enumerate}
\item E-step: Evaluate the responsibilities $r_{jk}^{(it)}\big(\boldsymbol{\theta}^{(it)}\big)$
using (\ref{eq:responsibilities}).
\item ELS-step: Compute the solution $\boldsymbol{\theta}^{(new)}$ using
(\ref{eq:els-solution}). If $\mathcal{L}(\boldsymbol{\theta}^{(new)})>\mathcal{L}(\boldsymbol{\theta}^{(it)}),$
evaluate $r_{jk}^{(new)}\big(\boldsymbol{\theta}^{(new)}\big)$ and
set $r_{jk}^{(it)}=r_{jk}^{(new)}$, else, keep $r_{jk}^{(it)}$
from the E-step.
\item M-step: Update Gaussian parameters $\boldsymbol{\theta}^{(it+1)}$
using $r_{jk}^{(it)}$.
\end{enumerate}
It was pointed out in Ref. \cite{xiang2020exact} that the cost of
the ELS-step is the same as the cost of the E-step, since it needs
an additional evaluation of the log-likelihood function and the re-evaluation
of responsibilities $r_{jk}$ when the new solution $\boldsymbol{\theta}^{(new)}$
is better. This makes one iteration of ELS-EM about two times more
expensive than one EM iteration. We have implemented ELS for standard
(non-adaptive) EM-GMM, and we have confirmed that this is the case
(as we will show). Moreover, we have found a reduction in iteration
count of only $\sim2-2.5\times$ with ELS-EM for our synthetic data
sets, resulting in almost no wall-clock-time speed up. This is in
contrast with our proposed algorithm, described in the next section,
where we demonstrate wall-clock-time speed-ups up to $\sim$$60\times$
for the same data sets.

\subsection{Anderson acceleration of EM\label{sec:AAEM-applications}}

AA is a common accelerator for nonlinear Picard iterative procedures
\cite{walker2011anderson,carlson1998fpa,anderson1965}. EM-GMM is
indeed a Picard iteration, which can show slow convergence especially
in the case where the Gaussians in the mixture are highly overlapping.
Therefore, in principle, it can be accelerated by Anderson acceleration.
A direct application of AA for EM (AAEM), taken from \cite{walker2011anderson},
is given in Algorithm \ref{alg:AA-EM}. In the algorithm, $\boldsymbol{G}(\boldsymbol{\theta})$
is one step of either standard or adaptive EM-GMM.
\begin{algorithm}[th]
\caption{Anderson accelerated EM algorithm (AAEM)\label{alg:AA-EM}}

Given initial solution, $\boldsymbol{\theta}^{(0)}$ and maximum number
of residuals, $m_{AA}\geqslant1$.

Evaluate $\boldsymbol{\theta}^{(1)}=\boldsymbol{G}(\boldsymbol{\theta}^{(0)}).$

Do $it=1,2,...$ until converged:
\begin{enumerate}
\item Set the number of residuals in AA, $m=min(it,m_{AA})$.
\item Set $\boldsymbol{F}_{it}=(\boldsymbol{f}_{it-m},...,\boldsymbol{f}_{it})$
where $\boldsymbol{f}_{i}=\boldsymbol{G}(\boldsymbol{\theta}^{(i)})-\boldsymbol{\theta}^{(i)}$.
\item Solve for $\boldsymbol{\alpha}^{(it)}$ such that 
\begin{equation}
\boldsymbol{\alpha}^{(it)}=\mbox{argmin}_{\boldsymbol{\alpha}}\;\vert\vert\boldsymbol{F}_{it}\boldsymbol{\alpha}\vert\vert\;\mbox{subject to}\;\ensuremath{\sum_{i=0}^{m}\alpha_{i}=1}.\label{eq:AA-constrainedLSP}
\end{equation}
\item $\boldsymbol{\theta}^{(it+1)}=\sum_{i=0}^{m}\alpha_{i}^{(it)}\boldsymbol{G}(\boldsymbol{\theta}^{(it-m+i)}).$
\item Check for convergence by monitoring log-likelihood (use (\ref{eq:llh-gmm})
for standard EM or (\ref{eq:pllh-gmm}) for adaptive EM).
\end{enumerate}
\end{algorithm}

Unfortunately, several problems arise when one naively applies AA
to EM-GMM. Firstly, for both standard and adaptive cases, our numerical
experiments show that AAEM only conserves the zeroth moment of the
given data set. Secondly, the positive-definiteness property of
the Gaussian weights and covariance matrices is not ensured with standard
AA since the Anderson iterate is expressed as a non-convex linear
combination of positive-definite solutions \cite{walker2017,chen2019convergence}
(i.e., some of the AA coefficients $\alpha_{i}$ in (\ref{eq:AA-constrainedLSP})
can be negative). Thirdly, the monotonicity of the log-likelihood
function is not ensured. From our numerical results, we often see
that the log-likelihood of the Anderson solution is less than the
log-likelihood of the current EM iterate, i.e., $\mathcal{L}(\boldsymbol{\theta}^{(AA)})<\mathcal{L}(\boldsymbol{\theta}^{(it)})$
in the standard case, or $\mathcal{PL}(\boldsymbol{\theta}^{(AA)})<\mathcal{PL}(\boldsymbol{\theta}^{(it)})$
in the adaptive case. Fourthly, in adaptive EM, since the exact number
of Gaussian components is unknown, we often start with a larger number
of components than needed for a given data set. In this case, when
applying AA for adaptive EM-GMM, we observe that the method does not
produce the right number of Gaussian components and converges to a
non-optimal solution. In some situations, the convergence rate of
adaptive AAEM is observed to be slower than that of the adaptive non-accelerated
EM, as we will demonstrate in a later section.

We have explored various solutions proposed in the literature to address
these issues, with varied success. To preserve up to second moments
at every iteration, we considered accelerating Gaussian moments $\mathcal{\boldsymbol{M}}_{k}=\omega_{k}(1,\boldsymbol{\mu}_{k},\boldsymbol{\Sigma}_{k}+\boldsymbol{\mu}_{k}\boldsymbol{\mu}_{k}^{T})$
instead of the Gaussian parameters $\boldsymbol{\theta}_{k}=(\omega_{k},\boldsymbol{\mu}_{k},\boldsymbol{\Sigma}_{k})$.
We find that this approach conserves up to second moments in the non-adaptive
case with fixed number of Gaussian components, but not in the adaptive
case because of the renormalization of the Gaussian weights in the
M-step. In addition, the moment-acceleration strategy may break the
positive-definiteness of $\boldsymbol{\Sigma}_{k}$ when computed
from the second moment matrix $\boldsymbol{\mathcal{M}}_{k,2}$.
To address the positive-definiteness of Gaussian weights and covariance
matrices, we employed the AA globalization technique proposed in \cite{chen2019convergence},
i.e., an additional constraint for the positivity of the AA coefficients
is added to the least square problem (\ref{eq:AA-constrainedLSP})
as follows

\begin{equation}
\begin{aligned}\mbox{Find}\;\boldsymbol{\alpha}^{(it)}\;\mbox{s.t.}\;\boldsymbol{\alpha}^{(it)}=\mbox{argmin}_{\boldsymbol{\alpha}}\,\vert\vert\boldsymbol{F}_{it}\boldsymbol{\alpha}\vert\vert\\
\mbox{subject to}\,\sum_{i=0}^{m}\alpha_{i}=1\,\mbox{and}\,\alpha_{i}>0\;\;\forall\,i,
\end{aligned}
\label{eq:AA-constrainedLSP2}
\end{equation}
where $m$ is the number of past residuals in AA at the $(it)\mbox{th}$
iteration. We found from our numerical experiments that this approach
significantly slows down the local convergence speed of AA (since
it restricts the optimization domain), and that, at a later phase
in the AAEM iteration, it frequently defaults back to EM, i.e., it
returns $\alpha_{m}=1$ and $\alpha_{0}=\cdots=\alpha_{m-1}=0$.

To address successfully the AAEM problem of local convergence to a
non-optimal solution, we follow the AA regularization approach proposed
by Henderson et al. \cite{henderson2019daarem}. Specifically, the
constrained least-squares problem (\ref{eq:AA-constrainedLSP}) can
be reformulated as an unconstrained least-squares problem to which
a regularization term $\lambda\boldsymbol{I}_{m\times m}$ is added
as follows (see \cite{walker2011anderson,henderson2019daarem}):
\begin{equation}
\mbox{At }(it)\mbox{th}\mbox{ iteration, find}\;\boldsymbol{\gamma}^{(it)}\;\mbox{s.t. \ensuremath{\big(}\ensuremath{\ensuremath{\boldsymbol{\mathcal{F}}_{it}^{T}\boldsymbol{\mathcal{F}}_{it}}+\ensuremath{\lambda\boldsymbol{I}}}\big)}\boldsymbol{\gamma}^{(it)}=\mathcal{\boldsymbol{F}}_{it}^{T}\boldsymbol{f}_{it}\,,\label{eq:AA-unconstrained_regLSP}
\end{equation}
where $\boldsymbol{\mathcal{F}}_{it}=(\Delta\boldsymbol{f}_{it-m},\cdots,\Delta\boldsymbol{f}_{it-1})$,
$\Delta\boldsymbol{f}_{i}=\boldsymbol{f}_{i+1}-\boldsymbol{f}_{i}$
and $\boldsymbol{f}_{i}=\boldsymbol{G}(\boldsymbol{\theta}^{(i)})-\boldsymbol{\theta}^{(i)}$.
As remarked in Ref. \cite{walker2011anderson}, there exists a one-to-one
correspondence between the coefficients $\boldsymbol{\alpha}^{(it)}$
given by (\ref{eq:AA-constrainedLSP}) and the coefficients $\boldsymbol{\gamma}^{(it)}$
given by (\ref{eq:AA-unconstrained_regLSP}) when $\lambda=0$, that
is:
\[
\begin{aligned}\alpha_{0}^{(it)} & =\gamma_{0}^{(it)}\,,\,\alpha_{i}^{(it)}=\gamma_{i}^{(it)}-\gamma_{i-1}^{(it)}\;\mbox{for}\,i=1,\cdots,m-1\,,\\
\alpha_{m}^{(it)} & =1-\gamma_{m-1}^{(it)}\,.
\end{aligned}
\]
 In this case, at the $(it)\mbox{th}$ iteration, the updated Anderson
iterate can be written as
\begin{equation}
\boldsymbol{\theta}^{(it+1)}=\boldsymbol{G}(\boldsymbol{\theta}^{(it)})-\sum_{i=0}^{m-1}\gamma_{i}^{(it)}\big[\boldsymbol{G}(\boldsymbol{\theta}^{(it-m+i+1)})-\boldsymbol{G}(\boldsymbol{\theta}^{(it-m+i)})\big].\label{eq:AAiterate}
\end{equation}
According to Henderson et al. \cite{henderson2019daarem}, the use
of the regularization term $\lambda\boldsymbol{I}$ helps combine
the convergence robustness of EM and the local convergence speed of
AA. One can see that if $\lambda=0$, we recover AA, and if $\lambda\gg1$,
we recover EM. We refer to Ref. \cite{henderson2019daarem} for the
strategy of computing $\lambda$ at each EM iteration, which we follow
strictly. This approach, together with a novel, very efficient monotonicity-control
implementation for the log-likelihood function (discussed in the next
section) results in an algorithm that captures the right number of
components and quickly converges to the right solution nearly without
run-time penalty.

\section{The adaptive accelerated-monotonicity-preserving Expectation-Maximization
(A-AMEM) algorithm\label{sec:AMEM}}

The main goal of this paper is to make the adaptive EM (A-EM) algorithm
faster. To this end, we apply an Anderson acceleration to A-EM while
maintaining key features of both adaptive and standard EM such as
component adaptivity, conservation of up to second moments, preservation
of positive-definiteness of Gaussian weights and covariance matrices,
and monotonicity preservation of the log-likelihood function during
the iteration. The adaptive, accelerated, monotonicity-preserving
EM (A-AMEM) algorithm for GMM is outlined in Section \ref{sec:main-alg},
with the initialization and implementation details discussed in Sections
\ref{subsec:initialization-details} and \ref{subsec:Implementation-details},
respectively.

\subsection{The main A-AMEM algorithm\label{sec:main-alg}}

The accelerated algorithm for the adaptive EM-GMM iteration is detailed
in Algorithm \ref{alg:AMEM-alg}. We implement the regularization
term \cite{henderson2019daarem} and employ the AA periodical restart
\cite{henderson2019daarem,carlson1998fpa,Pratapa2015restartpulay}
to address some of the pitfalls of a naive AAEM implementation for
adaptive GMM. 
\begin{algorithm}[t]
\caption{Adaptive accelerated-monotonicity EM (A-AMEM) algorithm\label{alg:AMEM-alg}}

Given
\begin{itemize}
\item $m_{AA}$, the maximum number of residuals for AA.
\item $K_{init},$ the initial number of components to be used.
\end{itemize}
\textbf{Initialization:} Perform K-means clustering algorithm with
$K_{init}$ clusters (multiple times and select the best run) to obtain
the initial solutions $\boldsymbol{\theta}^{(0)}$.

\textbf{The adaptive AMEM:} Do $it=0,1,2,...$ until converged:
\begin{enumerate}
\item Perform \textbf{EM} step: $\boldsymbol{\theta}^{(it+1)}=\boldsymbol{G}(\boldsymbol{\theta}^{(it)})$.
During this step, Gaussian(s) may be killed.

Restart AA if a Gaussian component is eliminated.
\item Apply \textbf{regularized AA}:

Solve the least square problem (\ref{eq:AA-unconstrained_regLSP})
for $\boldsymbol{\gamma}^{(it)}$.

Compute the Anderson iterate, $\boldsymbol{\theta}^{(AA)}$, using
(\ref{eq:AAiterate}). If  any Gaussian weight becomes negative, then
use EM solution and continue to the next iteration.

Else, go to Step 3.
\item \textbf{Monotonicity control}:

If $\mathcal{PL}(\boldsymbol{\theta}^{(AA)})-\mathcal{PL}(\boldsymbol{\theta}^{(it)})>-\epsilon$
then set $\boldsymbol{\theta}^{(it+1)}=\boldsymbol{\theta}^{(AA)},$

Else, use EM solution.
\item \textbf{Restart AA} if number of residual vectors reaches $m_{AA}.$
\item Check for convergence by monitoring the penalized log-likelihood given
by (\ref{eq:pllh-gmm}).
\end{enumerate}
\textbf{Conservation of moments:} Perform one final standard EM step
after convergence.
\end{algorithm}
In Algorithm \ref{alg:AMEM-alg}, $\boldsymbol{G}(\boldsymbol{\theta})$
represents the fixed-point EM step of A-EM, $\lambda$ is the regularization
factor and $\epsilon$ is the log-likelihood controlling parameter.
Details on Algorithm \ref{alg:AMEM-alg} and further discussions for
the values of $\lambda$ and $\epsilon$ are given next.

\subsection{Initialization of A-AMEM\label{subsec:initialization-details}}

\textcolor{black}{We use K-means clustering for Gaussian initialization.
The initial centroids of the clusters in each K-means call are selected
from the data points with an improved seeding technique \cite{Arthur07k-means++}.
To obtain the best guess for the number of components, we employ the
gap statistic (GS) method \cite{Tibshirani2001gapstatistics}, using
the so-called the gap statistic value (GSV). (We have also explored
other K-means techniques \cite{raykov2016adaptiveKmeans,Darken1990fastadaptkmeans,Rousseeuw1987silhouette,Nanjundan2019silhouette}
and we will discuss them in Section \ref{subsec:Adaptive-multi-KMeans}.)
The GSV associated to $K$ clusters  is defined as (cf. \cite{Tibshirani2001gapstatistics}):}

\textcolor{black}{
\begin{equation}
GSV(K)=E^{*}\big[\mbox{ln}(SSE_{k})\big]-\mbox{ln}(SSE_{k})\,.\label{eq:GSV(K)-Kmeans}
\end{equation}
}In (\ref{eq:GSV(K)-Kmeans}), $E^{*}$ denotes the expectation under
a sample from the reference distribution, and $SSE_{k}$ is given
as:\textcolor{black}{
\begin{equation}
SSE_{K}=\sum_{k=1}^{K}\sum_{\boldsymbol{x}_{i}\,\in\,\mathcal{C}_{k}}\zeta_{i}\vert\boldsymbol{x}_{i}-\boldsymbol{c}_{k}\vert^{2}\,,\label{eq:KM-inertia}
\end{equation}
}where \textcolor{black}{$\mathcal{C}_{k}$ is the $k\mbox{th}$ cluster
and $\zeta_{i}$ is the weight for point $\boldsymbol{x}_{i}\in\mathcal{C}_{k}$.}

\textcolor{black}{The computation of the expectation in $GSV(K)$
is done with Monte-Carlo from reference data sets (see Ref. \cite{Tibshirani2001gapstatistics}).
We have found that generating the reference sets from a uniform distribution
over a box aligned with the principal components of the observed sample
\cite{Tibshirani2001gapstatistics} (which is the one we use in our
numerical simulations in Section \ref{subsec:Adaptive-multi-KMeans}),
or from a unit normal distribution with parameters taken to be the
sample's mean and covariance matrix yields reliable results for our
data sets.}

\textcolor{black}{The optimal number of clusters $K_{opt}$ is found
from the following condition:}
\begin{equation}
K_{opt}=\mbox{smallest \emph{K} s.t. }GSV(K)>GSV(K+1)+\tau\times s_{K+1}\label{eq:GS-method-criterion-choosing-Kopt}
\end{equation}
\textcolor{black}{for $K=K_{min},\cdots,K_{max}$. In (\ref{eq:GS-method-criterion-choosing-Kopt}),
$s_{K}$ is the standard deviation term which accounts for the Monte
Carlo simulation error in evaluating $GSV(K)$, and $\tau$ is user-input
factor that represents the amount of standard deviation used. We refer
the reader to Ref. \cite{Tibshirani2001gapstatistics} for more details
on the evaluation of $GSV(K)$ and $s_{K}$. In our application, choosing
$\tau=0\;\mbox{or}\;1$ in (\ref{eq:GS-method-criterion-choosing-Kopt})
instead of $\tau=-1$ as in Ref. \cite{Tibshirani2001gapstatistics}
helps avoid possible under-estimation of number of clusters by the
GS method. Although it may potentially over-estimate the number of
clusters by a few in some cases, this is acceptable in our application
since, for robustness, A-AMEM should begin with more components than
the expected number. Once the optimal number of clusters is obtained,
we set the initial number of clusters as $K=K_{opt}+K_{adjust}$ for
some $K_{adjust}>0$, to further avoid under-estimation of the model.
We then repeat K-means with $K$ clusters multiple times and the centroids
associated with the best trial are selected. The best trial is the
one that yields the smallest inertia (SSE) value. The initial Gaussian
means $\boldsymbol{\mu}_{k}^{(0)},\,k=1,\cdots,K$ are assigned from
the clusters' centroids of the best run. The initial Gaussians weights
are computed from the K-means best run as:}

\textcolor{black}{
\[
\omega_{k}^{(0)}=\frac{n_{k}}{N}
\]
where $n_{k}=\sum_{i\in\mathcal{C}_{k}}\zeta_{i}$ is the total weight
of points that belong to the $k\mbox{th}$ cluster and $N$ is the
total number of observed points in the data set. The initial Gaussians'
covariance matrices can be assigned to the clusters' so-called within-covariances,
which are evaluated as:}

\textcolor{black}{
\[
\boldsymbol{\Sigma}_{k}^{(0)}=\frac{1}{n_{k}}\sum_{j=1}^{n_{k}}\zeta_{i}(\boldsymbol{x}_{j}-\boldsymbol{\mu}_{k}^{(0)})(\boldsymbol{x}_{j}-\boldsymbol{\mu}_{k}^{(0)})^{T}\:,
\]
where $\boldsymbol{x}_{j}\,,j=1,\cdots,n_{k}$ are the points assigned
to the $k\mbox{th}$ cluster at the end of the K-means iteration.
In general, the K-means initialization algorithm for EM is quite inexpensive
compared to EM, taking a small fraction (5-10\%) of the total wall-clock
time, and can have a large impact in the efficiency of the overall
algorithm.}

\subsection{Implementation details of A-AMEM\label{subsec:Implementation-details}}

For the adaptive EM algorithms, we follow the steps outlined in Section
\ref{subsec:The-adaptive-EM} to compute the updated Gaussians' parameters.
We remark that the $k\mbox{th}$ Gaussian is removed from the mixture
if the updated weight is negative. Additionally, we perform a restart
of AA if a Gaussian component is killed during this step, since the
past AA solution history of the removed component is no longer valid.

Once we obtain the updated values for the Gaussian parameters, we
solve the unconstrained regularized least-square problem (\ref{eq:AA-unconstrained_regLSP}),
and compute the updated AA solution, $\boldsymbol{\theta}^{(AA)}$,
using (\ref{eq:AAiterate}). To ensure positive-definiteness of covariance
matrices, we accelerate the entries of the matrices $\boldsymbol{L}_{k},\,k=1,\ldots,K$
where $\boldsymbol{\Sigma}_{k}=\boldsymbol{L}_{k}\boldsymbol{L}_{k}^{T}$
is the Cholesky decomposition \cite{horn2012matrix} of $\boldsymbol{\Sigma}_{k}$
for $k=1,\ldots,K$. After the acceleration procedure, the covariance
matrices $\boldsymbol{\Sigma}_{k}$ can be recovered using the lower-triangular
matrices $\boldsymbol{L}_{k}$ for $k=1,\cdots,K$. As for the Gaussian
weights, if AA returns $\omega_{l}<0$ for some $l=1,\ldots,K$, then
we simply roll back to the EM solution, which is guaranteed to keep
Gaussian weights positive and increase the penalized log-likelihood
function. We note that the violation of positive-definiteness of the
Gaussian weights does not occur frequently during the A-AMEM iteration,
about $1\%-5\%$ of the time for our synthetic data sets. Hence, the
local convergence speed of AA is not much affected by the development
of negative weights, and thus the rollback-to-EM strategy seems reasonable.

The monotonicity control step (step 3 in Algorithm \ref{alg:AMEM-alg})
is expensive because in principle it requires an additional evaluation
of the log-likelihood function (\ref{eq:pllh-gmm}), which involves
loops over all samples and available Gaussian components and expensive
logarithmic and exponential evaluations. As a result, one iteration
of A-AMEM becomes twice as expensive as one A-EM iteration. To avoid
evaluating the penalized log-likelihood function for the AA iterate,
$\mathcal{PL}(\boldsymbol{\theta}^{(AA)})$, in the monotonicity control
step, we approximate the computation of $\mathcal{PL}(\boldsymbol{\theta}^{(AA)})-\mathcal{PL}(\boldsymbol{\theta}^{(it)})$
using a first-order Taylor expansion, which relies on the exact evaluation
of score functions. We recall that the score function is the derivative
of the log-likelihood function with respect to the Gaussian unknowns.
In particular, instead of checking

\begin{equation}
\mathcal{PL}(\boldsymbol{\theta}^{(AA)})-\mathcal{PL}(\boldsymbol{\theta}^{(it)})>-\epsilon\,,\label{eq:MC-original}
\end{equation}
we check

\begin{equation}
\frac{\partial\mathcal{PL}(\boldsymbol{\theta})}{\partial\boldsymbol{\theta}^{(it)}}\cdot\big(\boldsymbol{\theta}^{(AA)}-\boldsymbol{\theta}^{(it)}\big)>-\epsilon\,,\label{eq:MC-Taylor-expansion}
\end{equation}
where $\epsilon>0$ is the monotonicity parameter, and

\begin{equation}
\frac{\partial\mathcal{PL}(\boldsymbol{\theta})}{\partial\omega_{k}^{(it)}}=\frac{N_{k}^{(it)}}{\omega_{k}^{(it)}}-\frac{T}{2\omega_{k}^{(it)}}-N+\frac{TK}{2}\,,\label{eq:dPL_dwk}
\end{equation}
\begin{equation}
\frac{\partial\mathcal{PL}(\boldsymbol{\theta})}{\partial\boldsymbol{\mu}_{k}^{(it)}}=\big(\boldsymbol{\Sigma}_{k}^{(it)}\big)^{-1}\bigg[\sum_{j=1}^{N}r_{jk}^{(it)}(\boldsymbol{x}_{j}-\boldsymbol{\mu}_{k}^{(it)})\bigg]\,,\label{eq:dPL_dmuk}
\end{equation}
\begin{equation}
\begin{aligned}\frac{\partial\mathcal{PL}(\boldsymbol{\theta})}{\partial\boldsymbol{\Sigma}_{k}^{(it)}} & =\big(\boldsymbol{\Sigma}_{k}^{(it)}\big)^{-1}\bigg\{\sum_{j=1}^{N}\frac{r_{jk}^{(it)}}{2}\bigg[-\boldsymbol{\Sigma}_{k}^{(it)}+\\
 & \big(\boldsymbol{x}_{j}-\boldsymbol{\mu}_{k}^{(it+1)}\big)\big(\boldsymbol{x}_{j}-\boldsymbol{\mu}_{k}^{(it+1)}\big)^{T}\bigg]\bigg\}\big(\boldsymbol{\Sigma}_{k}^{(it)}\big)^{-1}\,,
\end{aligned}
\label{eq:dPL-dsigmak}
\end{equation}
for $k=1,\cdots,K$. Details on the derivation of (\ref{eq:dPL_dwk}),
(\ref{eq:dPL_dmuk}) and (\ref{eq:dPL-dsigmak}) are given in Appendix \ref{sec:Derivative-LLH-appendix}. Using (\ref{eq:MC-Taylor-expansion})
is cheap because most quantities in (\ref{eq:dPL_dwk}), (\ref{eq:dPL_dmuk})
and (\ref{eq:dPL-dsigmak}) can be re-used from the adaptive M step
in Algorithm \ref{alg:AMEM-alg}. Thus, the evaluation complexity
for the gradient $\frac{\mathcal{\partial PL}(\boldsymbol{\theta})}{\partial\boldsymbol{\theta}^{(it)}}$
is only of order $O(KD)$ where $D$ is the dimension of $\boldsymbol{\mu}_{k}^{(it)}.$
This is much more efficient than the direct evaluation of $\mathcal{PL}(\boldsymbol{\theta}^{(AA)})$,
which has computational complexity of $O(NKD)$, where $N\gg1$ is
the number of sample data points. The Taylor expansion approach for
approximating $\mathcal{PL}(\boldsymbol{\theta}^{(AA)})-\mathcal{PL}(\boldsymbol{\theta}^{(it)})$
renders the cost of one A-AMEM iteration comparable to one A-EM iteration,
and is a key contributor to the efficiency improvement of our implementation.
As for the monotonicity parameter $\epsilon$, we find that using
values of $\epsilon\in[0.001,\;0.01]$ works well for our simulations.
Choosing $\epsilon=0.01$ means that likelihood ratios between current
and accelerated solutions are allowed to be no greater than $e^{\epsilon}\approx1.01$
\cite{henderson2019daarem}. Further discussion about the choice of
$\epsilon$ can be found in the same reference.

We apply a periodic restart strategy for AA in A-AMEM to help improve
the overall robustness of the algorithm, as suggested i\textcolor{black}{n
\cite{henderson2019daarem,Pratapa2015restartpulay}}. The AA restart
solution proposed in \cite{Pratapa2015restartpulay} kept the last
column of $\boldsymbol{\mathcal{F}}_{it}$. We have tested both resetting
all AA residuals to zero and keeping the latest column of $\boldsymbol{\mathcal{F}}_{it}$,
and found that they yield comparable performance for our numerical
tests. 

Finally, once the algorithm converges to a solution with an optimal
number of Gaussian components, we perform one final standard EM iteration
(see Section \ref{subsec:The-standard-EM} and \cite{chen2020CR-EM-GM})
to recover the conservation up to second moments of the observed data
set.

\section{Numerical Results\label{sec:Numerical-Results}}

We apply A-AMEM (Algorithm \ref{alg:AMEM-alg}) and A-EM to several
synthetic GMM data sets and compare the performance and results of
the two algorithms. Each synthetic data set is a convex linear combination
of $K_{exact}=3$ Gaussian distributions with different overlap. Each
set consists of $N=1000$ points. The separation between Gaussians
can be measured by the Euclidean distances of the Gaussians' means
and the Gaussians' shapes, determined by covariance matrices. We generate
three synthetic data sets, namely Very Well Separated (VWS), Poorly
Separated (PS) and Very Poorly Separated (VPS), with Gaussian means
given as follows:
\[
\mbox{VWS:}\;\boldsymbol{\mu}_{1}=\left[\begin{array}{c}
-3\\
-3\\
-3
\end{array}\right],\;\boldsymbol{\mu}_{2}=\left[\begin{array}{c}
0\\
0\\
0
\end{array}\right],\;\boldsymbol{\mu}_{3}=\left[\begin{array}{c}
3\\
3\\
3
\end{array}\right],
\]
\[
\mbox{PS:}\;\boldsymbol{\mu}_{1}=\left[\begin{array}{c}
-2\\
-2\\
-2
\end{array}\right],\;\boldsymbol{\mu}_{2}=\left[\begin{array}{c}
0\\
0\\
0
\end{array}\right],\;\boldsymbol{\mu}_{3}=\left[\begin{array}{c}
2\\
2\\
2
\end{array}\right],
\]
\[
\mbox{VPS:}\;\boldsymbol{\mu}_{1}=\left[\begin{array}{c}
-1\\
-1\\
-1
\end{array}\right],\;\boldsymbol{\mu}_{2}=\left[\begin{array}{c}
0\\
0\\
0
\end{array}\right],\;\boldsymbol{\mu}_{3}=\left[\begin{array}{c}
1\\
1\\
1
\end{array}\right].
\]
 For the Gaussian weights and covariance matrices, we choose:
\[
\omega_{1}=0.3,\;\omega_{2}=0.3,\;\omega_{3}=0.4,
\]
\[
\boldsymbol{\Sigma}_{1}=\mbox{diag}(1,1,1),\;\boldsymbol{\Sigma}_{2}=1.5\boldsymbol{\Sigma}_{1},\;\boldsymbol{\Sigma}_{3}=0.75\boldsymbol{\Sigma}_{1},
\]
for the three manufactured GMM data sets. We also apply Algorithm
\ref{alg:AMEM-alg} to real data sets generated from collisionless
plasma particle-in-cell (PIC) simulations \cite{chen2020CR-EM-GM}.
These real data sets consist of electrons' velocity points in the
three dimensional (3D) velocity space, and will be described in detail
later.

Our goal is to study the efficiency of A-AMEM when compared to A-EM
for different initial number of Gaussians components, $K_{init}$.
To this end, we define the iteration reduction factor (IRF) and CPU
time reduction factor (TRF) between accelerated and non-accelerated
EM algorithms as:

\begin{equation}
IRF=\frac{\mbox{number of standard EM iterations}}{\mbox{number of accelerated EM iterations}},\label{eq:IRF}
\end{equation}

\begin{equation}
TRF=\frac{\mbox{standard EM CPU time}}{\mbox{accelerated EM CPU time}}.\label{eq:TRF}
\end{equation}
The same terminating tolerance is used when applying both A-AMEM and
A-EM to the data sets. In order for both A-EM and A-AMEM to kill enough
unnecessary components before converging to the desired optimal solutions,
we use a small terminating tolerance $TOL$. In particular, we set
$TOL=10^{\mathrm{-10}}$ for the synthetic data sets, and $TOL=10^{\mathrm{-12}}$
for the PIC data sets. 

As for the choice of $m_{AA}$, the maximum number of past residuals
in AA, we acknowledge that the performance of AA with respect to this
number is problem-dependent, which was also remarked in \cite{walker2011anderson}.
We find that using $m_{AA}$ between 5 and 10 works well for our simulations.
We set $m_{AA}=5$ for $K_{init}=3$, and $m_{AA}=10$ for $3<K_{init}\leq10$.

\subsection{Visualization of the manufactured GMM data sets\label{subsec:Visualization-data-sets}}

\begin{figure*}[tbh]
\begin{centering}
\includegraphics[width=5cm,height=5cm]{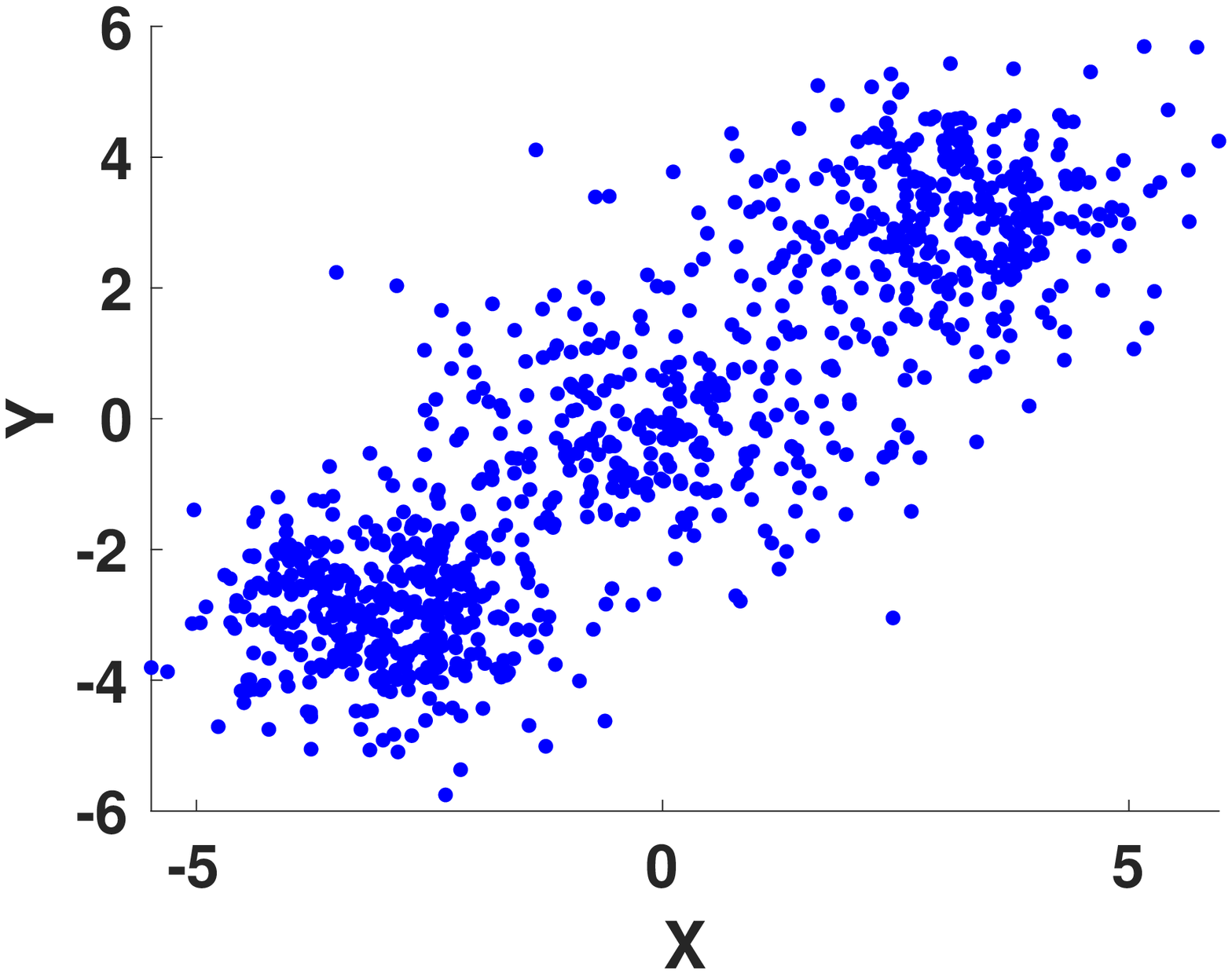}\includegraphics[width=5cm,height=5cm]{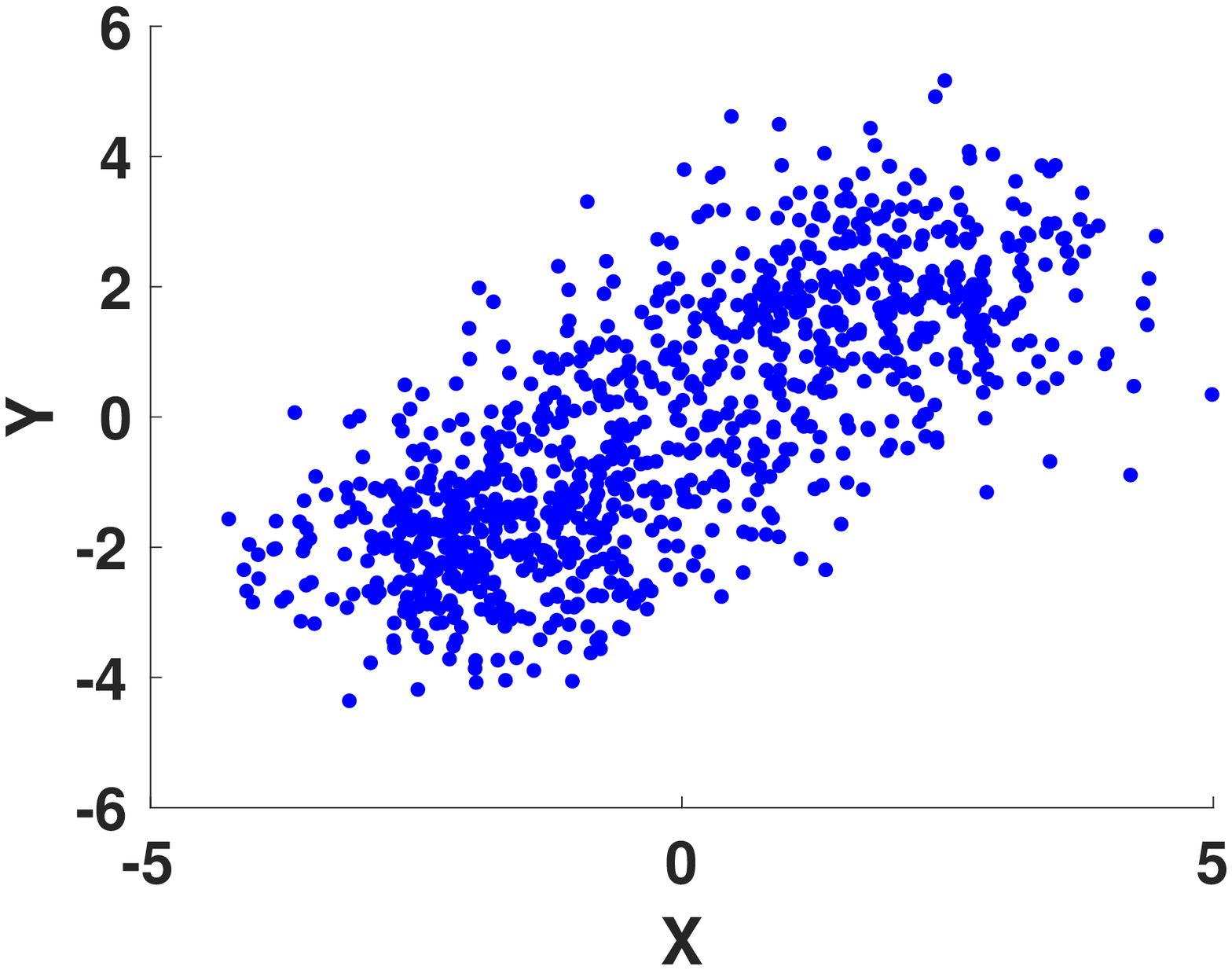}\includegraphics[width=5cm,height=5cm]{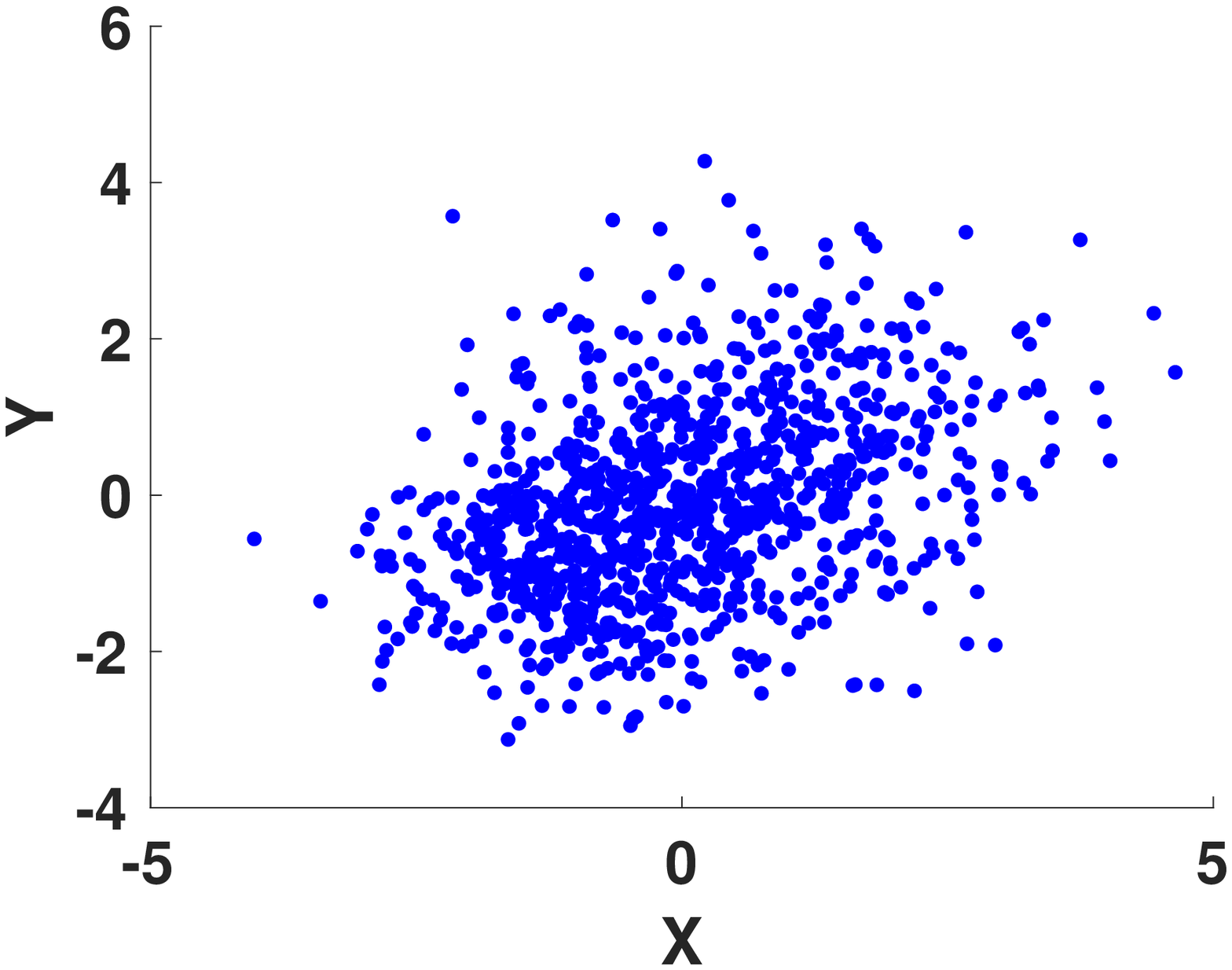}
\par\end{centering}
\caption{Visualization of manufactured GMM data sets in the X-Y plane. (Left)
VWS data set. (Center) PS data set. (Right) VPS data set.\label{fig:Manufactured-data-sets}}
\end{figure*}

The 2D view of the synthetic data sets in the X-Y plane is given in
Fig. \ref{fig:Manufactured-data-sets}. The views in the Y-Z and X-Z
planes are identical to the X-Y plane's view, since we use diagonal
covariance matrices for the three components in the mixture.

We can clearly observe the presence of the three Gaussian components
in the VWS data set and somewhat clearly in the PS data set. However,
they are impossible to tell by the naked eye in the VPS data set.
It is expected that A-EM will be able to detect the right number of
components and converge fairly quickly for the VWS data set. It is
also anticipated that A-EM will detect the right number of Gaussians
for the PS data set but with a slower convergence rate. For the VPS
data set, however, A-EM is expected to converge extremely slowly due
to the significant overlap among Gaussian components in the mixture,
and perhaps even underestimate the number of components.

\subsection{Performance of the non-adaptive ELS-EM\label{subsec:ELS-EM-perf}}

We illustrate the performance of ELS-EM for GMM by applying it to
the synthetic data sets in Section \ref{subsec:Visualization-data-sets}
and comparing the results with standard EM. For these tests, we use
$K_{init}=3$, fixed, i.e., we assume the exact model. We use the
same tolerance for terminating both algorithms. Both ELS-EM and standard
EM are initialized using the best K-means result, i.e., with the smallest
inertia (SSE) values, out of multiple trials with $K=3$ clusters,
as described in Section \ref{subsec:initialization-details}. In these
simulations, we note that ELS-EM converges to the same solutions as
EM. We report the ratios for IRF and TRF between ELS-EM and EM in
Table \ref{tab:RF-ELS-EM-Kinit=00003D3}. We observe from the Table
that the ratios of IRF over TRF are approximately equal to $2.0$.
This verifies that, on average, one iteration of ELS-EM is twice as
expensive as one iteration of EM. It is also apparent that ELS-ES
is only able to speed up the convergence by a factor of $\lesssim2$
for all cases, resulting in no actual CPU speed up (as demonstrated
by the TRF values in Table \ref{tab:RF-ELS-EM-Kinit=00003D3}).

\begin{table}[tbh]
\centering{}\caption{Reduction factors for ELS-EM.\label{tab:RF-ELS-EM-Kinit=00003D3}}
\begin{tabular}{|c|c|c|c|}
\hline 
 & IRF & TRF & IRF/TRF\tabularnewline
\hline 
\hline 
VWS & 1.50 & 0.81 & \textbf{1.85}\tabularnewline
\hline 
PS & 1.90 & 0.89 & \textbf{2.13}\tabularnewline
\hline 
VPS & 1.89 & 0.90 & \textbf{2.10}\tabularnewline
\hline 
\end{tabular}
\end{table}

\subsection{Application of A-AMEM to manufactured data sets with $K_{init}=3$\label{AMEM-Kinit=00003D3}}

\begin{figure*}[tbh]
\begin{centering}
\includegraphics[width=5cm,height=5cm]{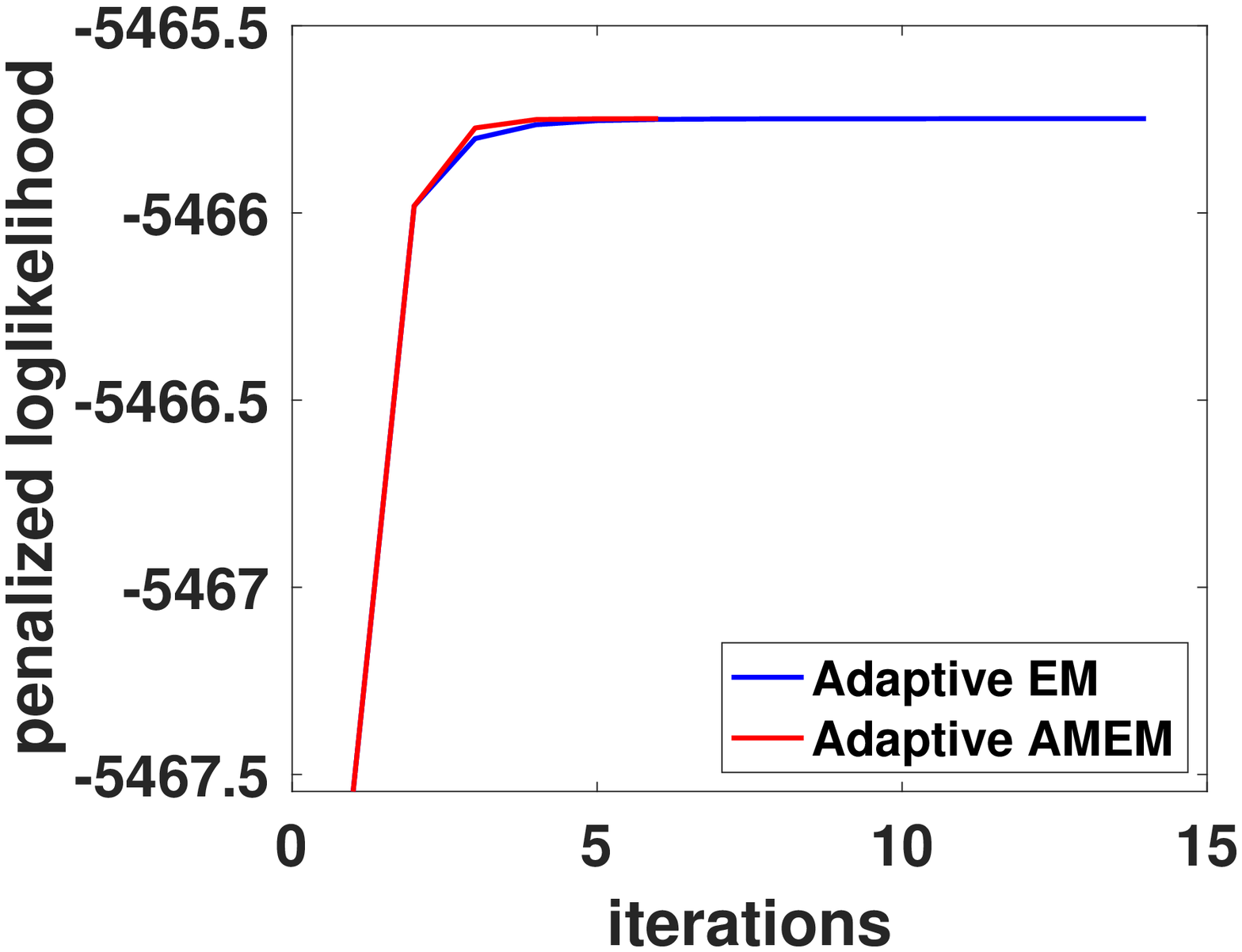}\includegraphics[width=5cm,height=5cm]{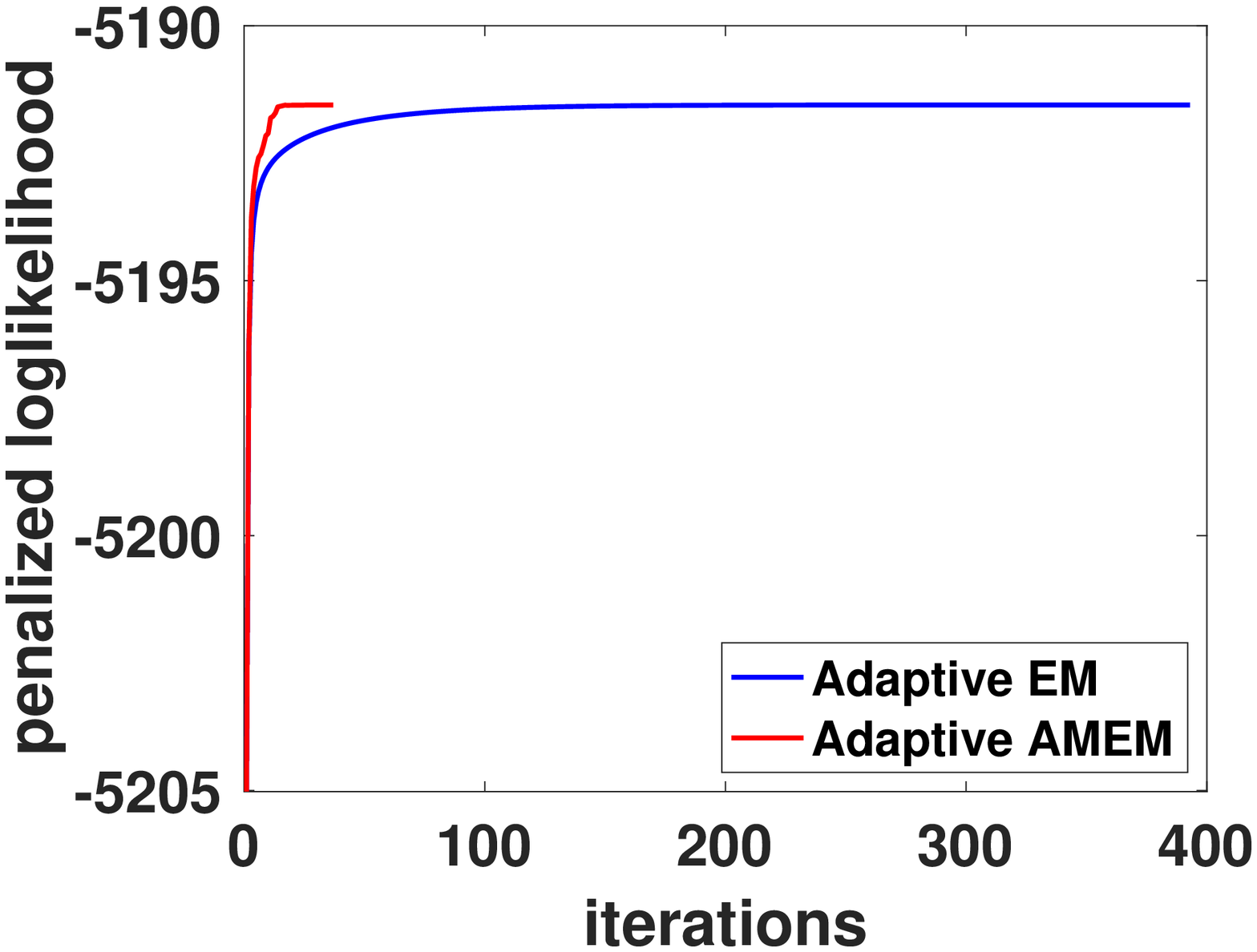}\includegraphics[width=5cm,height=5cm]{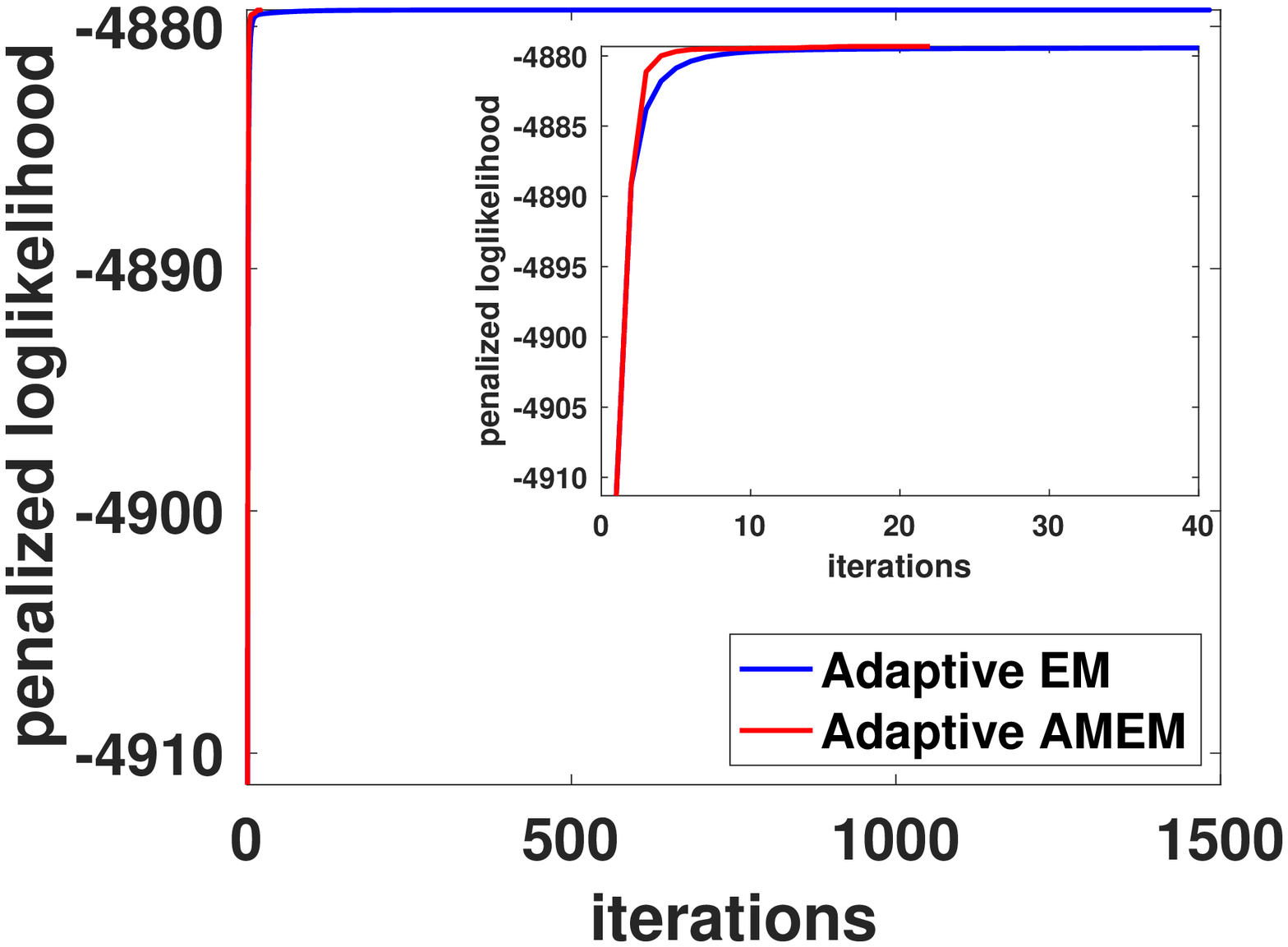}
\par\end{centering}
\caption{Case $K_{init}=3$: History of penalized log-likelihood values as
a function of the number of iterations for synthetic data sets. (Left)
VWS data set. (Center) PS data set. (Right) VPS data set.\label{fig:PLLH-Kinit=00003D3}}
\end{figure*}

We apply A-EM and A-AMEM to the manufactured data sets using the exact
initial number of components $K_{init}=3$. We initialize the Gaussians
as in the previous section. Fig. \ref{fig:PLLH-Kinit=00003D3} depicts
the convergence history of the log-likelihood of the two algorithms.
We have added a subplot inside the convergence plot of the VPS case
to zoom into the first 40 iterations.

As expected, the convergence rate of A-EM worsens with increasing
Gaussian-component overlap. However, A-AMEM seems to converge fairly
quickly, regardless of component overlap, to the same penalized log-likelihood
values as the non-accelerated version. We also remark that both A-EM
and A-AMEM give the correct number of components ($K_{final}=3$)
for the three manufactured data sets, i.e., both algorithms are able
to recognize the true number of components within each data set and
do not kill any component. Table \ref{tab:RF-Kinit=00003D3} records
the iteration and the wall-clock time reduction factors for this case,
and shows that A-AMEM outperforms A-EM dramatically, especially for
the hard VPS case (by a factor of 60 in wall-clock time).

\begin{table}[tbh]
\centering{}\caption{Reduction factors for $K_{init}=3.$\label{tab:RF-Kinit=00003D3}}
\begin{tabular}{|c|c|c|c|}
\hline 
 & IRF & TRF & IRF/TRF\tabularnewline
\hline 
\hline 
VWS & 2.33 & 1.98 & \textbf{1.18}\tabularnewline
\hline 
PS & 10.62 & 10.22 & \textbf{1.04}\tabularnewline
\hline 
VPS & 67.45 & 60.96 & \textbf{1.11}\tabularnewline
\hline 
\end{tabular}
\end{table}

\subsection{A failed application of AA without monotonicity control to A-EM\label{subsec:Failed-AA-application}}

\begin{figure*}[tbh]
\begin{centering}
\includegraphics[width=5cm,height=5cm]{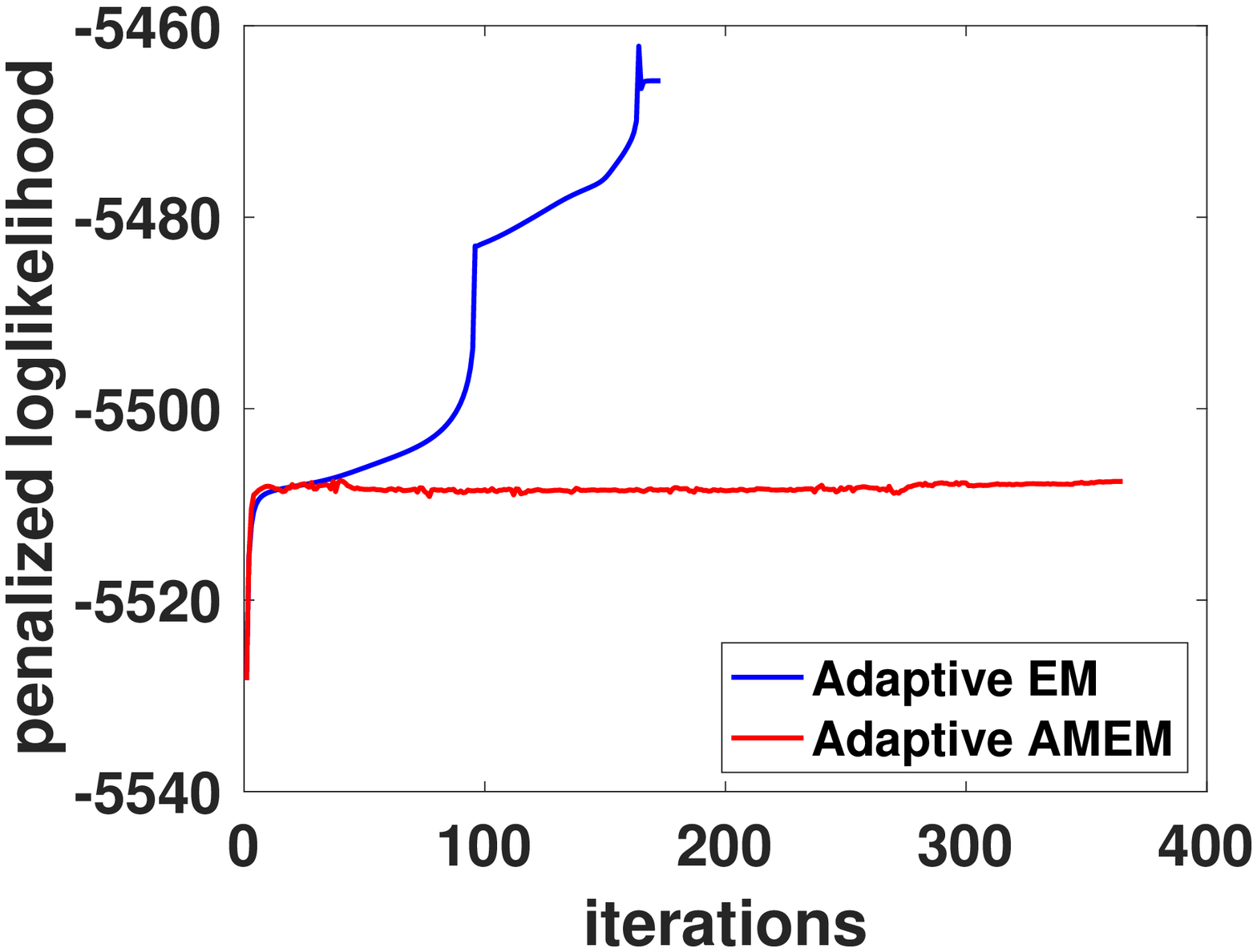}\includegraphics[width=5cm,height=5cm]{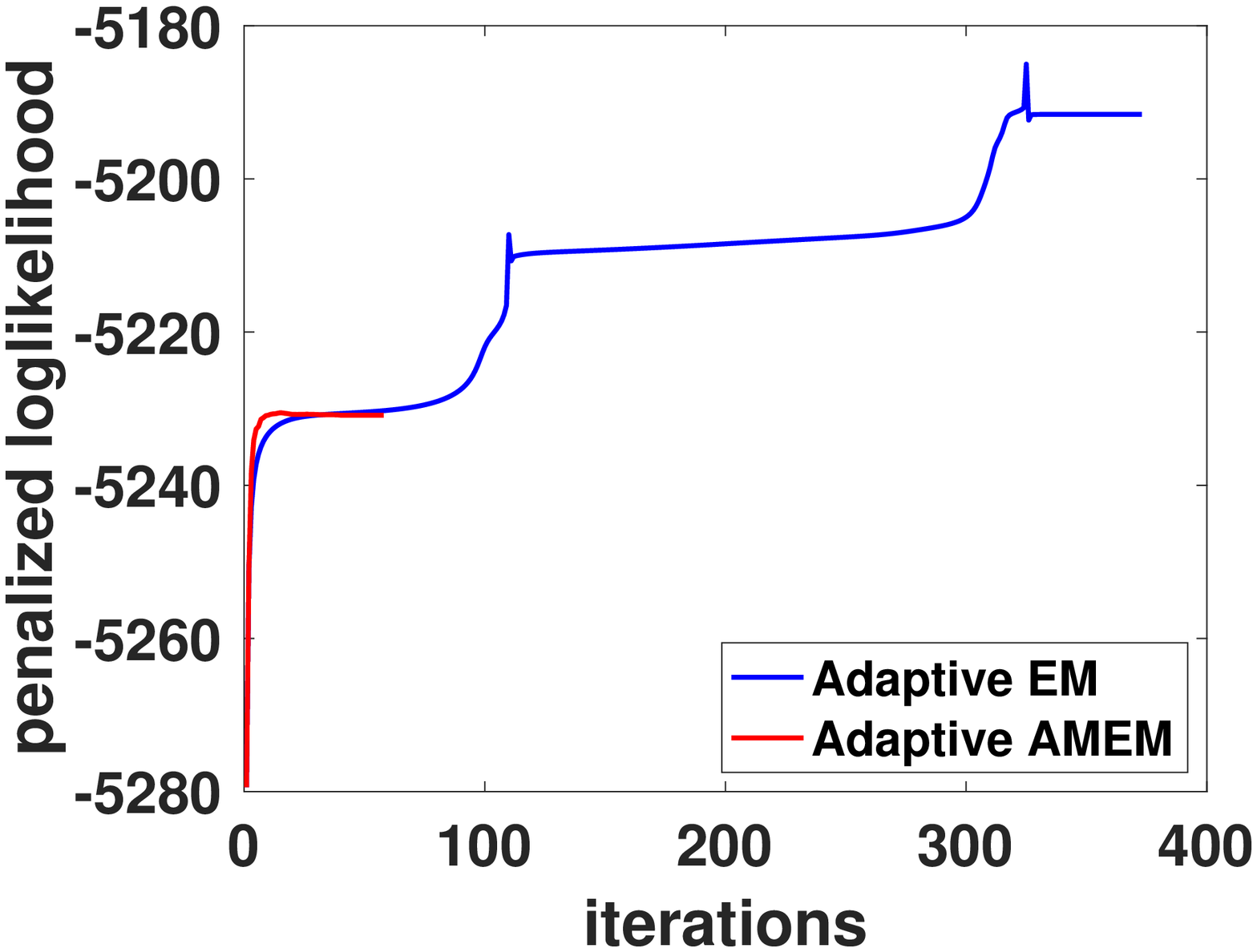}\includegraphics[width=5cm,height=5cm]{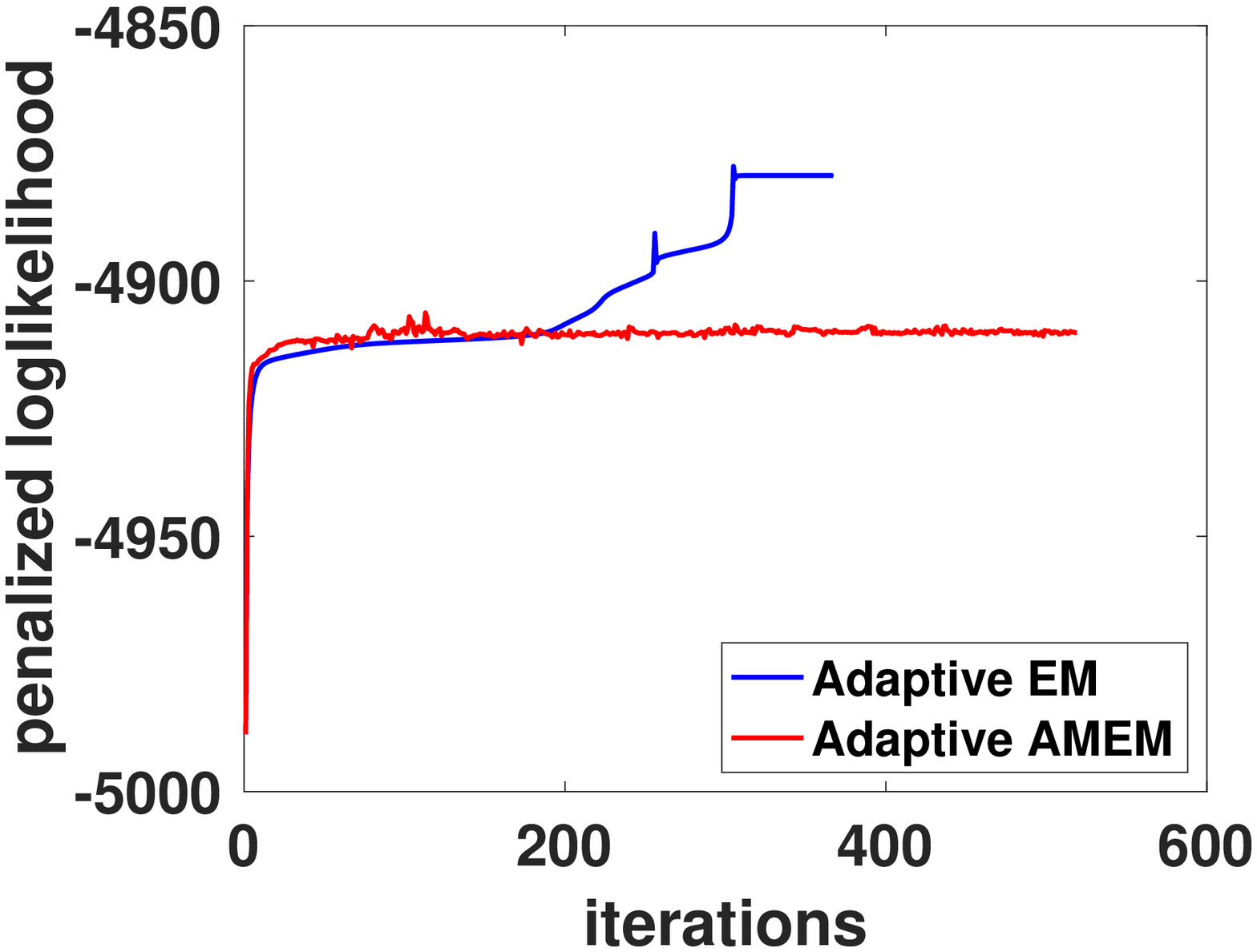}
\par\end{centering}
\caption{Case $K_{init}=5$ \textendash{} Application of AA to A-EM without
monotonicity control: History of penalized log-likelihood values as
a function of the number of iterations for synthetic data sets. (Left)
VWS data set. (Center) PS data set. (Right) VPS data set.\label{fig:without-MC-Kinit=00003D5}}
\end{figure*}

In practice, we often do not know in advance the exact model of the
Gaussian mixtures. Therefore, it is wise to start with a number of
Gaussian components larger than the suspected number of groups, and
rely on component adaptivity to find the correct model. However, the
application of AA without monotonicity control to A-EM (which we term
A-AAEM) fails catastrophically. To demonstrate this, we compare the
outcomes of A-EM with and without AA (with no monotonicity control)
to the synthetic data sets with $K_{init}=5$. Gaussian parameters
are initialized from the best run out of multiple trials of K-means
clustering with  $K=5$ clusters. The histories of the penalized log-likelihood
values as a function of the number of iterations are given in Fig.
\ref{fig:without-MC-Kinit=00003D5}.

From the figure, we observe that A-AAEM fails to annihilate unnecessary
Gaussian components, converges to the wrong solutions, and sometimes
is even slower than the non-accelerated version. Using $K_{init}>K_{exact}$
is equivalent to expanding the solution space, i.e., many local maxima
are created, and the accelerated A-EM manages to converge to one of
those local maxima in the expanded subspace. That AA+EM finds a suboptimal
solution is also the case without EM adaptivity (AAEM) when $K_{init}>K_{exact}$,
even if standard EM is able to find the correct solution. More specifically,
in our simulations with $K_{init}=5>K_{exact}$, standard EM finds
three Gaussians approximately matching the exact ones, and two Gaussians
with very small weights. However, in the converged AAEM solutions,
we observe that all five Gaussians have comparable weights, and the
means and covariance matrices for these Gaussians are far from the
exact parameters. As a result, at convergence, $\mathcal{L}(\boldsymbol{\theta}(\mbox{AAEM}))<\mathcal{L}(\boldsymbol{\theta}(\mbox{EM}))$
for the standard case and $\mathcal{PL}(\boldsymbol{\theta}(\mbox{A-AAEM}))<\mathcal{PL}(\boldsymbol{\theta}(\mbox{A-EM}))$
for the adaptive case, as we observed in Fig. \ref{fig:without-MC-Kinit=00003D5}.
We conclude that the application of AA to EM-GMM without monotonicity
control yields unreliable solutions and no performance advantage when
the exact number of components in the mixture is unknown.

\subsection{Application of AA to A-EM with monotonicity control (A-AMEM) to manufactured
data sets with $K_{init}=5$\label{subsec:AMEM-Kinit=00003D5}}

\begin{figure*}[tbh]
\begin{centering}
\includegraphics[width=5cm,height=5cm]{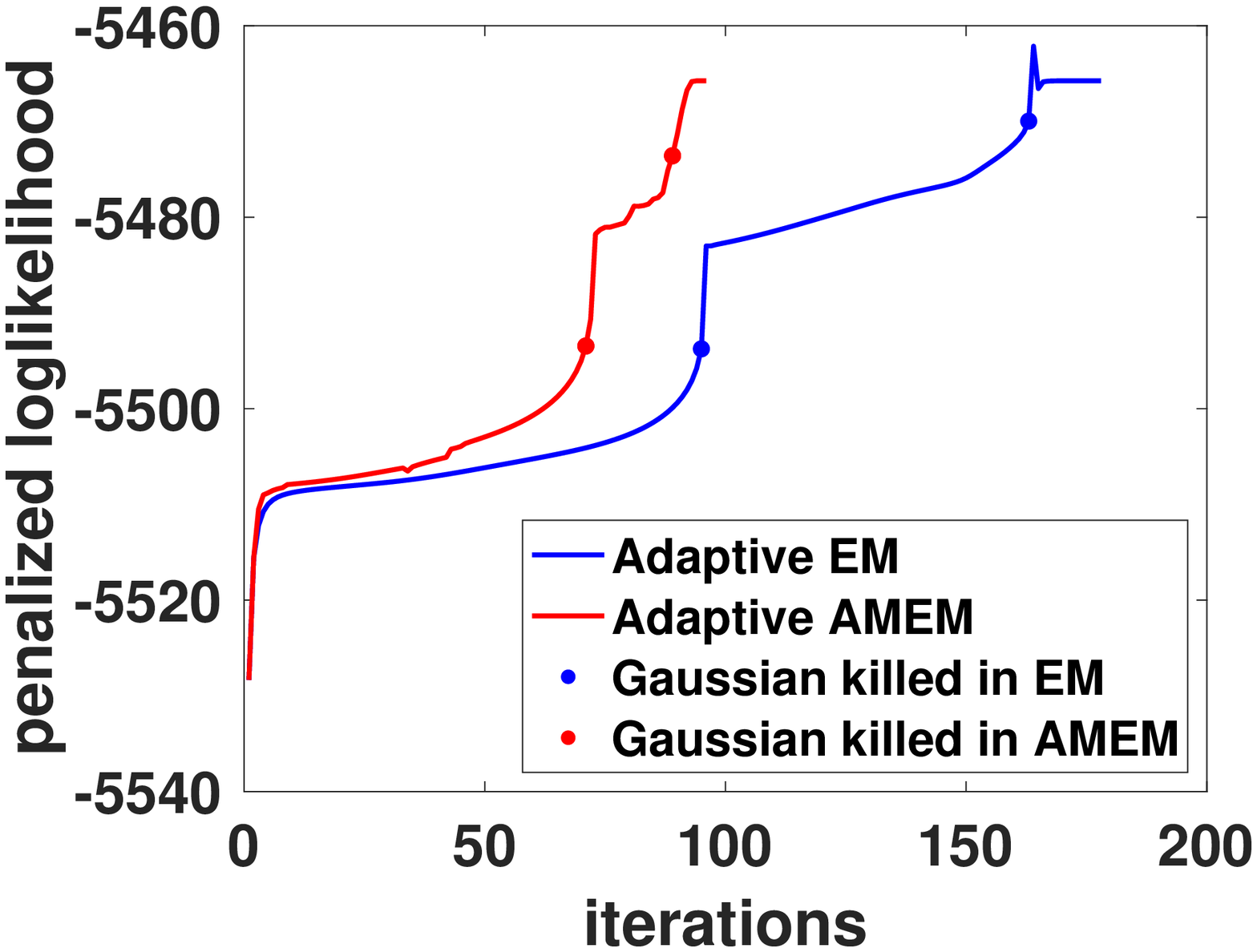}\includegraphics[width=5cm,height=5cm]{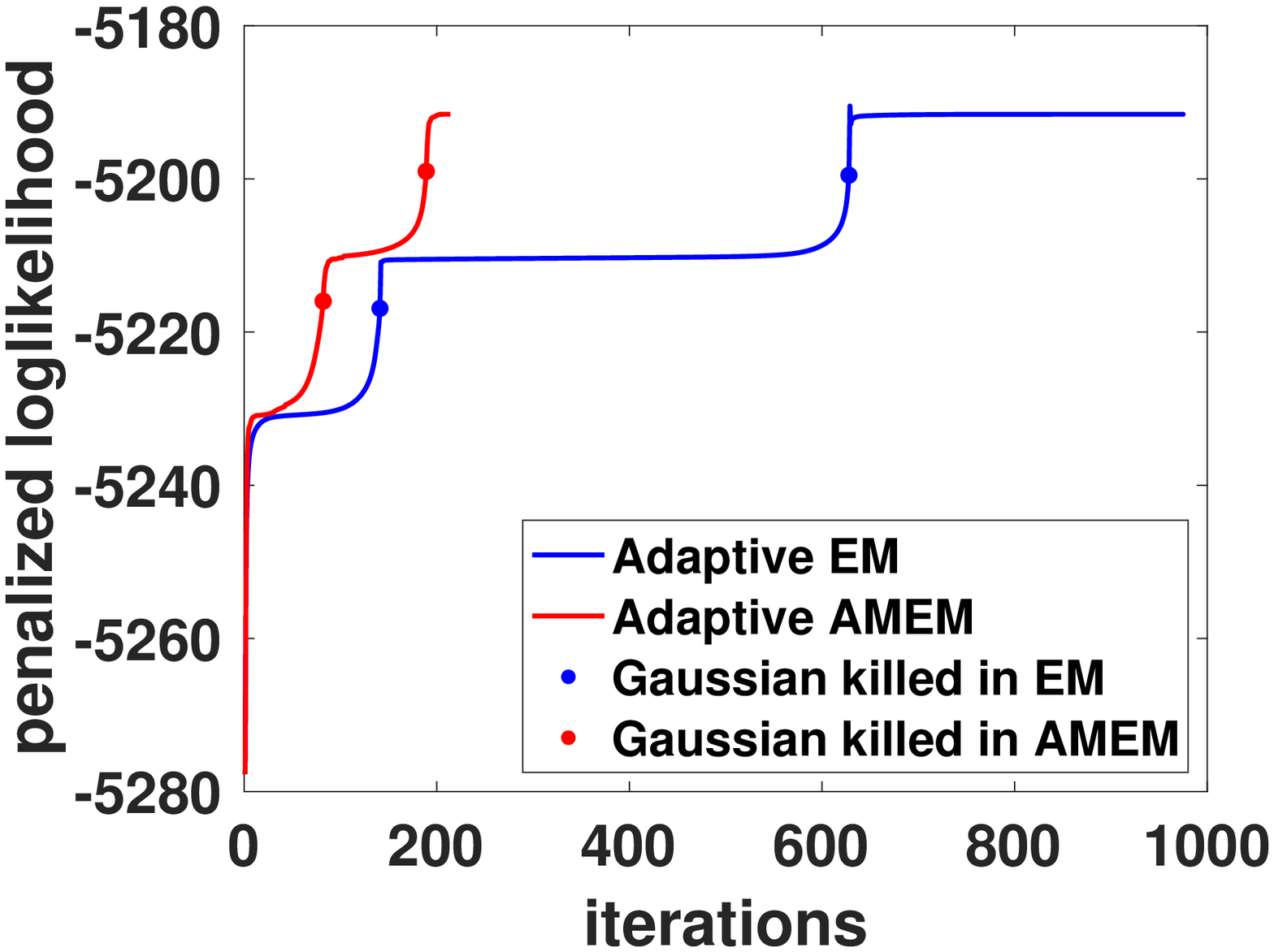}\includegraphics[width=5cm,height=5cm]{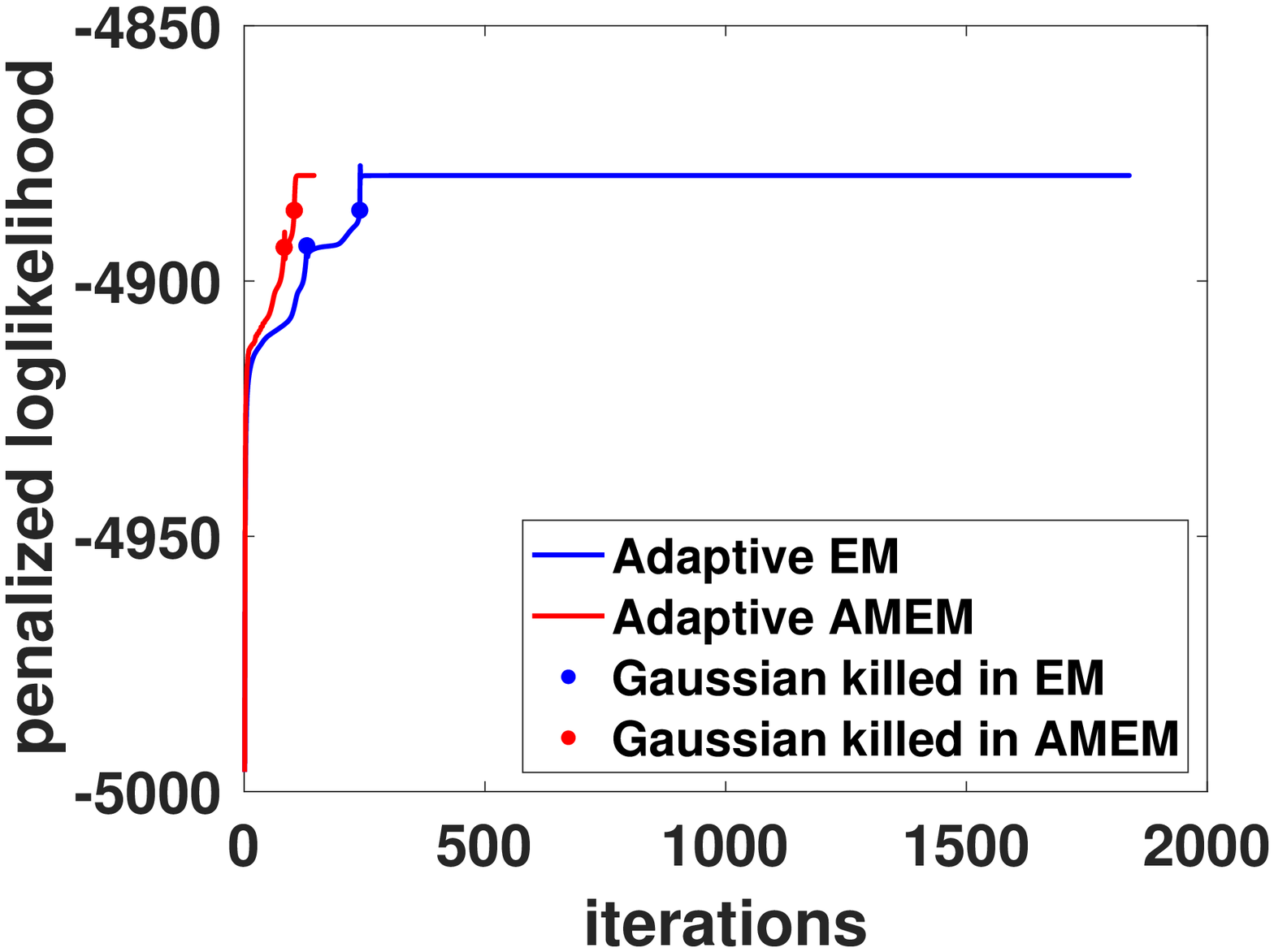}
\par\end{centering}
\caption{Case $K_{init}=5$: History of penalized log-likelihood values as
a function of the number of iterations for synthetic data sets. (Left)
VWS data set. (Center) PS data set. (Right) VPS data set. The blue
and red dots indicate the iterations where Gaussians are killed.\label{fig:PLLH-Kinit=00003D5}}
\end{figure*}

We use $K_{init}=5$ in both A-AMEM and A-EM, and the initialization
of Gaussian parameters is done in the same manner as described in
Section \ref{subsec:Failed-AA-application}. Note that we turn on
the monotonicity control step for this test. As shown in Fig. \ref{fig:PLLH-Kinit=00003D5},
both algorithms can detect the right number of Gaussian components
at convergence for the manufactured data sets, and find the same solution.
A-AMEM is able to annihilate Gaussian components somewhat faster than
A-EM, but converges very fast once the optimal number of Gaussian
components is reached, while A-EM continues to struggle to converge,
especially for the VPS data set.

\begin{figure*}[tbh]
\begin{centering}
\includegraphics[width=5cm,height=5cm]{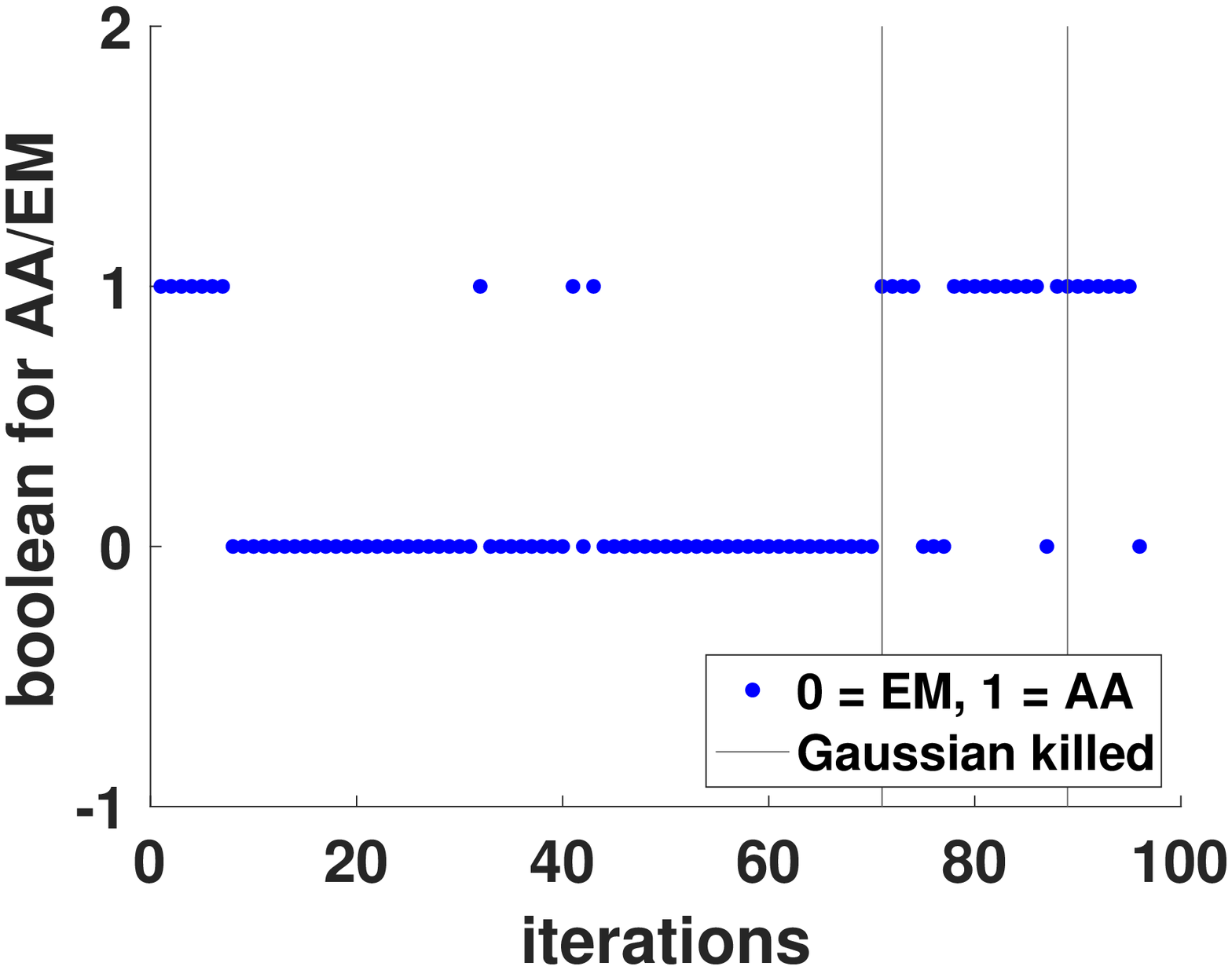}\includegraphics[width=5cm,height=5cm]{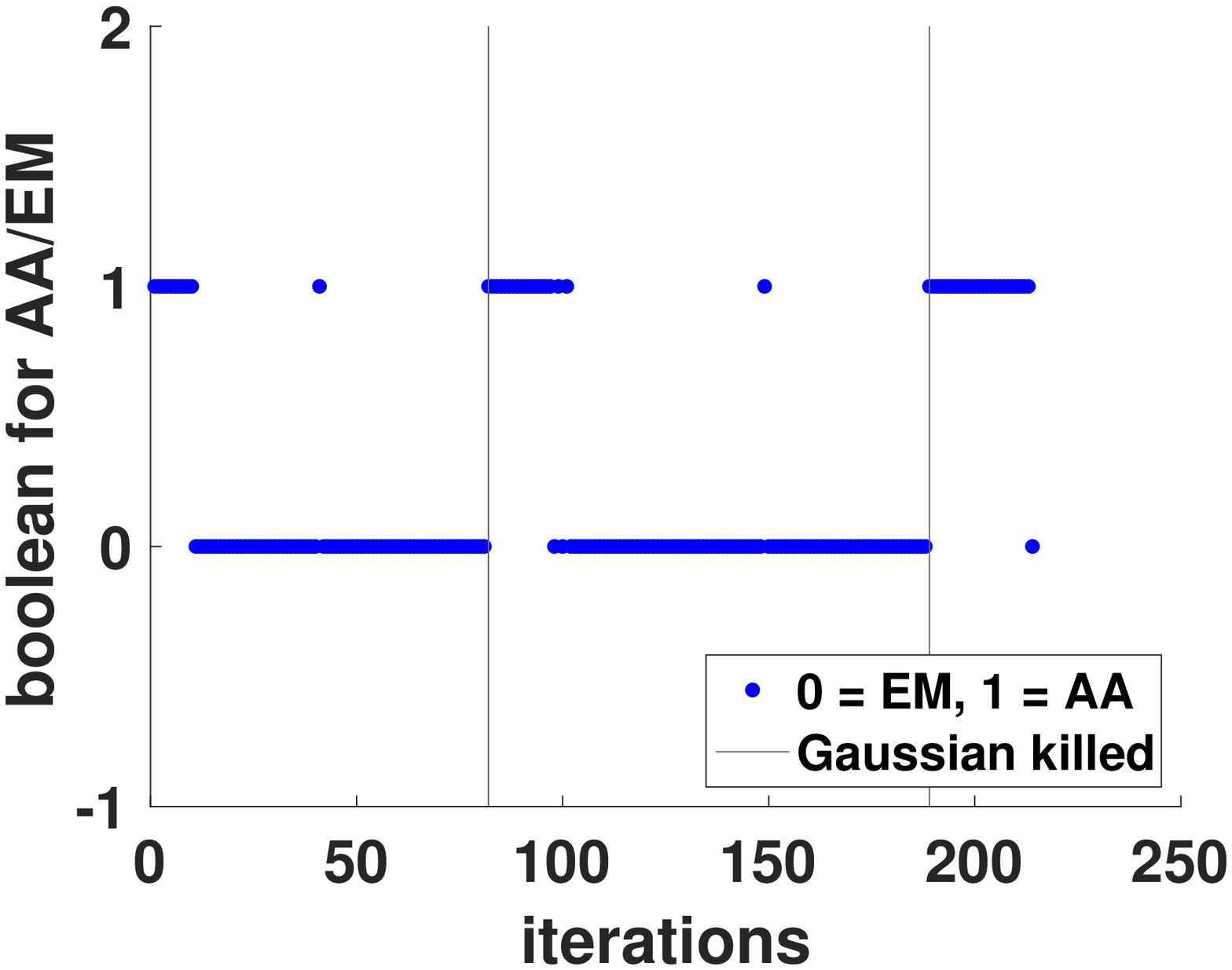}\includegraphics[width=5cm,height=5cm]{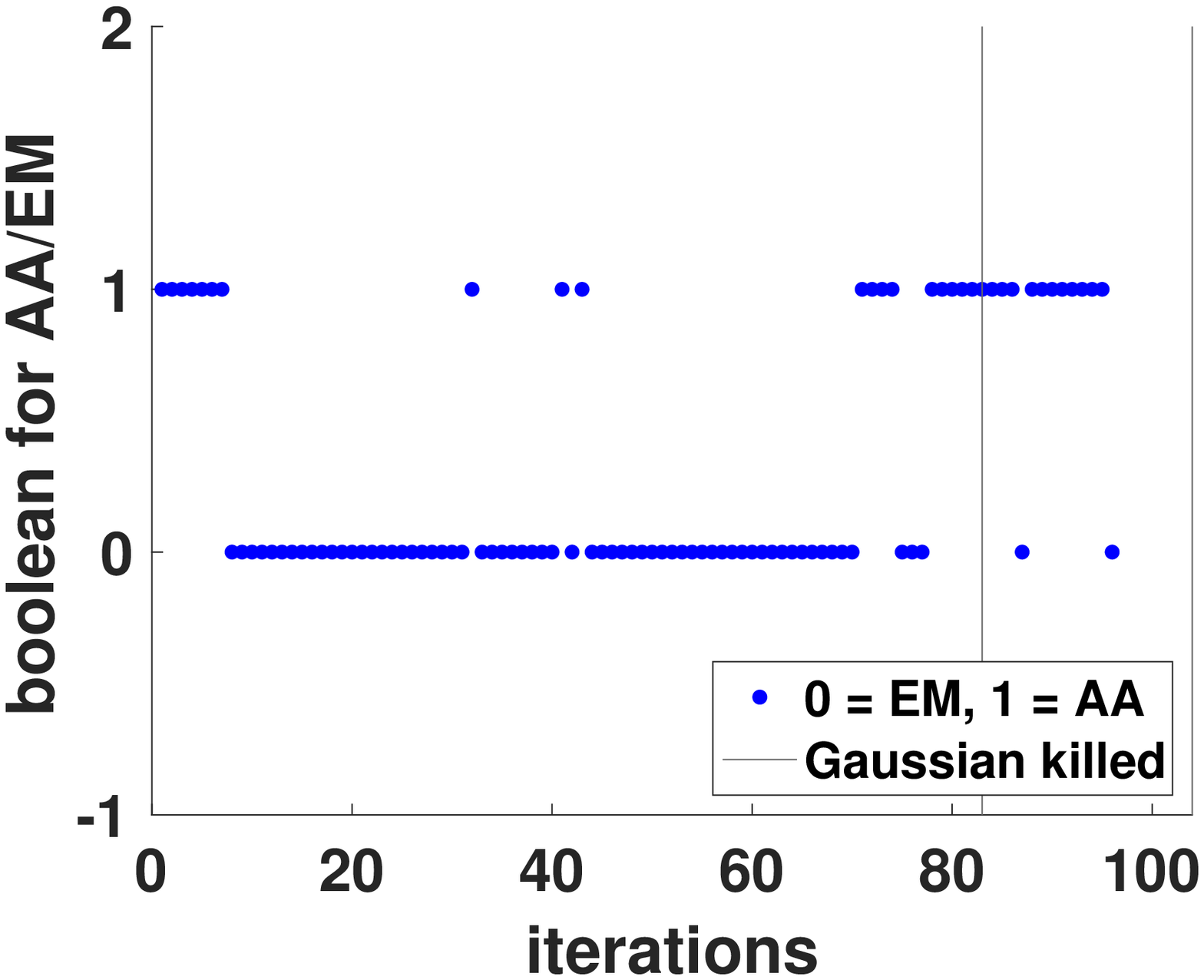}
\par\end{centering}
\caption{Case $K_{init}=5$: History of solution choices in A-AMEM iterations
for synthetic data sets. (Left) VWS data set. (Center) PS data set.
(Right) VPS data set.\label{fig:Solution-choices-Kinit=00003D5}}
\end{figure*}

We also investigate the dynamics of the component removal process
in A-AMEM due to the monotonicity control step. To this end, in
Fig. \ref{fig:Solution-choices-Kinit=00003D5} we show the binary
plots of the solution choices, AA iterate or EM iterate, during the
A-AMEM iteration for the three synthetic GMM data sets. In the binary
plots, the y-value for a given iteration is set to $0$ only when
A-AMEM falls back to the EM solution due to lack of monotonicity (and
not because of violations of positivity). Also, the vertical lines
in these binary plots represent the iterations when Gaussians are
killed.

From Fig. \ref{fig:Solution-choices-Kinit=00003D5} we see that A-AMEM
frequently reverts back to EM before reaching the optimal number of
Gaussian components in order to maintain the monotonicity of the penalized
log-likelihood function. This explains the fact that A-AMEM kills
Gaussians components at a comparable rate to A-EM. However, once the
optimal number of components is reached, the acceleration kicks in
aggressively and A-AMEM barely rolls back to EM. The last dot (with
y-value equal to zero) in each of the binary plots of Fig. \ref{fig:Solution-choices-Kinit=00003D5}
indicates the application of a final standard EM step for conservation
up to the second moments of the data points \cite{chen2020CR-EM-GM}.

Table \ref{tab:RF-Kinit=00003D5} shows the IRF and TRF for $K_{init}=5$,
demonstrating that A-AMEM converges faster than A-EM, especially for
the hard VPS data set, with speed-ups reaching an order of magnitude
for that case.
\begin{table}[H]
\centering{}\caption{Reduction factors for $K_{init}=5$.\label{tab:RF-Kinit=00003D5}}
\begin{tabular}{|c|c|c|c|}
\hline 
 & IRF & TRF & IRF/TRF\tabularnewline
\hline 
\hline 
VWS & 1.85 & 1.83 & \textbf{1.01}\tabularnewline
\hline 
PS & 4.56 & 3.94 & \textbf{1.16}\tabularnewline
\hline 
VPS & 12.60 & 10.01 & \textbf{1.26}\tabularnewline
\hline 
\end{tabular}
\end{table}

\subsection{Application of A-AMEM with monotonicity control to manufactured data
sets with $K_{init}=8$\label{subsec:AMEM-Kinit=00003D8}}

\begin{figure*}[tbh]
\begin{centering}
\includegraphics[width=5cm,height=5cm]{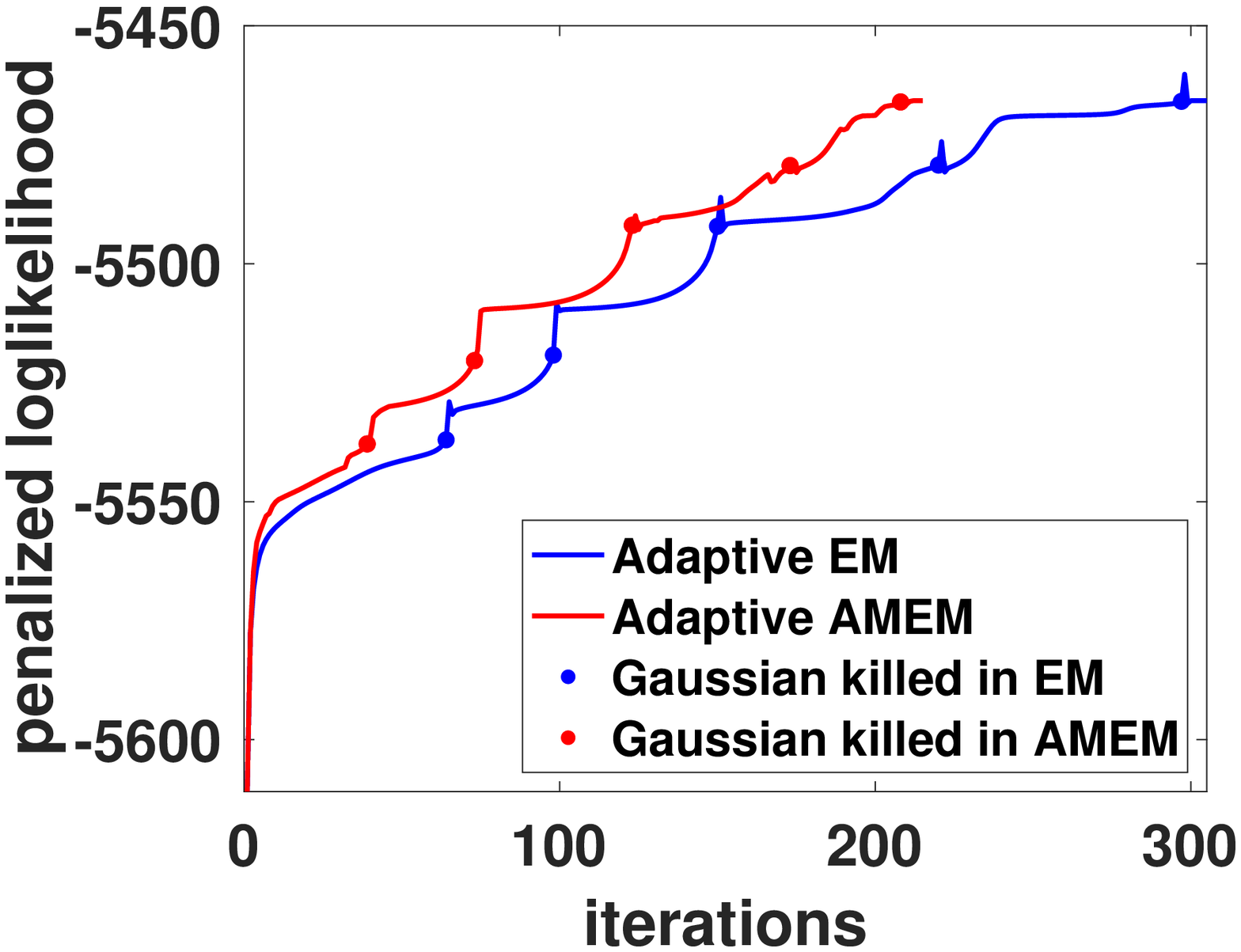}\includegraphics[width=5cm,height=5cm]{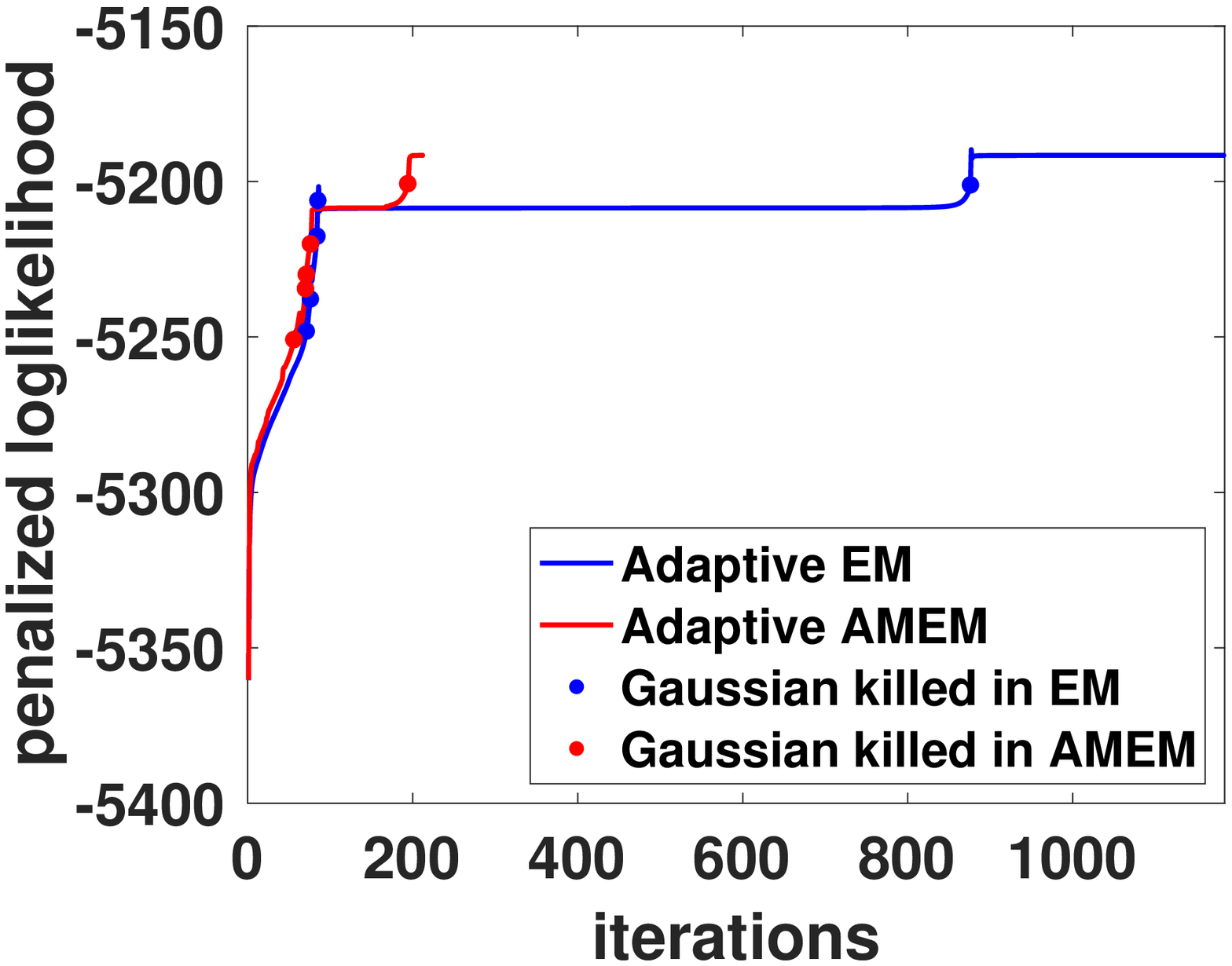}\includegraphics[width=5cm,height=5cm]{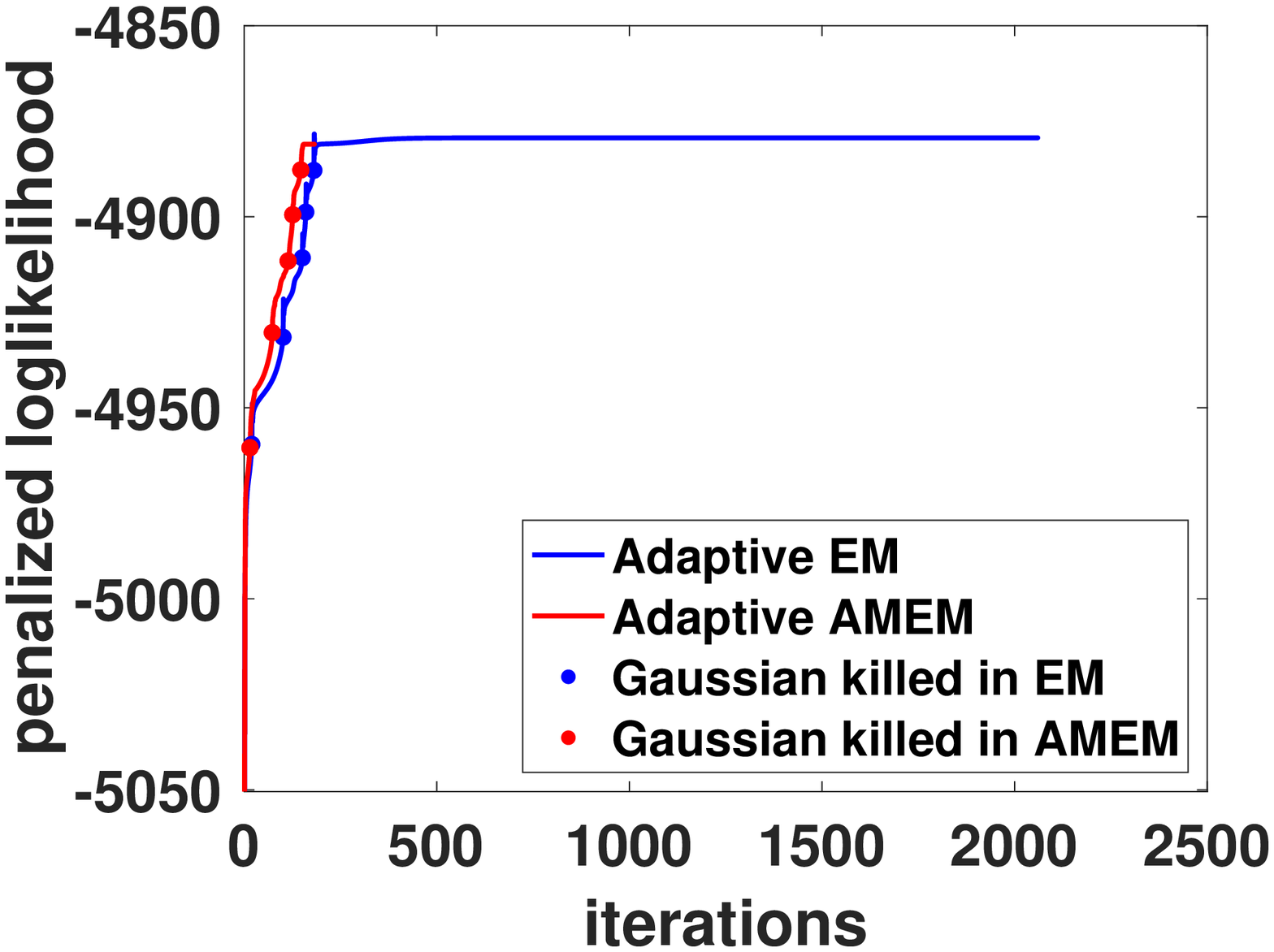}
\par\end{centering}
\caption{Case $K_{init}=8$: History of penalized log-likelihood values as
a function of the number of iterations for synthetic data sets. (Left)
VWS data set. (Center) PS data set. (Right) VPS data set. The blue
and red dots indicate the iterations where Gaussians are killed.\label{fig:PLLH-Kinit=00003D8}}
\end{figure*}

Next, we consider $K_{init}=8$, and perform the same numerical simulations
for the manufactured GMM data set using both adaptive non-accelerated
and accelerated EM. As before, Gaussian parameters are initialized
from the best run out of multiple runs of K-means clustering, but
with $K=8$ clusters. We present the plots of the penalized log-likelihood
values as a function of the number of iterations in Fig. \ref{fig:PLLH-Kinit=00003D8}.

Observations for this case are very similar to the $K_{init}=5$ case.
Again, in the beginning A-AMEM kills Gaussians components at a similar
rate to A-EM. However, once the optimal number of Gaussians components
is reached, A-AMEM quickly converges while A-EM struggles. In terms
of accuracy, both A-AMEM and A-EM converge to the same solutions and
yield the same final penalized log-likelihood value. The IRF and TRF
for this case are presented in Table \ref{tab:RF-Kinit=00003D8},
again demonstrating a significant speed-up.

It is clear from Figs. \ref{fig:PLLH-Kinit=00003D5} and \ref{fig:PLLH-Kinit=00003D8}
that A-AMEM does not remove Gaussian components much more efficiently
than A-EM. Since a larger $K_{init}$ requires more Gaussian killing,
it delays the onset of the fast convergence stage of the iteration,
resulting in the increase (degradation) of the IRF/TRF ratio between
the $K_{init}=5$ and $K_{init}=8$ cases observed in Tables \ref{tab:RF-Kinit=00003D5}
and \ref{tab:RF-Kinit=00003D8}. This result highlights the importance
of finding good guesses for the initial number of components. We will
discuss in detail our strategy for finding best guess for $K_{init}$
in Section \ref{subsec:Adaptive-multi-KMeans}.

\begin{table}[tbh]
\centering{}\caption{Reduction factors for $K_{init}=8$.\label{tab:RF-Kinit=00003D8}}
\begin{tabular}{|c|c|c|c|}
\hline 
 & IRF & TRF & IRF/TRF\tabularnewline
\hline 
\hline 
VWS & 1.42 & 1.46 & \textbf{0.97}\tabularnewline
\hline 
PS & 5.56 & 4.12 & \textbf{1.35}\tabularnewline
\hline 
VPS & 11.26 & 6.73 & \textbf{1.67}\tabularnewline
\hline 
\end{tabular}
\end{table}

\subsection{Performance of A-AMEM without Taylor expansion in the monotonicity
control step}

Previous results in Section \ref{subsec:AMEM-Kinit=00003D5} and Section
\ref{subsec:AMEM-Kinit=00003D8} have demonstrated that the cost per
iteration of A-AMEM is comparable to A-EM. To further highlight the
importance of using the Taylor expansion for the monotonicity control
step, we examine the performance of A-AMEM (vs. A-EM) using (\ref{eq:MC-original})
instead of (\ref{eq:MC-Taylor-expansion}). This requires an extra
evaluation of the penalized log-likelihood function (\ref{eq:pllh-gmm})
at every A-AMEM iteration. We consider the synthetic data sets with
$K_{init}=5$ and $K_{init}=8$. Table \ref{tab:RFs-no-TaylorExpand}
shows that the IRF/TRF ratios for these cases are $\sim2$, confirming
that one iteration of A-AMEM without Taylor expansion is about twice
as expensive as one A-EM iteration. 

\begin{table}[tbh]
\centering{}\caption{Reduction factors for $K_{init}=5,\,8$ without using Taylor expansion
in the monotonicity control step.\label{tab:RFs-no-TaylorExpand}}
\begin{tabular}{|c|c|c|c|c|c|c|}
\hline 
\multirow{2}{*}{} & \multicolumn{3}{c|}{$K_{init}=5$} & \multicolumn{3}{c|}{$K_{init}=8$}\tabularnewline
\cline{2-7} \cline{3-7} \cline{4-7} \cline{5-7} \cline{6-7} \cline{7-7} 
 & IRF & TRF & IRF/TRF & IRF & TRF & IRF/TRF\tabularnewline
\hline 
VWS & 1.73 & 1.01 & \textbf{1.71} & 1.44 & 0.88 & \textbf{1.64}\tabularnewline
\hline 
PS & 4.51 & 2.43 & \textbf{1.86} & 4.45 & 2.11 & \textbf{2.11}\tabularnewline
\hline 
VPS & 10.90 & 5.07 & \textbf{2.15} & 8.63 & 3.78 & \textbf{2.28}\tabularnewline
\hline 
\end{tabular}
\end{table}

\subsection{Application of A-AMEM to particle-in-cell (PIC) data sets\label{subsec:AMEM-to-PIC-data-sets}}

As a final test for A-AMEM, we consider data sets generated from PIC
simulations (see \cite{birdsall2004plasma} for detailed description
of PIC and \cite{chen2020CR-EM-GM} for a specific PIC application
of GMM). In particular, we consider the same 2D-3V Weibel electromagnetic
instability \cite{weibel1959spontaneously} as in Ref. \cite{chen2020CR-EM-GM}.
We partition the 2D spatial domain into $16\times16$ cells, with
$N\approx1024$ particles per cell per species. We run the simulations
until time $t=50$ (in inverse plasma frequency units) and record
the velocity points of all particles within each cell. The particle
velocities in the 3D velocity space in each cell provide the data
set. We then test the algorithms with selected cells. By applying
both A-AMEM and A-EM to these cells, we assume that the velocity distribution
functions (VDFs) can be approximated by a Gaussian mixture model of
unknown number of components. At time $t=50$, the simulations are
in the nonlinear phase and the electron VDFs strongly deviate from
the Maxwellian distribution. Therefore, for most of the cells, we
expect at least a few (anisotropic) components.

For demonstration purposes, we  choose cells 83, 155, 170, 243, with
the cell numbers defined lexicographically on the 2D spatial mesh
from-left-to-right and from-bottom-to-top. We apply A-EM and A-AMEM
to these data sets using $K_{init}=8$ and $K_{init}=10$ Gaussian
components. The plots of penalized log-likelihood values as functions
of iterations are given in Fig. \ref{fig:PLLH-DPIC-Kinit=00003D8,10}
and the IRFs and TRFs are recorded in Table \ref{tab:RF-Kinit=00003D8,10-DPIC}.

\begin{figure*}[tbh]
\begin{centering}
\includegraphics[width=4cm,height=4cm]{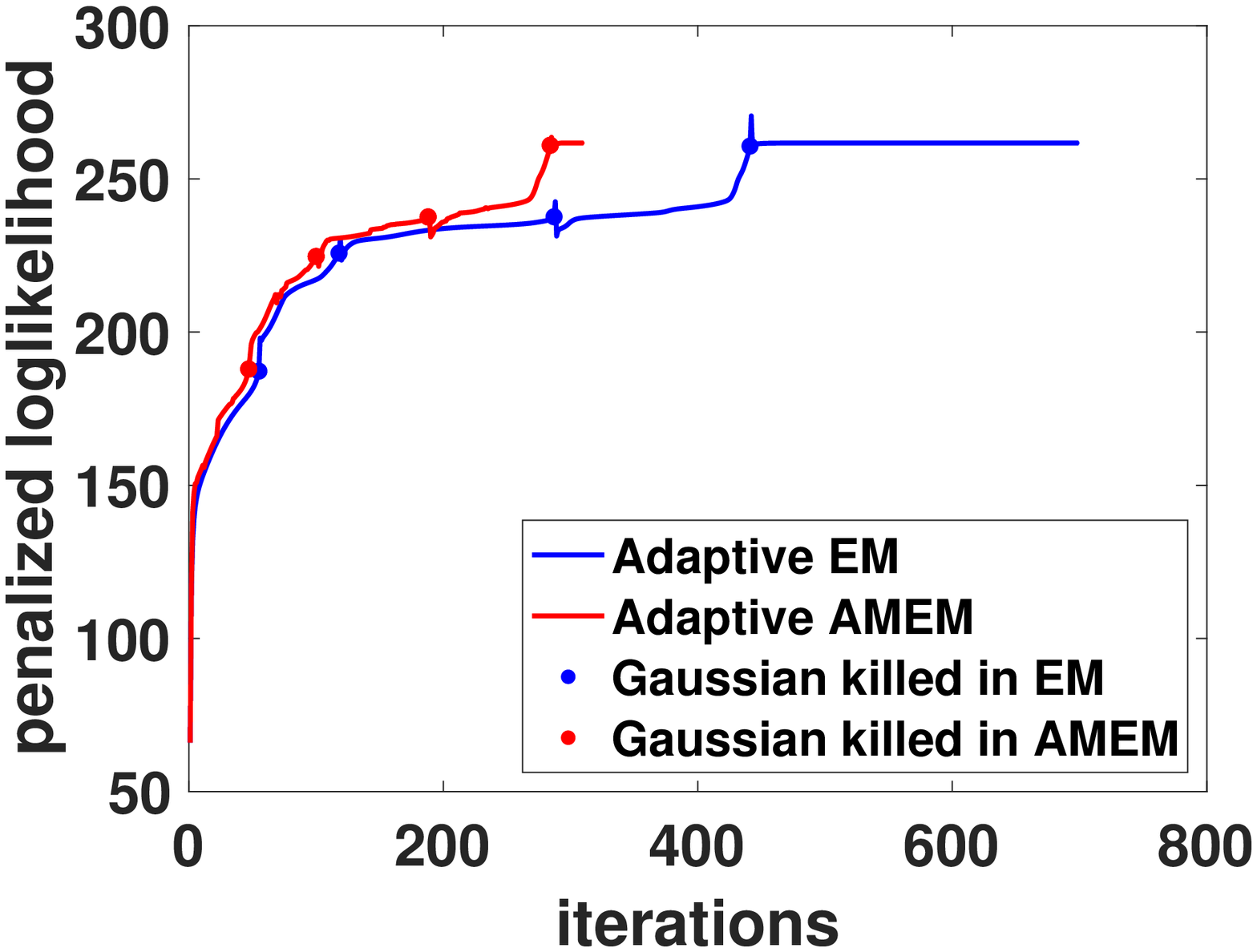}\includegraphics[width=4cm,height=4cm]{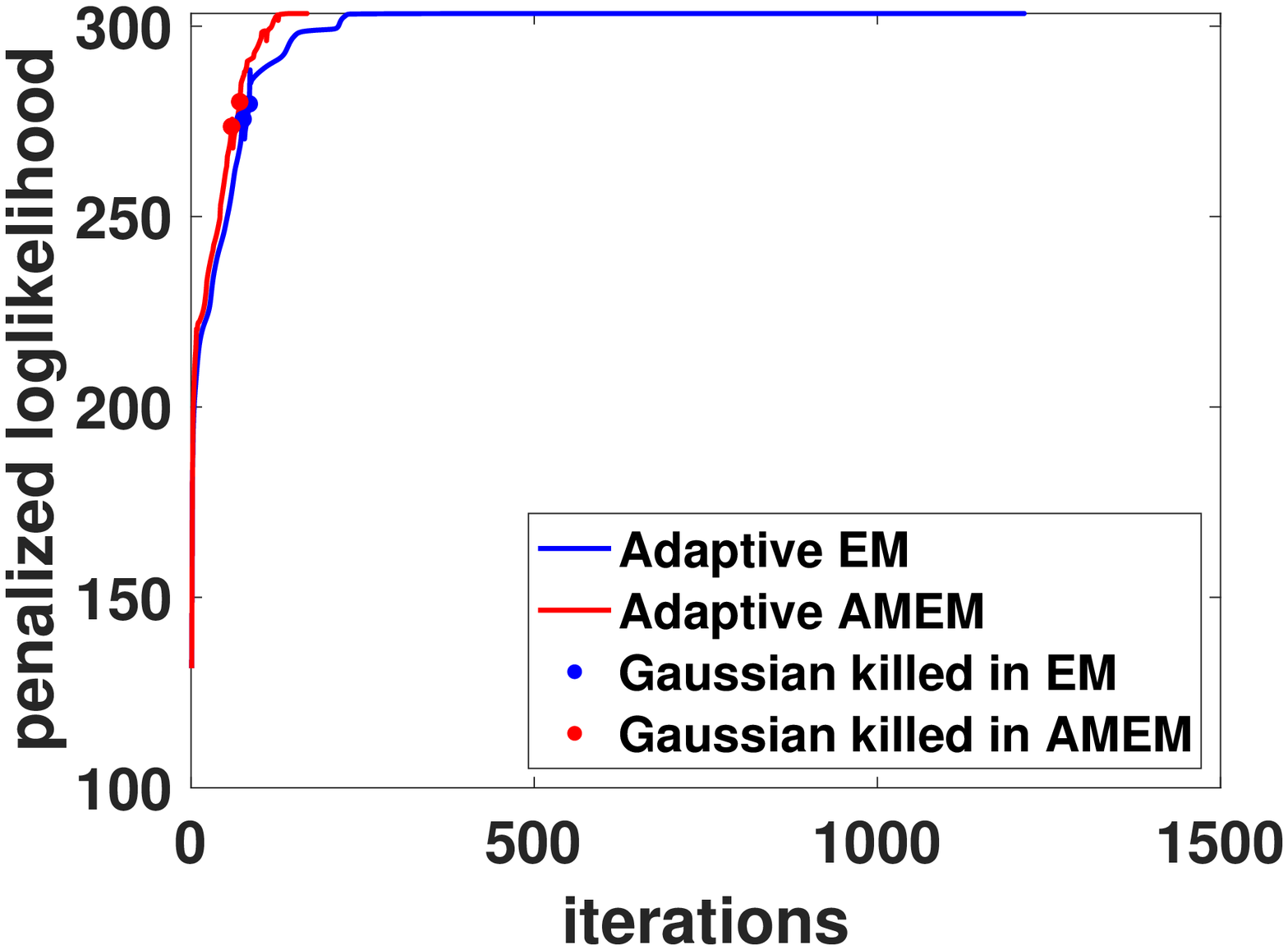}\includegraphics[width=4cm,height=4cm]{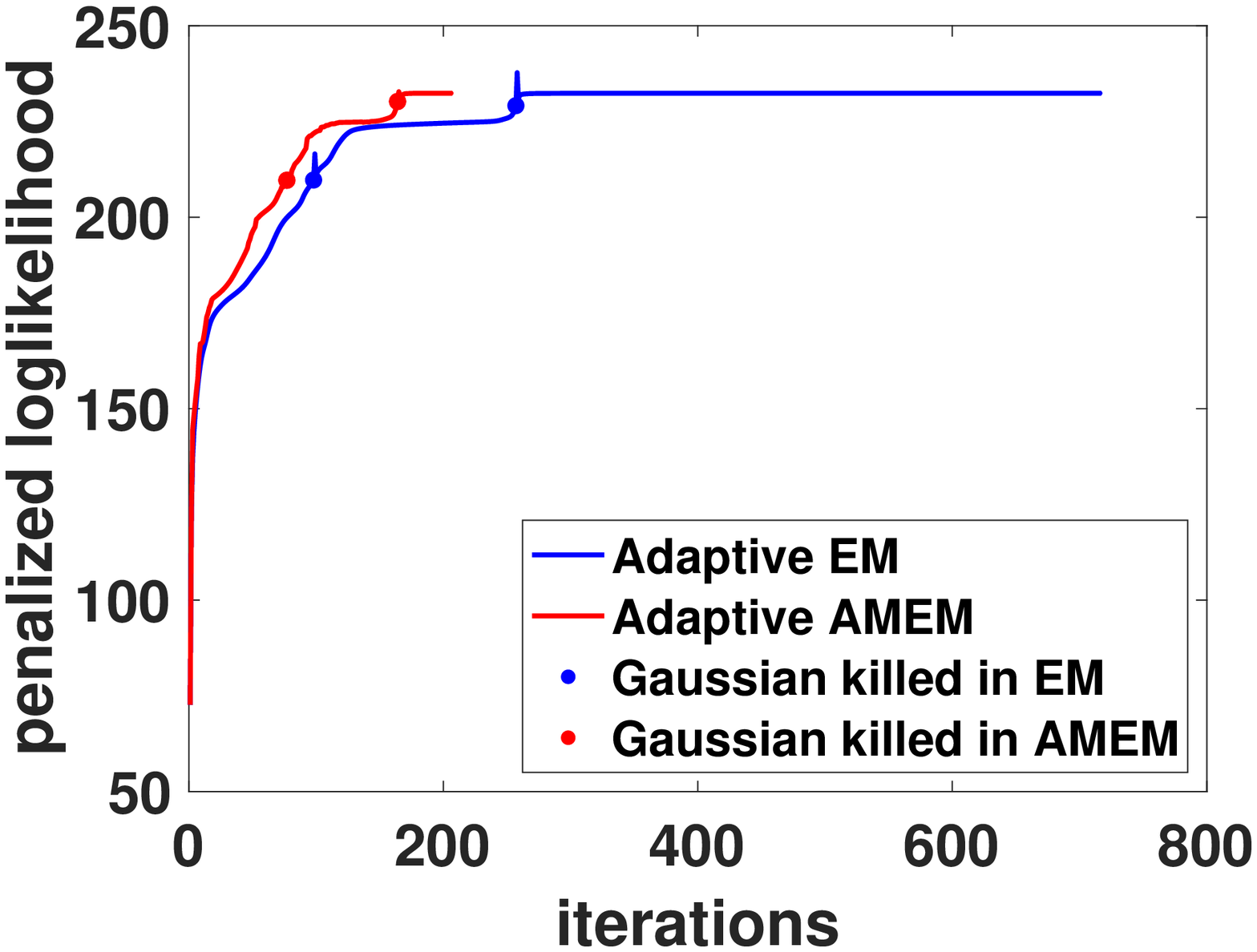}\includegraphics[width=4cm,height=4cm]{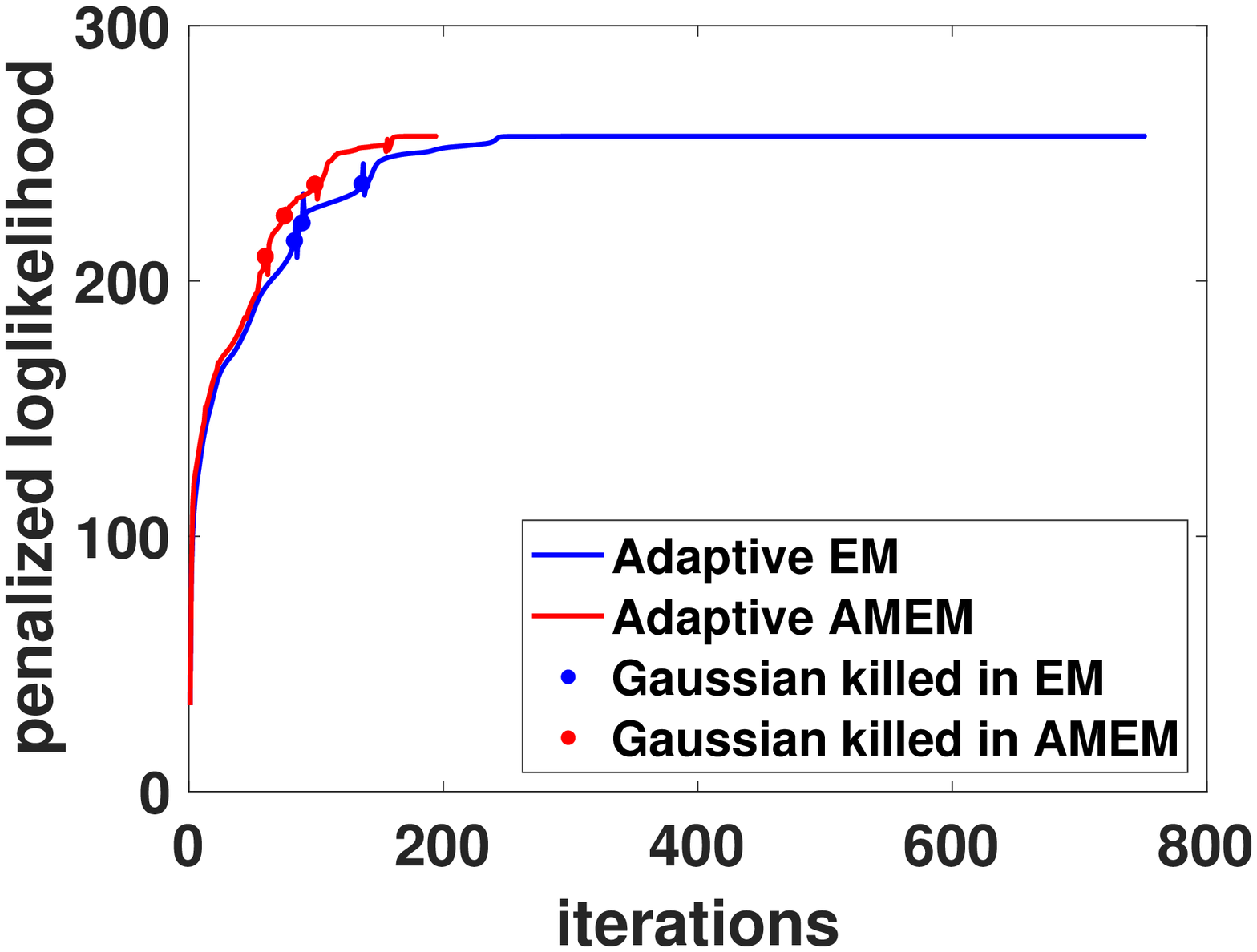}
\par\end{centering}
\[
\mbox{(a) }K_{init}=8
\]

\begin{centering}
\includegraphics[width=4cm,height=4cm]{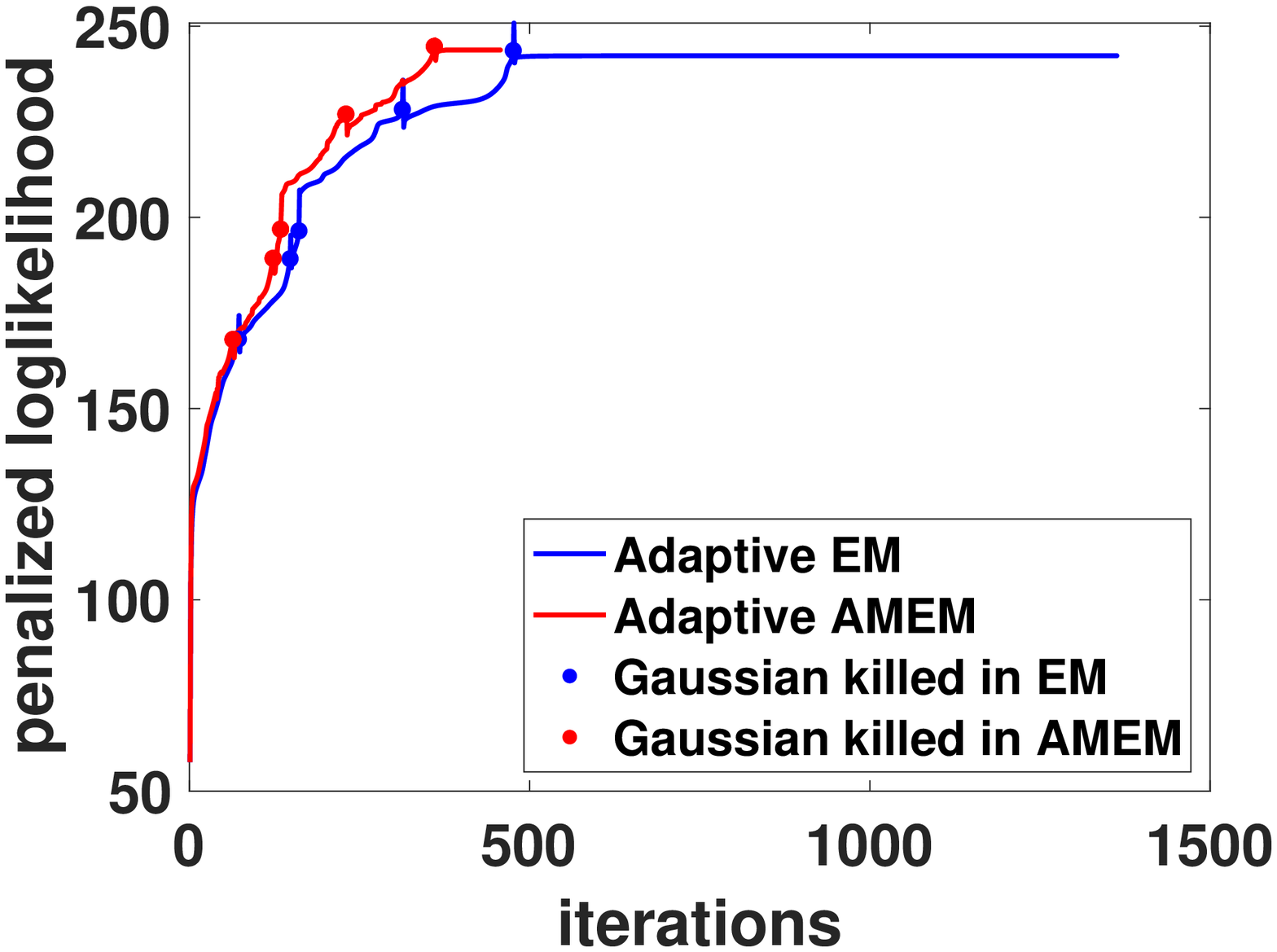}\includegraphics[width=4cm,height=4cm]{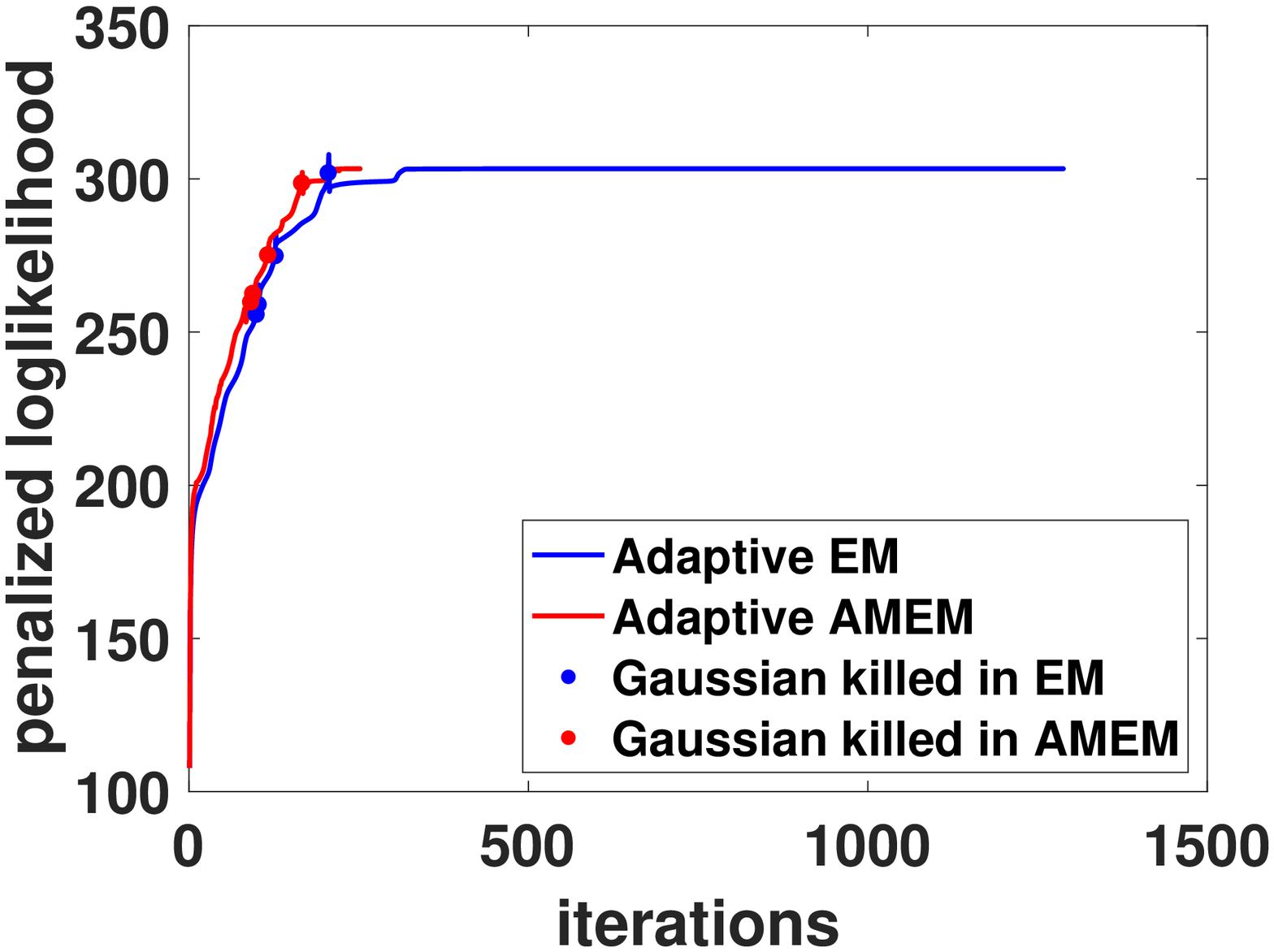}\includegraphics[width=4cm,height=4cm]{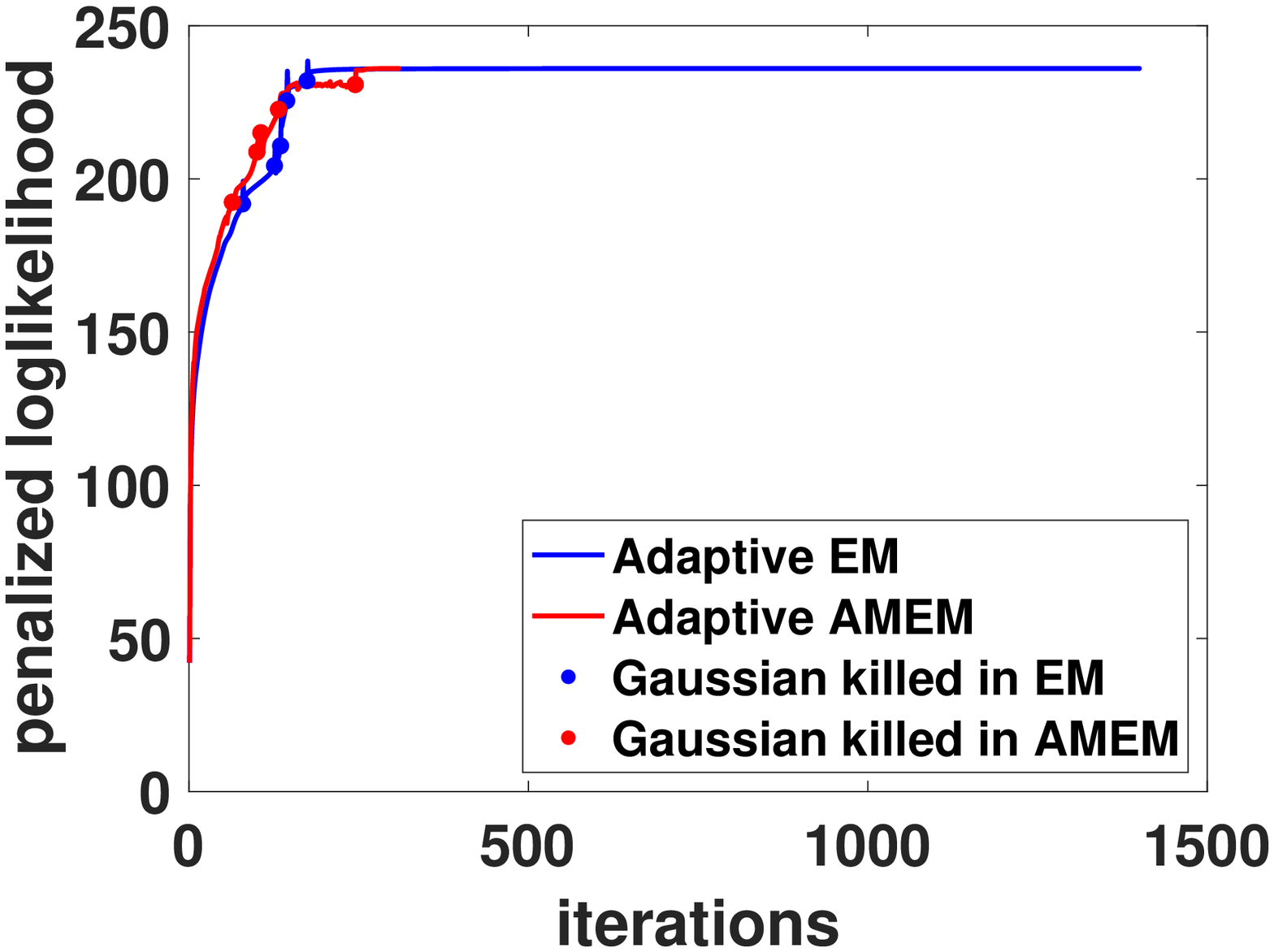}\includegraphics[width=4cm,height=4cm]{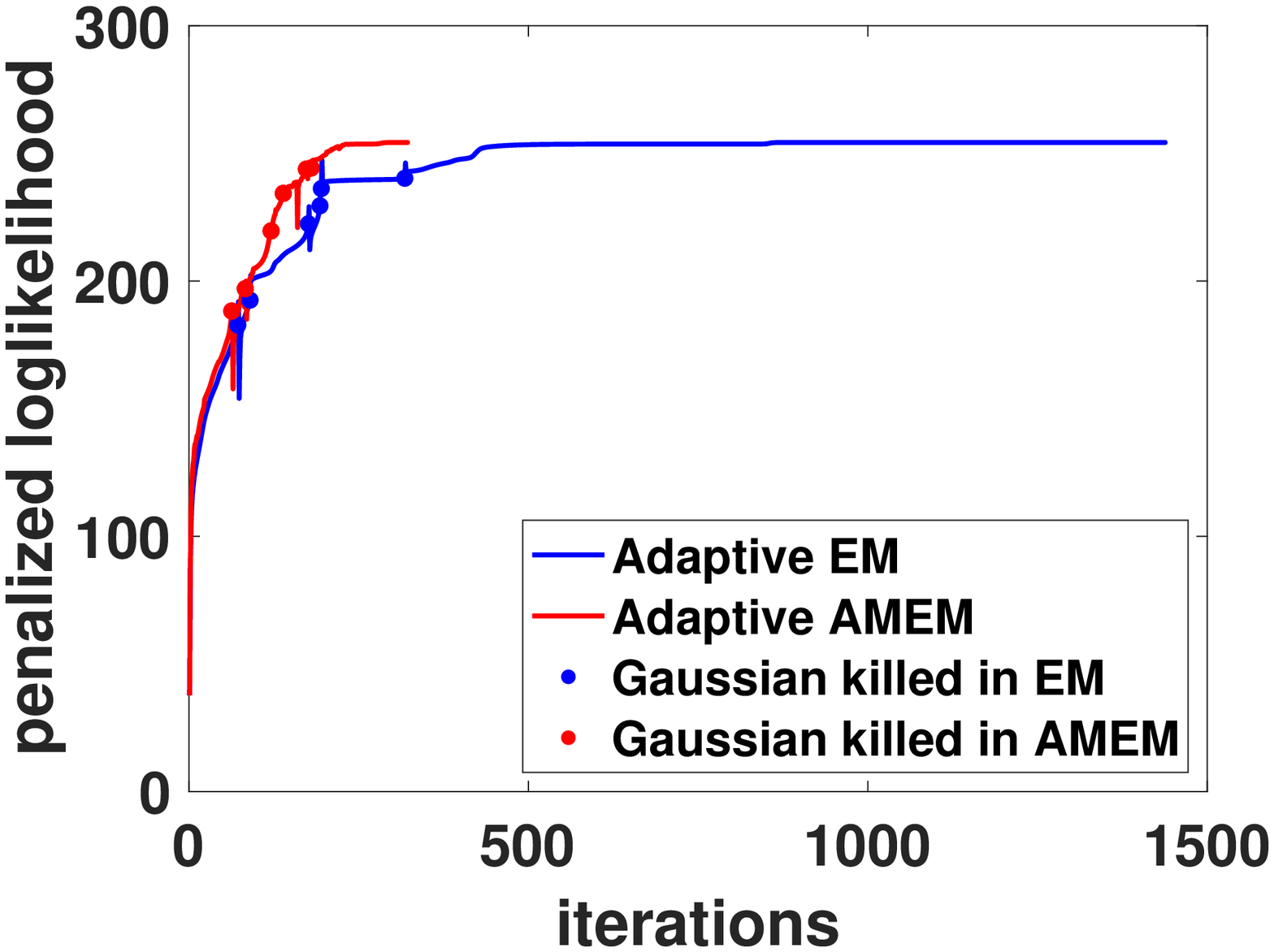}
\par\end{centering}
\[
\mbox{(b) }K_{init}=10
\]

\caption{Application of A-AMEM for PIC data sets with $K_{init}=8,\;10$: History
of penalized log-likelihood values as a function of the number of
iterations for selected particle-in-cell data sets. (From left to
right) Cell 83, 155, 170, 243. The blue and red dots indicate the
iterations where Gaussians are killed.\label{fig:PLLH-DPIC-Kinit=00003D8,10}}
\end{figure*}

\begin{table}[tbh]
\centering{}\caption{Reduction factors for $K_{init}=8,\;10$ for particle-in-cell data
sets.\label{tab:RF-Kinit=00003D8,10-DPIC}}
\begin{tabular}{|c|c|c|c|c|c|c|}
\hline 
\multirow{2}{*}{} & \multicolumn{3}{c|}{$K_{init}=8$} & \multicolumn{3}{c|}{$K_{init}=10$}\tabularnewline
\cline{2-7} \cline{3-7} \cline{4-7} \cline{5-7} \cline{6-7} \cline{7-7} 
 & IRF & TRF & IRF/TRF & IRF & TRF & IRF/TRF\tabularnewline
\hline 
Cell 83 & 2.25 & 1.95 & \textbf{1.15} & 2.97 & 2.54 & \textbf{1.17}\tabularnewline
\hline 
Cell 155 & 7.15 & 6.16 & \textbf{1.16} & 5.09 & 4.21 & \textbf{1.21}\tabularnewline
\hline 
Cell 170 & 3.69 & 3.25 & \textbf{1.14} & 4.52 & 3.10 & \textbf{1.46}\tabularnewline
\hline 
Cell 243 & 3.86 & 3.34 & \textbf{1.16} & 4.46 & 3.03 & \textbf{1.47}\tabularnewline
\hline 
\end{tabular}
\end{table}

We observe from Fig. \ref{fig:PLLH-DPIC-Kinit=00003D8,10} and Table
\ref{tab:RF-Kinit=00003D8,10-DPIC} that A-AMEM converges two to six
times faster than A-EM to the same solution, and that on average the
largest accelerations and smaller IRF/TRF ratios occur for smaller
$K_{init}$, again emphasizing the need for good initial guesses for
the number of components to maximize performance.

\subsection{Gap-statistic K-Means multi-initialization strategy\label{subsec:Adaptive-multi-KMeans}}

\textcolor{black}{The previous results highlight the importance of
choosing $K_{init}$ wisely. In addition to the gap statistic (GS)
method, we have explored other extensions of K-means clustering algorithms
proposed in the literature \cite{raykov2016adaptiveKmeans,Darken1990fastadaptkmeans,Rousseeuw1987silhouette,Nanjundan2019silhouette}
to obtain a good initial guess of the number of components for both
A-EM and A-AMEM. Firstly, we tested the so-called map-dp clustering
method introduced in \cite{raykov2016adaptiveKmeans}. This approach
works well for well separated data sets but not for poorly separated
ones. We observed that, when the Gaussians are poorly separated, this
approach usually yields a single $K=1$ component. In addition, the
effectiveness of the approach strongly depends on the choice of parameters,
making it brittle. Secondly, we tried the mean silhouette approach
\cite{Nanjundan2019silhouette,Rousseeuw1987silhouette,scikit-learn-selectNG}.
This approach works well for our data sets. However, it is very expensive
since it requires distance evaluation between all points in the data
sets, i.e., it has computational complexity of order $O(N^{2}D)$,
much greater than the complexity of the K-means algorithm, $O(NKD)$
for $N\gg K$.}

\textcolor{black}{We find that GS \cite{Tibshirani2001gapstatistics}
is cheap to use with K-means as the clustering method, and gives good
estimates for the initial number of components for our data sets (as
we show below). The initialization procedure for the Gaussian parameters
using GS (as described in Section \ref{subsec:initialization-details})
is summarized in Algorithm \ref{alg:Adaptive-K-means-alg}. We use
$K_{min}=2,\,K_{max}=10$, and $K_{adjust}=2$ for our numerical simulations.}
\begin{algorithm}[tbh]
\textcolor{black}{\caption{GS\textendash K-means algorithm\label{alg:Adaptive-K-means-alg}}
}

\textcolor{black}{Given data set $\mathbf{X}=(\mathbf{x}_{1,},\cdots,\mathbf{x}_{N}),$
the minimum number of cluster $K_{min}$ and the maximum number of
cluster $K_{max}$}
\begin{enumerate}
\item \textcolor{black}{Use the gap statistic (GS) method \cite{Tibshirani2001gapstatistics}
to estimate the number of components among $K_{min},\cdots,K_{max}$
clusters, i.e., $K_{opt}=\mbox{GS-method}(K_{min,},K_{max})$.}
\item \textcolor{black}{Set $K_{init}=K_{opt}+K_{adjust}$ for some $K_{adjust}>0$
to further avoid possible under-estimations.}
\item \textcolor{black}{Perform K-means algorithm using $K_{init}$ clusters
multiple times and select centroids from the best run which yields
the smallest inertia (SSE).}
\item \textcolor{black}{Set the initial means to be the centroids obtained
from Step 3 and compute the initial values for Gaussian weights and
covariance matrices.}
\end{enumerate}
\end{algorithm}

\begin{figure*}[tbh]
\begin{centering}
\includegraphics[width=4cm,height=4cm]{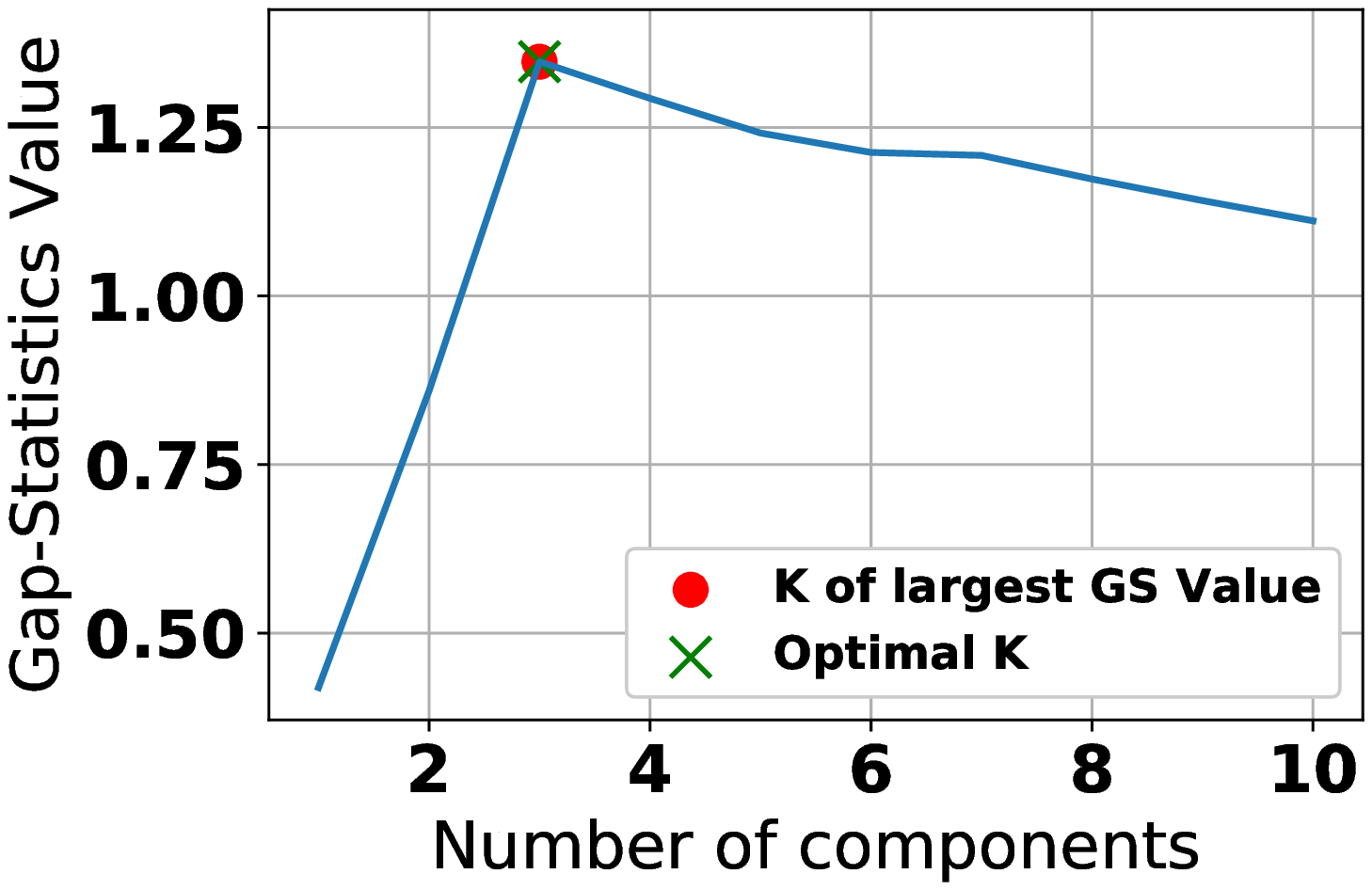}\includegraphics[width=4cm,height=4cm]{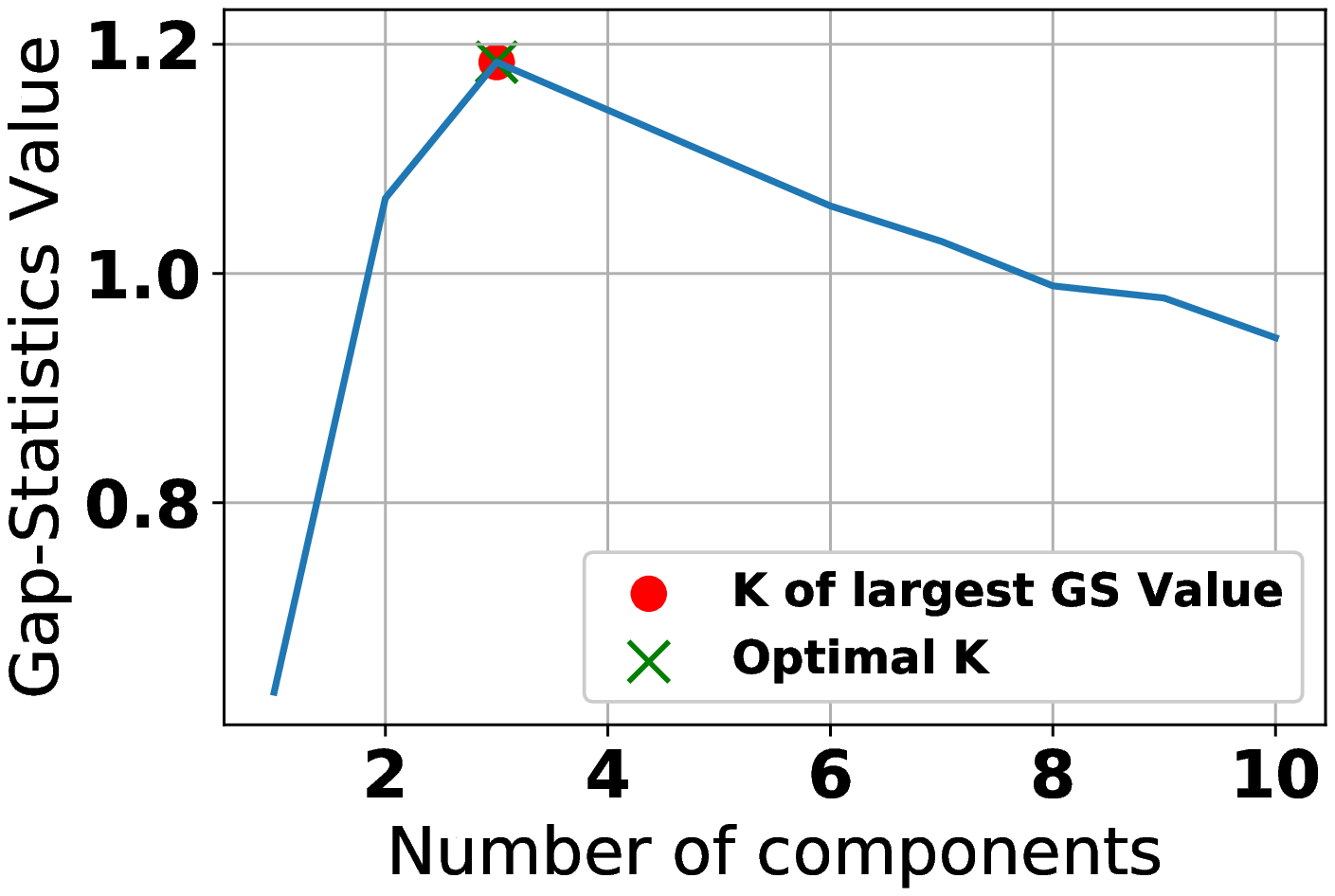}\includegraphics[width=4cm,height=4cm]{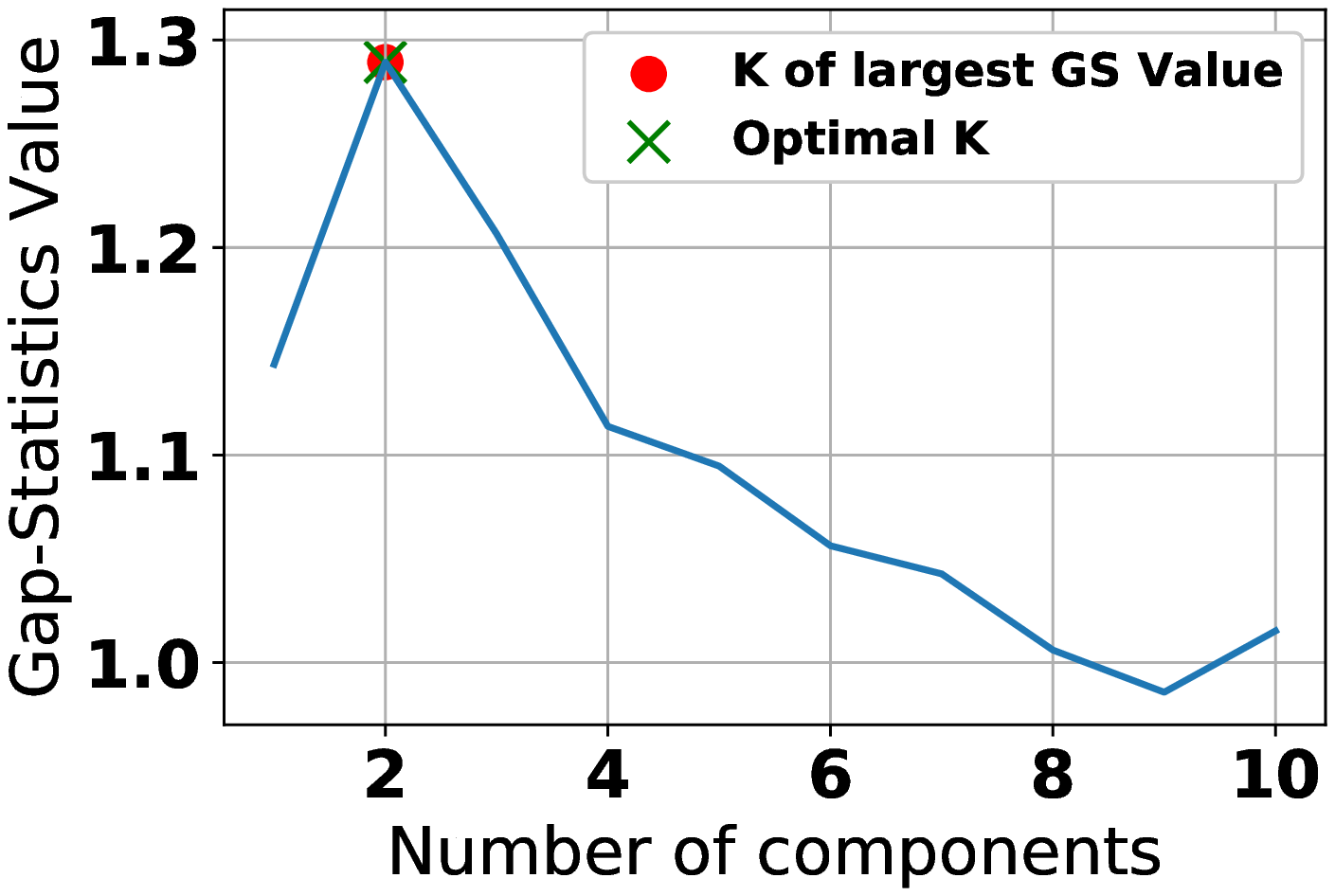}
\par\end{centering}
\caption{Estimated number of components by GS\textendash K-means for synthetic
data sets. (Left) VWS data set. (Center) PS data set. (Right) VPS
data set.\label{fig:decisions-gap-statistics}.}
\end{figure*}

\textcolor{black}{Fig. \ref{fig:decisions-gap-statistics} shows the
results for the number of components from the GS\textendash K-means
method for the synthetic data sets. Table \ref{tab:A-KM-A-AMEM--Results}
presents numerical results for applying A-EM and A-AMEM }initialized
with GS\textcolor{black}{\textendash }K-means to both synthetic and
PIC data sets. We remark that in these simulations, A-EM and A-AMEM
converge to the same solutions. We observe that the number of components
returned by the GS\textcolor{black}{\textendash }K-means initialization
for the synthetic data set is very accurate for the VWS and PS data
sets. For the VPS data set, since the Gaussians are highly overlapping,
the method under-estimates the number of components by only one. For
the synthetic tests, the optimal final number of components $K_{final}$
returned by A-EM and A-AMEM is equal to the number of components predicted
by GS\textendash K-means. For the PIC data sets, we observe that the
GS\textendash K-means underestimates the optimal number of components
by one or two groups, justifying the need to adjust $K_{est}$ by
some amount $K_{adjust}$ in Alg. \ref{alg:Adaptive-K-means-alg}.

Finally, we note that on average the wall-clock-time for the GS\textendash K-means
initialization step is about an order of magnitude smaller than adaptive
EM in our experiments, but this cost is likely amortized by the performance
gains from accurately guessing the number of components of the mixture.

\begin{table}[tbh]
\centering{}\caption{Results of using A-EM and A-AMEM with GS\textendash K-means for both
synthetic and particle-in-cell data sets\label{tab:A-KM-A-AMEM--Results}.
$K_{final}$ is the optimal number of groups returned by both A-EM
and A-AMEM.}
\begin{tabular}{|c|c|c|c|c|c|}
\hline 
\multirow{2}{*}[5cm]{} & \multirow{2}{*}{$K_{est}$} & \multirow{2}{*}{$K_{init}$} & \multirow{2}{*}{IRF} & \multirow{2}{*}{TRF} & \multirow{2}{*}{$K_{final}$}\tabularnewline
 &  &  &  &  & \tabularnewline
\hline 
\hline 
VWS & 3 & 5 & 1.85 & 1.76 & 3\tabularnewline
\hline 
PS & 3 & 5 & 4.39 & 4.03 & 3\tabularnewline
\hline 
VPS & 2 & 4 & 2.28 & 2.08 & 2\tabularnewline
\hline 
Cell 83 & 2 & 4 & 1.84 & 1.87 & 3\tabularnewline
\hline 
Cell 155 & 2 & 4 & 12.02 & 11.85 & 4\tabularnewline
\hline 
Cell 170 & 3 & 5 & 4.80 & 4.53 & 5\tabularnewline
\hline 
Cell 243 & 2 & 4 & 3.63 & 3.26 & 3\tabularnewline
\hline 
\end{tabular}
\end{table}

\section{Conclusion\label{sec:Conclusion}}

We propose for the first time an acce\textcolor{black}{lerated, monotonicity-preserving
algorithm for the }\textcolor{black}{\emph{adaptive}}\textcolor{black}{{}
EM-GMM algorithm that is significantly faster (in wall-clock time)
than its non-accelerated counterpart. The method combines the minimum-message-length
Bayesian information criterion with a monotonicity-controlled Anderson
acceleration (AA) solver. The targeted use of exact score functions
in the monotonicity control step of AA avoids computations of the
log-likelihood function, which is very expensive for GMM, and delivers
an overall very competitive method. The resulting A-AMEM converges
to the same solution as the non-accelerated version, strictly conserves
up to second moments of the observed data points and ensures the positive-definiteness
property of the solutions. The method has been tested on several synthetic
and data sets generated from PIC simulations. It shows significant
acceleration (from a few times up to more than an order of magnitude)
in terms of both iteration count and wall-clock time. Finally, we
have explored the use of a GS\textendash K-means initialization strategy,
which provides as good a guess for the number of components as practical
at a fraction of the cost of the adaptive EM algorithm. By eliminating
the guess work in the number of components (and thus avoiding the
necessary culling of unneeded mixture components), this strategy significantly
enhances both the efficiency and robustness of our approach in practical
applications.}


\appendix{}

\section*{Derivation of the Derivatives of Log-likelihood function w.r.t Gaussian
parameters\label{sec:Derivative-LLH-appendix}}

We consider the log-likelihood function with particle weight $\zeta_{j}$
for all $j=1,\cdots,N$ as follows:

\begin{equation}
\mathcal{L}(\boldsymbol{\theta})=\sum_{j=1}^{N}\zeta_{j}\;\mbox{ln}\bigg(\sum_{k=1}^{K}\omega_{k}G_{k}(\boldsymbol{x}_{j};\boldsymbol{\mu}_{k},\boldsymbol{\Sigma}_{k})\bigg)+\eta\,\bigg(\sum_{k=1}^{K}\omega_{k}-1\bigg)\label{eq:llh-gmm-appendix}
\end{equation}
where $N$ is the total number of observed points in the data set,
$K$ is the number of Gaussian components in the mixture and $\boldsymbol{x}_{j},\,j=1,\cdots,N$
is the observed data points with weights $\zeta_{j}$, and $\eta\,\bigg(\sum_{k=1}^{K}\omega_{k}-1\bigg)$
is the Lagrange multiplier term that enforces the normalization constraint
$\sum_{k=1}^{K}\omega_{k}=1$. We assume that $\sum_{j=1}^{N}\zeta_{j}=N$.
Parameters $\omega_{k},\boldsymbol{\mu}_{k}$ and $\boldsymbol{\Sigma}_{k}$
represent the weights, means, and covariance matrices of the $k\mbox{th}$
Gaussian. The Gaussian distribution $k\mbox{th}$ in the $D$-dimensional
space is defined as

\begin{equation}
G_{k}(\boldsymbol{x};\boldsymbol{\mu}_{k},\boldsymbol{\Sigma}_{k})=\frac{1}{\sqrt{(2\pi)^{D}\vert\boldsymbol{\Sigma}_{k}\vert}}e^{-\frac{1}{2}(\boldsymbol{x}-\boldsymbol{\mu}_{k})^{T}\boldsymbol{\Sigma}_{k}^{-1}(\boldsymbol{x}-\boldsymbol{\mu}_{k})}.\label{eq:gauss_dist-appendix}
\end{equation}

Firstly, we employ the following identities from the Matrix Cook Book
\cite{MatrixCookBook}:
\begin{equation}
\frac{\partial}{\partial\boldsymbol{s}}(\boldsymbol{x}-\boldsymbol{s})^{T}\boldsymbol{W}(\boldsymbol{x}-\boldsymbol{s})=-2\boldsymbol{W}(\boldsymbol{x}-\boldsymbol{s})\;,\label{eq:d/ds(x-s)^TW(x-s)}
\end{equation}

\begin{equation}
\frac{\partial}{\partial\boldsymbol{A}}\boldsymbol{v}^{T}\boldsymbol{A}^{-1}\boldsymbol{v}=-\boldsymbol{A}^{-1}\boldsymbol{v}\boldsymbol{v}^{T}\boldsymbol{A}^{-1}\;,\label{eq:ddAv^TA^-1v}
\end{equation}

\begin{equation}
\frac{\partial}{\partial\boldsymbol{A}}\vert\boldsymbol{A}\vert=\vert\boldsymbol{A}\vert\boldsymbol{A}^{-1}\;.\label{eq:ddAdetA}
\end{equation}
where $\boldsymbol{x},\,\boldsymbol{s}$ are vectors and $\boldsymbol{W},\,\boldsymbol{A}$
are matrices.

Secondly, we define the following quantities
\begin{equation}
f(\boldsymbol{x}_{j})=f(\boldsymbol{x}_{j};\,\boldsymbol{\theta})=\sum_{l=1}^{K}\omega_{l}G_{l}(\boldsymbol{x}_{j};\,\boldsymbol{\mu}_{l},\boldsymbol{\Sigma}_{l})\;,\label{eq:gmm-appendix}
\end{equation}

\begin{equation}
r_{jk}=\frac{\zeta_{j}\omega_{k}G_{k}(\boldsymbol{x}_{j};\,\boldsymbol{\mu}_{k},\boldsymbol{\Sigma}_{k})}{f(\boldsymbol{x}_{j})}\;,\label{eq:responsibilities-appendix}
\end{equation}
where $\boldsymbol{\theta}=(\boldsymbol{\theta}_{1},\cdots,\boldsymbol{\theta}_{k}),\,k=1,\cdots,K$
and $\boldsymbol{\theta}_{k}=(\omega_{k},\boldsymbol{\mu}_{k},\boldsymbol{\Sigma}_{k})$.
$K$ is the total number of components in the Gaussian mixture (GM).

Next, taking the derivative of $\mathcal{L}(\boldsymbol{\theta})$
w.r.t. the mean $\boldsymbol{\mu}_{k},$ we have
\begin{equation}
\begin{aligned}\frac{\partial\mathcal{L}(\boldsymbol{\theta})}{\partial\boldsymbol{\mu}_{k}} & =\sum_{j=1}^{N}\frac{\zeta_{j}}{f(\boldsymbol{x}_{j})}\frac{\partial}{\partial\boldsymbol{\mu}_{k}}\bigg[\frac{\omega_{k}e^{-\frac{1}{2}(\boldsymbol{x}_{j}-\boldsymbol{\mu}_{k})^{T}\Sigma_{k}^{-1}(\boldsymbol{x}_{j}-\boldsymbol{\mu}_{k})}}{(2\pi)^{D/2}\vert\boldsymbol{\Sigma}_{k}\vert^{1/2}}\bigg]\\
 & =\sum_{j=1}^{N}\frac{\zeta_{j}\omega_{k}G_{k}(\boldsymbol{x}_{j};\,\boldsymbol{\mu}_{k},\boldsymbol{\Sigma}_{k})}{f(\boldsymbol{x}_{j})}\,\times\\
 & \quad\quad\quad\frac{\partial\big(-\frac{1}{2}(\boldsymbol{x}_{j}-\boldsymbol{\mu}_{k})^{T}\boldsymbol{\Sigma}_{k}^{-1}(\boldsymbol{x}_{j}-\boldsymbol{\mu}_{k})\big)}{\partial\boldsymbol{\mu}_{k}}\\
 & =\sum_{j=1}^{N}-\frac{r_{jk}}{2}\,\frac{\partial}{\partial\boldsymbol{\mu}_{k}}\big((\boldsymbol{x}_{j}-\boldsymbol{\mu}_{k})^{T}\boldsymbol{\Sigma}_{k}^{-1}(\boldsymbol{x}_{j}-\boldsymbol{\mu}_{k})\big)\\
 & =\sum_{j=1}^{N}r_{jk}\,\boldsymbol{\Sigma}_{k}^{-1}\big(\boldsymbol{x}_{j}-\boldsymbol{\mu}_{k}\big).
\end{aligned}
\label{eq:dL-dmuk-appendix}
\end{equation}
where in the last equality of (\ref{eq:dL-dmuk-appendix}), we use
(\ref{eq:d/ds(x-s)^TW(x-s)}).

Taking the derivative of $\mathcal{L}(\boldsymbol{\theta})$ w.r.t.
the covariance matrix $\boldsymbol{\Sigma}_{k},$ we have
\begin{equation}
\begin{aligned}\frac{\partial\mathcal{L}(\boldsymbol{\theta})}{\partial\boldsymbol{\Sigma}_{k}} & =\sum_{j=1}^{N}\frac{\zeta_{j}}{f(\boldsymbol{x}_{j})}\,\frac{\partial}{\partial\boldsymbol{\Sigma}_{k}}\bigg[\frac{\omega_{k}e^{-\frac{1}{2}(\boldsymbol{x}_{j}-\boldsymbol{\mu}_{k})^{T}\boldsymbol{\Sigma}_{k}^{-1}(\boldsymbol{x}_{j}-\boldsymbol{\mu}_{k})}}{(2\pi)^{D/2}\vert\boldsymbol{\Sigma}_{k}\vert^{1/2}}\bigg]\\
 & =\sum_{j=1}^{N}\frac{\zeta_{j}\omega_{k}}{f(\boldsymbol{x}_{j})}\bigg\{\frac{e^{-\frac{1}{2}(\boldsymbol{x}_{j}-\boldsymbol{\mu}_{k})^{T}\boldsymbol{\Sigma}_{k}^{-1}(\boldsymbol{x}_{j}-\boldsymbol{\mu}_{k})}}{(2\pi)^{D/2}}\frac{\partial\vert\boldsymbol{\Sigma}_{k}\vert^{-\frac{1}{2}}}{\partial\boldsymbol{\Sigma}_{k}}\\
 & +G_{k}(\boldsymbol{x}_{j};\,\boldsymbol{\mu}_{k},\boldsymbol{\Sigma}_{k})\frac{\partial}{\partial\boldsymbol{\Sigma}_{k}}\big[-\frac{1}{2}(\boldsymbol{x}_{j}-\boldsymbol{\mu}_{k})^{T}\boldsymbol{\Sigma}_{k}^{-1}(\boldsymbol{x}_{j}-\boldsymbol{\mu}_{k})\big]\bigg\}\\
 & =\sum_{j=1}^{N}\frac{\zeta_{j}\omega_{k}}{f(\boldsymbol{x}_{j})}\bigg\{-\frac{1}{2}G_{k}(\boldsymbol{x}_{j};\,\boldsymbol{\mu}_{k},\boldsymbol{\Sigma}_{k})\boldsymbol{\Sigma}_{k}^{-1}\\
 & +\frac{1}{2}G_{k}(\boldsymbol{x}_{j};\,\boldsymbol{\mu}_{k},\boldsymbol{\Sigma}_{k})\boldsymbol{\Sigma}_{k}^{-1}(\boldsymbol{x}_{j}-\boldsymbol{\mu}_{k})(\boldsymbol{x}_{j}-\boldsymbol{\mu}_{k})^{T}\boldsymbol{\Sigma}_{k}^{-1}\bigg\}\\
 & =\sum_{j=1}^{N}\frac{r_{jk}}{2}\bigg\{-\boldsymbol{\Sigma}_{k}^{-1}+\boldsymbol{\Sigma}_{k}^{-1}(\boldsymbol{x}_{j}-\boldsymbol{\mu}_{k})(\boldsymbol{x}_{j}-\boldsymbol{\mu}_{k})^{T}\boldsymbol{\Sigma}_{k}^{-1}\bigg\}\,,
\end{aligned}
\label{eq:dL-dsigmak-appendix}
\end{equation}
in which we use (\ref{eq:ddAv^TA^-1v}) and (\ref{eq:ddAdetA}) to
go from the second equality to the third equality.

Taking the derivative of $\mathcal{L}(\boldsymbol{\theta})$ w.r.t.
the weight $\omega_{k}$ subject to the constraint $\sum_{k=1}^{K}\omega_{k}=1$,
we have
\begin{equation}
\begin{aligned}\frac{\partial\mathcal{L}(\boldsymbol{\theta})}{\partial\omega_{k}} & =\frac{\partial}{\partial\omega_{k}}\bigg\{\sum_{j=1}^{N}\zeta_{j}\;\mbox{ln}\bigg(\sum_{k=1}^{K}\omega_{k}G_{k}(\boldsymbol{x}_{j};\,\boldsymbol{\mu}_{k},\boldsymbol{\Sigma}_{k})\bigg)\\
 & +\eta\bigg(\sum_{k=1}^{K}\omega_{k}-1\bigg)\bigg\}\\
 & =\sum_{j=1}^{N}\frac{\zeta_{j}G_{k}(\boldsymbol{x}_{j};\,\boldsymbol{\mu}_{k},\boldsymbol{\Sigma}_{k})}{f(\boldsymbol{x}_{j})}+\eta\\
 & =\frac{1}{\omega_{k}}\sum_{j=1}^{N}r_{jk}+\eta\,,
\end{aligned}
\label{eq:dL-domegak-appendix}
\end{equation}
where $\eta=-N$ in (\ref{eq:dL-domegak-appendix}). In the case of
the penalized log-likelihood function, taking into account the penalized
terms, we have
\begin{equation}
\begin{aligned}\frac{\partial\mathcal{L}(\boldsymbol{\theta})}{\partial\omega_{k}} & =\frac{\partial}{\partial\omega_{k}}\bigg\{\sum_{j=1}^{N}\zeta_{j}\;\mbox{ln}\bigg(\sum_{k=1}^{K}\omega_{k}G_{k}(\boldsymbol{x}_{j};\,\boldsymbol{\mu}_{k},\boldsymbol{\Sigma}_{k})\bigg)\\
 & -\frac{T}{2}\sum_{k=1}^{K}\mbox{ln}(\omega_{k})+\eta\bigg(\sum_{k=1}^{K}\omega_{k}-1\bigg)\bigg\}\\
 & =\sum_{j=1}^{N}\frac{\zeta_{j}G_{k}(\boldsymbol{x}_{j};\,\boldsymbol{\mu}_{k},\boldsymbol{\Sigma}_{k})}{f(\boldsymbol{x}_{j})}-\frac{T}{2\omega_{k}}+\eta\\
 & =\frac{1}{\omega_{k}}\sum_{j=1}^{N}r_{jk}-\frac{T}{2\omega_{k}}+\eta\,,
\end{aligned}
\label{eq:dPL-domegak-appendix}
\end{equation}
where $\eta=-N+0.5TK$ for this case.

Setting (\ref{eq:dL-dmuk-appendix}), (\ref{eq:dL-dsigmak-appendix}),
and (\ref{eq:dL-domegak-appendix}) equal to zero and using $\eta=-N$,
we arrive at
\begin{equation}
\boldsymbol{\Sigma}_{k}^{-1}\bigg(\sum_{j=1}^{N}r_{jk}\,(\boldsymbol{x}_{j}-\boldsymbol{\mu}_{k})\bigg)=0\;\Leftrightarrow\;\boldsymbol{\mu}_{k}=\frac{\sum_{j=1}^{N}r_{jk}\boldsymbol{x}_{j}}{\sum_{j=1}^{N}r_{jk}}\,,\label{eq:dL-dmuk=00003D0-appendix}
\end{equation}

\begin{equation}
\begin{aligned} & \boldsymbol{\Sigma}_{k}^{-1}\bigg(\sum_{j=1}^{N}\frac{r_{jk}}{2}\bigg\{-\boldsymbol{\Sigma}_{k}+(\boldsymbol{x}_{j}-\boldsymbol{\mu}_{k})(\boldsymbol{x}_{j}-\boldsymbol{\mu}_{k})^{T}\bigg\}\bigg)\boldsymbol{\Sigma}_{k}^{-1}=0\;\\
\Leftrightarrow\, & \boldsymbol{\Sigma}_{k}=\frac{\sum_{j=1}^{N}r_{jk}(\boldsymbol{x}_{j}-\boldsymbol{\mu}_{k})(\boldsymbol{x}_{j}-\boldsymbol{\mu}_{k})^{T}}{\sum_{j=1}^{N}r_{jk}}\,,
\end{aligned}
\label{eq:dL-dsigmak=00003D0-appendix}
\end{equation}
and
\begin{equation}
\frac{1}{\omega_{k}}\sum_{j=1}^{N}r_{jk}=-\eta=N\;\Leftrightarrow\;\omega_{k}=\frac{\sum_{j=1}^{N}r_{jk}}{N}\,.\label{eq:dL-domegak=00003D0-appendix}
\end{equation}

Similarly, for the penalized log-likelihood case, setting (\ref{eq:dPL-domegak-appendix})
to zero and using $\eta=-N+0.5TK,$ we have

\begin{equation}
\begin{aligned} & \frac{1}{\omega_{k}}\sum_{j=1}^{N}r_{jk}-\frac{T}{2\omega_{k}}=-\eta=N-0.5TK\;\\
\Leftrightarrow\, & \omega_{k}=\frac{\sum_{j=1}^{N}r_{jk}-0.5T}{N-0.5TK}\,.
\end{aligned}
\label{eq:dPL-domegak=00003D0-appendix}
\end{equation}
Here, we note that $\frac{\partial\mathcal{L}(\boldsymbol{\theta})}{\partial\boldsymbol{\mu}_{k}}=\frac{\partial\mathcal{PL}(\boldsymbol{\theta})}{\partial\boldsymbol{\mu}_{k}}$
and $\frac{\partial\mathcal{L}(\boldsymbol{\theta})}{\partial\boldsymbol{\Sigma}_{k}}=\frac{\partial\mathcal{PL}(\boldsymbol{\theta})}{\partial\boldsymbol{\Sigma}_{k}}$.

\section*{Acknowledgment}

This work was supported by the U.S. Department of Energy, Office of
Science, through the Scientific Discovery through Advanced Computation
(SciDAC) Fusion Energy Sciences/Applied Scientific Computing Research
partnership program. This research used resources provided by the
Los Alamos National Laboratory Institutional Computing Program, and
was performed under the auspices of the National Nuclear Security
Administration of the U.S. Department of Energy at Los Alamos National
Laboratory, managed by Triad National Security, LLC under contract
89233218CNA000001.

\bibliographystyle{IEEEtran}
\bibliography{IEEEabrv,IEEEexample,2020NCC-AAMEM-preprint}

\begin{thebibliography}{10}
\providecommand{\url}[1]{#1}
\csname url@samestyle\endcsname
\providecommand{\newblock}{\relax}
\providecommand{\bibinfo}[2]{#2}
\providecommand{\BIBentrySTDinterwordspacing}{\spaceskip=0pt\relax}
\providecommand{\BIBentryALTinterwordstretchfactor}{4}
\providecommand{\BIBentryALTinterwordspacing}{\spaceskip=\fontdimen2\font plus
\BIBentryALTinterwordstretchfactor\fontdimen3\font minus
  \fontdimen4\font\relax}
\providecommand{\BIBforeignlanguage}[2]{{%
\expandafter\ifx\csname l@#1\endcsname\relax
\typeout{** WARNING: IEEEtran.bst: No hyphenation pattern has been}%
\typeout{** loaded for the language `#1'. Using the pattern for}%
\typeout{** the default language instead.}%
\else
\language=\csname l@#1\endcsname
\fi
#2}}
\providecommand{\BIBdecl}{\relax}
\BIBdecl

\bibitem{titterington1985statistical}
D.~M. Titterington, A.~F. Smith, and U.~E. Makov, \emph{Statistical analysis of
  finite mixture distributions}.\hskip 1em plus 0.5em minus 0.4em\relax Wiley,,
  1985.

\bibitem{mclachlan2004finite}
G.~J. McLachlan and D.~Peel, \emph{Finite mixture models}, ser. Wiley series in
  probability and statistics: Applied probability and statistics.\hskip 1em
  plus 0.5em minus 0.4em\relax John Wiley \& Sons, 2004.

\bibitem{mclachlan2019finite}
G.~J. McLachlan, S.~X. Lee, and S.~I. Rathnayake, ``Finite mixture models,''
  \emph{Annual review of statistics and its application}, vol.~6, pp. 355--378,
  2019.

\bibitem{fruhwirth2019handbook}
S.~Fruhwirth-Schnatter, G.~Celeux, and C.~P. Robert, \emph{Handbook of mixture
  analysis}.\hskip 1em plus 0.5em minus 0.4em\relax CRC press, 2019.

\bibitem{bishop2006pattern}
C.~M. Bishop, \emph{Pattern recognition and machine learning (Information
  science and statistics)}.\hskip 1em plus 0.5em minus 0.4em\relax Berlin,
  Heidelberg: Springer-Verlag, 2006.

\bibitem{murphy2012probabilistic}
K.~P. Murphy, \emph{Machine learning: A probabilistic perspective}.\hskip 1em
  plus 0.5em minus 0.4em\relax MIT Press, 2012.

\bibitem{liu2016bigdata}
S.~Liu, J.~McGree, Z.~Ge, and Y.~Xie, \emph{Computational and statistical
  methods for analysing big data with applications}.\hskip 1em plus 0.5em minus
  0.4em\relax Academic Press, 2016.

\bibitem{ullah2019BayesianMM}
I.~Ullah and K.~Mengersen, ``Bayesian mixture models and their big data
  implementations with application to invasive species presence-only data,''
  \emph{Journal of Big Data}, vol.~6, no.~29, 2019.

\bibitem{BouchonMeunier2006informationprocessing}
B.~Bouchon-Meunier, G.~Coletti, and R.~R. Yager, \emph{Modern information
  processing}.\hskip 1em plus 0.5em minus 0.4em\relax Elsevier Science, 2006.

\bibitem{yu2014}
Z.~Yu, C.~Chen, X.~Zheng, W.~Ding, and D.~Chen, ``Context-aware trust aided
  recommendation via {O}ntology and {G}aussian mixture model in big data
  environment,'' in \emph{2014 International Conference on Service
  Sciences}.\hskip 1em plus 0.5em minus 0.4em\relax IEEE, 2014, pp. 85--90.

\bibitem{Bi2018GMM-IS}
H.~Bi, H.~Tang, G.~Yang, H.~Shu, and J.-L. Dillenseger, ``Accurate image
  segmentation using {G}aussian mixture model with saliency map,''
  \emph{Pattern Analysis and Applications}, vol.~21, pp. 869--878, 2018.

\bibitem{Deledalle2018GGMM-ID}
C.-A. Deledalle, S.~Parameswaran, and T.~Q. Nguyen, ``Image denoising with
  generalized {G}aussian mixture model patch priors,'' \emph{SIAM Journal on
  Imaging Sciences}, vol.~11, no.~4, pp. 2568--2609, 2018.

\bibitem{Kalti2014-IS}
K.~Kalti and M.~Mahjoub, ``Image segmentation by {G}aussian mixture models and
  modified {FCM} algorithm,'' \emph{The International Arab Journal of
  Information Technology}, vol.~11, no.~1, pp. 11--18, 2014.

\bibitem{TMNguyen2011DGMM-IS}
T.~M. Nguyen and J.~Wu, ``Dirichlet {G}aussian mixture model: Application to
  image segmentation,'' \emph{Image and Vision Computing}, vol.~29, no.~12, pp.
  818--828, 2011.

\bibitem{Farnoosh2008GMM-IS}
R.~Farnoosh and B.~Zarpak, ``Image segmentation using {G}aussian mixture
  model,'' \emph{IUST International Journal of Engineering Science}, vol.~9,
  no. 1--2, pp. 29--32, 2008.

\bibitem{alekseenko2018bci}
A.~Alekseenko, T.~Nguyen, and A.~Wood, ``A deterministic-stochastic method for
  computing the {B}oltzmann collision integral in {O(MN)} operations,''
  \emph{Kinetic \& Related Models}, vol.~11, no.~5, pp. 1211--1234, 2018.

\bibitem{chen2020CR-EM-GM}
G.~Chen, L.~Chacon, and T.~Nguyen, ``An unsupervised machine-learning
  checkpoint-restart algorithm using {G}aussian mixtures for particle-in-cell
  simulations,'' \emph{submitted to Journal of Computational Physics,
  arXiv:2007.12273}, 2020.

\bibitem{dupuis2020characterizing}
R.~Dupuis, M.~V. Goldman, D.~L. Newman, J.~Amaya, and G.~Lapenta,
  ``Characterizing magnetic reconnection regions using {G}aussian mixture
  models on particle velocity distributions,'' \emph{The Astrophysical
  Journal}, vol. 889, no.~1, p.~22, 2020.

\bibitem{Barb2011GMM-Radiology}
A.~Barb, ``Gaussian mixture models for semantic ranking in domain specific
  databases with application in radiology,'' \emph{Central European Journal of
  Computer Science}, vol.~1, no.~3, pp. 266--279, 2011.

\bibitem{Reynolds2000GMM-SpeakerVerification}
D.~Reynolds, T.~Quatieri, and R.~Dunn, ``Speaker verification using adapted
  {G}aussian mixture models,'' \emph{Digital Signal Processing}, vol.~10, no.
  1--3, pp. 19--41, 2000.

\bibitem{Plataniotis200GMM-SignalProcess}
B.~Plataniotis, ``Gaussian mixtures and their applications to signal
  processing,'' in \emph{Advanced Signal Processing Handbook: Theory and
  Implementation for Radar, Sonar, and Medical Imaging Real Time Systems},
  S.~Stergiopoulos, Ed.\hskip 1em plus 0.5em minus 0.4em\relax Taylor \&
  Francis Group, 2000, ch.~3.

\bibitem{yu2015gaussian}
D.~Yu and L.~Deng, ``Gaussian mixture models,'' in \emph{Automatic Speech
  Recognition}.\hskip 1em plus 0.5em minus 0.4em\relax Springer, 2015, pp.
  13--21.

\bibitem{dempster1977maximum}
A.~P. Dempster, N.~M. Laird, and D.~B. Rubin, ``Maximum likelihood from
  incomplete data via the {EM} algorithm,'' \emph{Journal of the royal
  statistical society. Series B (methodological)}, vol.~39, no.~1, pp. 1--38,
  1977.

\bibitem{redner1984mixture}
R.~A. Redner and H.~F. Walker, ``Mixture densities, maximum likelihood and the
  {EM} algorithm,'' \emph{SIAM review}, vol.~26, no.~2, pp. 195--239, 1984.

\bibitem{mclachlan2007algorithm}
G.~J. McLachlan and T.~Krishnan, \emph{The {EM} algorithm and
  extensions}.\hskip 1em plus 0.5em minus 0.4em\relax John Wiley \& Sons, 2007,
  vol. 382.

\bibitem{mclachlan2014number}
G.~J. McLachlan and S.~Rathnayake, ``On the number of components in a
  {G}aussian mixture model,'' \emph{Wiley Interdisciplinary Reviews: Data
  Mining and Knowledge Discovery}, vol.~4, no.~5, pp. 341--355, 2014.

\bibitem{figueiredo2000unsupervised}
M.~A. Figueiredo and A.~K. Jain, ``Unsupervised selection and estimation of
  finite mixture models,'' in \emph{Proceedings 15th International Conference
  on Pattern Recognition. ICPR-2000}, vol.~2.\hskip 1em plus 0.5em minus
  0.4em\relax IEEE, 2000, pp. 87--90.

\bibitem{figueiredo2002unsupervised}
------, ``Unsupervised learning of finite mixture models,'' \emph{IEEE
  Transactions on Pattern Analysis and Machine Intelligence}, vol.~24, no.~3,
  pp. 381--396, 2002.

\bibitem{wallace2005statistical}
C.~S. Wallace, \emph{Statistical and inductive inference by minimum message
  length}.\hskip 1em plus 0.5em minus 0.4em\relax Berlin, Heidelberg:
  Springer-Verlag, 2005.

\bibitem{hansen2001modelselection}
M.~Hansen and B.~Yu, ``Model selection and the principle of minimum description
  length,'' \emph{Journal of the American Statistical Association}, vol.~96,
  no. 454, pp. 746--774, 2001.

\bibitem{corduneanu2001variational}
A.~Corduneanu and C.~M. Bishop, ``Variational {B}ayesian model selection for
  mixture distributions,'' in \emph{Artificial intelligence and Statistics},
  vol. 2001.\hskip 1em plus 0.5em minus 0.4em\relax Morgan Kaufmann Waltham,
  MA, 2001, pp. 27--34.

\bibitem{lange2013optimization}
K.~Lange, \emph{Optimization}.\hskip 1em plus 0.5em minus 0.4em\relax Springer
  Science \& Business Media, 2013.

\bibitem{meng1993maximum}
X.-L. Meng and D.~B. Rubin, ``Maximum likelihood estimation via the {ECM}
  algorithm: A general framework,'' \emph{Biometrika}, vol.~80, no.~2, pp.
  267--278, 1993.

\bibitem{liu1994ecme}
C.~Liu and D.~B. Rubin, ``The {ECME} algorithm: a simple extension of {EM} and
  {ECM} with faster monotone convergence,'' \emph{Biometrika}, vol.~81, no.~4,
  pp. 633--648, 1994.

\bibitem{fessler1994space}
J.~A. Fessler and A.~O. Hero, ``Space-alternating generalized
  {E}xpectation-{M}aximization algorithm,'' \emph{IEEE Transactions on signal
  processing}, vol.~42, no.~10, pp. 2664--2677, 1994.

\bibitem{meng1997algorithm}
X.-L. Meng and D.~Van~Dyk, ``The {EM} algorithm, an old folk song sung to a
  fast new tune,'' \emph{Journal of the Royal Statistical Society: Series B
  (Statistical Methodology)}, vol.~59, no.~3, pp. 511--567, 1997.

\bibitem{liu1998parameter}
C.~Liu, D.~B. Rubin, and Y.~N. Wu, ``Parameter expansion to accelerate {EM} :
  the {PX-EM} algorithm,'' \emph{Biometrika}, vol.~85, no.~4, pp. 755--770,
  1998.

\bibitem{celeux2001component}
G.~Celeux, S.~Chr{\'e}tien, F.~Forbes, and A.~Mkhadri, ``A component-wise {EM}
  algorithm for mixtures,'' \emph{Journal of Computational and Graphical
  Statistics}, vol.~10, no.~4, pp. 697--712, 2001.

\bibitem{wolfe1970pattern}
J.~H. Wolfe, ``Pattern clustering by multivariate mixture analysis,''
  \emph{Multivariate Behavioral Research}, vol.~5, no.~3, pp. 329--350, 1970.

\bibitem{louis1982finding}
T.~A. Louis, ``Finding the observed information matrix when using the {EM}
  algorithm,'' \emph{Journal of the Royal Statistical Society: Series B
  (Methodological)}, vol.~44, no.~2, pp. 226--233, 1982.

\bibitem{varadhan2008squarem}
R.~Varadhan and C.~Roland, ``Simple and globally convergent methods for
  accelerating the convergence of any {EM} algorithm,'' \emph{Scandinavian
  Journal of Statistics}, vol.~35, no.~2, pp. 335--353, 2008.

\bibitem{berlinet2007acceleration}
A.~Berlinet and C.~Roland, ``Acceleration schemes with application to the {EM}
  algorithm,'' \emph{Computational statistics \& data analysis}, vol.~51,
  no.~8, pp. 3689--3702, 2007.

\bibitem{jamshidian1993conjugate}
M.~Jamshidian and R.~I. Jennrich, ``Conjugate gradient acceleration of the {EM}
  algorithm,'' \emph{Journal of the American Statistical Association}, vol.~88,
  no. 421, pp. 221--228, 1993.

\bibitem{salakhutdinov2003optimization}
R.~Salakhutdinov, S.~T. Roweis, and Z.~Ghahramani, ``Optimization with {EM} and
  expectation-conjugate-gradient,'' in \emph{Proceedings of the 20th
  International Conference on Machine Learning (ICML-03)}, 2003, pp. 672--679.

\bibitem{he2012dynamic}
Y.~He and C.~Liu, ``The dynamic "expectation--conditional maximization either"
  algorithm,'' \emph{Journal of the Royal Statistical Society: Series B
  (Statistical Methodology)}, vol.~74, no.~2, pp. 313--336, 2012.

\bibitem{lange1995quasi}
K.~Lange, ``A quasi-newtonian acceleration of the {EM} algorithm,''
  \emph{Statistica Sinica. v5}, pp. 1--18, 1995.

\bibitem{jamshidian1997acceleration}
M.~Jamshidian and R.~I. Jennrich, ``Acceleration of the {EM} algorithm by using
  quasi-{N}ewton methods,'' \emph{Journal of the Royal Statistical Society:
  Series B (Statistical Methodology)}, vol.~59, no.~3, pp. 569--587, 1997.

\bibitem{henderson2019daarem}
N.~Henderson and R.~Varadhan, ``Damped {A}nderson acceleration with restarts
  and monotonicity control for accelerating {EM} and {EM}-like algorithms,''
  \emph{Journal of Computational and Graphical Statistics}, vol.~28, no.~4, pp.
  834--846, 2019.

\bibitem{zhou2011quasi}
H.~Zhou, D.~Alexander, and K.~Lange, ``A quasi-{N}ewton acceleration for
  high-dimensional optimization algorithms,'' \emph{Statistics and computing},
  vol.~21, no.~2, pp. 261--273, 2011.

\bibitem{meilijson1989fast}
I.~Meilijson, ``A fast improvement to the {EM} algorithm on its own terms,''
  \emph{Journal of the Royal Statistical Society: Series B (Methodological)},
  vol.~51, no.~1, pp. 127--138, 1989.

\bibitem{lange1995gradient}
K.~Lange, ``A gradient algorithm locally equivalent to the {EM} algorithm,''
  \emph{Journal of the Royal Statistical Society: Series B (Methodological)},
  vol.~57, no.~2, pp. 425--437, 1995.

\bibitem{anderson1965}
D.~G. Anderson, ``Iterative procedures for nonlinear integral equations,''
  \emph{J. Assoc. Comput. Mach.}, vol.~12, pp. 547--560, 1965.

\bibitem{fang2009two}
H.-r. Fang and Y.~Saad, ``Two classes of multisecant methods for nonlinear
  acceleration,'' \emph{Numerical Linear Algebra with Applications}, vol.~16,
  no.~3, pp. 197--221, 2009.

\bibitem{walker2011anderson}
H.~F. Walker and P.~Ni, ``Anderson acceleration for fixed-point iterations,''
  \emph{SIAM Journal on Numerical Analysis}, vol.~49, no.~4, pp. 1715--1735,
  2011.

\bibitem{plasse2013algorithm}
J.~H. Plasse, ``The {EM} algorithm in multivariate {G}aussian mixture models
  using {A}nderson acceleration,'' Master's thesis, Worcester Polytechnic
  Institute, April 2013.

\bibitem{Tibshirani2001gapstatistics}
R.~Tibshirani, G.~Walther, and T.~Hastie, ``Estimating the number of clusters
  in a data set via the gap statistic,'' \emph{Journal of Royal Statistical
  Society: Series B (Statistical Methodology)}, vol.~63, no.~2, pp. 411--423,
  2001.

\bibitem{everitt2014finite}
B.~S. Everitt, ``Finite mixture distributions,'' \emph{Wiley StatsRef:
  Statistics Reference Online}, 2014.

\bibitem{hasselblad1966estimation}
V.~Hasselblad, ``Estimation of parameters for a mixture of normal
  distributions,'' \emph{Technometrics}, vol.~8, no.~3, pp. 431--444, 1966.

\bibitem{behboodian1970mixture}
J.~Behboodian, ``On a mixture of normal distributions,'' \emph{Biometrika},
  vol.~34, no. 57 Part 1, pp. 215--217, 1970.

\bibitem{mackay2003information}
D.~J.~C. MacKay, \emph{Information theory, inference and learning
  algorithms}.\hskip 1em plus 0.5em minus 0.4em\relax USA: Cambridge University
  Press, 2003.

\bibitem{rousseau2011asymptotic}
J.~Rousseau and K.~Mengersen, ``Asymptotic behaviour of the posterior
  distribution in overfitted mixture models,'' \emph{Journal of the Royal
  Statistical Society: Series B (Statistical Methodology)}, vol.~73, no.~5, pp.
  689--710, 2011.

\bibitem{raykov2016adaptiveKmeans}
Y.~Raykov, A.~Boukouvalas, F.~Baig, and M.~Little, ``What to do when {K}-means
  clustering fails: A simple yet principled alternative algorithm,'' \emph{PLoS
  ONE}, vol.~11, no.~9, 2016.

\bibitem{Darken1990fastadaptkmeans}
C.~Darken and J.~Moody, ``Fast adaptive {K}-means clustering: some empirical
  results,'' in \emph{1990 IJCNN International Joint Conference on Neural
  Networks}, vol.~2.\hskip 1em plus 0.5em minus 0.4em\relax IEEE, 1990, pp.
  233--238.

\bibitem{scikit-learn-selectNG}
\BIBentryALTinterwordspacing
{Scikit-Learn Library}, \emph{Selecting the number of clusters with silhouette
  analysis on {K}-Means clustering}, 2007-2020. [Online]. Available:
  \url{https://scikit-learn.org/stable/auto_examples/cluster/plot_kmeans_silhouette_analysis.html}
\BIBentrySTDinterwordspacing

\bibitem{Rousseeuw1987silhouette}
P.~J. Rousseeuw, ``Silhouettes: a graphical aid to the interpretation and
  validation of cluster analysis,'' \emph{Journal of Computational and Applied
  Mathematics}, vol.~20, pp. 53--65, 1987.

\bibitem{Nanjundan2019silhouette}
S.~Nanjundan, S.~Sankaran, C.~R. Arjun, and P.~Anand, ``Identifying the number
  of clusters for {K}-means: A hypersphere density based approach,'' in
  \emph{International Conference on Computers, Communication and Signal
  Processing - 2019}, 2019.

\bibitem{xiang2020exact}
W.~Xiang, A.~Karfoul, C.~Yang, H.~Shu, and R.~L.~B. Jeann{\`e}s, ``An exact
  line search scheme to accelerate the {EM} algorithm: Application to
  {G}aussian mixture models identification,'' \emph{Journal of Computational
  Science}, vol.~41, p. 101073, 2020.

\bibitem{carlson1998fpa}
N.~N. Carlson and K.~Miller, ``Design and application of a gradient-weighted
  moving finite element code i: In one dimension,'' \emph{SIAM J. Sci.
  Comput.}, vol.~19, no.~3, pp. 728--765, 1998.

\bibitem{walker2017}
H.~An, X.~Jia, and H.~F. Walker, ``Anderson acceleration and application to the
  three-temperature energy equations,'' \emph{J. Comput. Phys.}, vol. 347,
  no.~15, pp. 1--19, 2017.

\bibitem{chen2019convergence}
X.~Chen and C.~Kelley, ``Convergence of the {EDIIS} algorithm for nonlinear
  equations,'' \emph{SIAM Journal on Scientific Computing}, vol.~41, no.~1, pp.
  A365--A379, 2019.

\bibitem{Pratapa2015restartpulay}
P.~R. Pratapa and P.~Suryanarayana, ``Restarted {P}ulay mixing for efficient
  and robust acceleration of fixed-point iterations,'' \emph{Chemical Physical
  Letters}, vol. 635, pp. 69--74, 2015.

\bibitem{Arthur07k-means++}
D.~Arthur and S.~Vassilvitskii, ``K-means++: the advantages of careful
  seeding,'' in \emph{In Proceedings of the 18th Annual ACM-SIAM Symposium on
  Discrete Algorithms}, 2007, pp. 1027--1035.

\bibitem{horn2012matrix}
R.~A. Horn and C.~R. Johnson, \emph{Matrix analysis}.\hskip 1em plus 0.5em
  minus 0.4em\relax Cambridge university press, 2012.

\bibitem{birdsall2004plasma}
C.~K. Birdsall and A.~B. Langdon, \emph{Plasma physics via computer
  simulation}.\hskip 1em plus 0.5em minus 0.4em\relax CRC press, 2004.

\bibitem{weibel1959spontaneously}
E.~S. Weibel, ``Spontaneously growing transverse waves in a plasma due to an
  anisotropic velocity distribution,'' \emph{Physical Review Letters}, vol.~2,
  no.~3, p.~83, 1959.

\bibitem{MatrixCookBook}
\BIBentryALTinterwordspacing
K.~B. Petersen and M.~S. Pedersen, ``The matrix cookbook,'' nov 2012, version
  20121115. [Online]. Available:
  \url{http://www2.compute.dtu.dk/pubdb/pubs/3274-full.html}
\BIBentrySTDinterwordspacing

\end{thebibliography}

\end{document}